\documentclass[acmsmall]{acmart}

\usepackage{footnote}
\usepackage{amsmath,amssymb,multicol,mathrsfs}
\usepackage[ruled,linesnumbered]{algorithm2e}
\usepackage{graphicx}
\usepackage{balance}

\usepackage{epstopdf}
\usepackage{multirow}
\usepackage{url}
\usepackage[skip=0pt]{subcaption}
\usepackage{color,soul}

\usepackage{tabularx}
 
\usepackage[normalem]{ulem}
\usepackage{enumitem}
\setlist{leftmargin=5.5mm}

\renewcommand\vec[1]{\overrightarrow{#1}}
\newcommand\cev[1]{\overleftarrow{#1}}
\newcommand{\etal}{\textit{et~al.}}

\AtBeginDocument{%
  \providecommand\BibTeX{{%
    \normalfont B\kern-0.5em{\scshape i\kern-0.25em b}\kern-0.8em\TeX}}}

\setcopyright{acmcopyright}
\acmJournal{TDS}
\acmYear{2020} \acmVolume{1} \acmNumber{1} \acmArticle{1} \acmMonth{1} \acmPrice{15.00}\acmDOI{10.1145/3419106}

\begin{document}

\title{Neural Abstractive Text Summarization with Sequence-to-Sequence Models}

\author{Tian Shi}
\email{tshi@vt.edu}
\affiliation{
  \institution{Virginia Tech}
}
\author{Yaser Keneshloo}
\email{yaserkl@vt.edu}
\affiliation{
  \institution{Virginia Tech}
}
\author{Naren Ramakrishnan}
\email{naren@cs.vt.edu}
\affiliation{
  \institution{Virginia Tech}
}
\author{Chandan K. Reddy}
\email{reddy@cs.vt.edu}
\affiliation{
  \institution{Virginia Tech}
}

\renewcommand{\shortauthors}{Shi, et al.}

\begin{abstract}
  In the past few years, neural abstractive text summarization with sequence-to-sequence (seq2seq) models have gained a lot of popularity. Many interesting techniques have been proposed to improve seq2seq models, making them capable of handling different challenges, such as saliency, fluency and human readability, and generate high-quality summaries. Generally speaking, most of these techniques differ in one of these three categories: network structure, parameter inference, and decoding/generation. There are also other concerns, such as efficiency and parallelism for training a model. In this paper, we provide a comprehensive literature survey on different seq2seq models for abstractive text summarization from the viewpoint of network structures, training strategies, and summary generation algorithms. Several models were first proposed for language modeling and generation tasks, such as machine translation, and later applied to abstractive text summarization. Hence, we also provide a brief review of these models. As part of this survey, we also develop an open source library, namely, Neural Abstractive Text Summarizer (NATS) toolkit, for the abstractive text summarization. An extensive set of experiments have been conducted on the widely used CNN/Daily Mail dataset to examine the effectiveness of several different neural network components. Finally, we benchmark two models implemented in NATS on the two recently released datasets, namely, Newsroom and Bytecup.
\end{abstract}



\maketitle

\vspace{-1mm}
\section{Introduction}

In the modern era of big data, retrieving useful information from a large number of textual documents is a challenging task, due to the unprecedented growth in the availability of blogs, news articles, and reports are explosive.
Automatic text summarization provides an effective solution for summarizing these documents.
The task of the text summarization is to condense long documents into short summaries while preserving the important information and meaning of the documents~\cite{radev2002introduction,allahyari2017text}. 
Having such short summaries, the text content can be retrieved, processed and digested effectively and efficiently.
Generally speaking, there are two ways to perform text summarization: Extractive and Abstractive~\cite{mani1999advances}. 
A method is considered to be \textit{extractive} if words, phrases, and sentences in the summaries are selected from the source articles~\cite{gambhir2017recent,
verma2017extractive,bhatia2016automatic,
allahyari2017text, lloret2012text, saggion2013automatic,nenkova2011automatic,das2007survey}.
They are relatively simple and can produce grammatically correct sentences.
The generated summaries usually persist salient information of source articles and have a good performance w.r.t. human-written summaries~\cite{verma2017extractive,wu2018learning,see2017get,zhou2018neural}.
On the other hand, abstractive text summarization has attracted much attention since it is capable of generating novel words using language generation models conditioned on representation of source documents~\cite{nallapati2016abstractive,rush2015neural}.
Thus, they have a strong potential of producing high-quality summaries that are verbally innovative and can also easily incorporate external knowledge~\cite{see2017get}. In this category, many deep neural network based models have achieved better performance in terms of the commonly used evaluation measures (such as \textit{ROUGE}~\cite{lin2004rouge} score) compared to traditional extractive approaches~\cite{paulus2017deep,celikyilmaz2018deep}.
In this paper, we primarily focus on the recent advances of recurrent neural network (RNN) based sequence-to-sequence (seq2seq) models for the task of abstractive text summarization.

\vspace{-2mm}
\subsection{RNN-based Seq2seq Models and Pointer-Generator Network}

Seq2seq models (see Fig.~\ref{fig:basic_seq2seq})~\cite{sutskever2014sequence,cho2014learning} have been successfully applied to a variety of natural language processing (NLP) tasks, such as machine translation~\cite{luong2015effective,wu2016google,bahdanau2014neural,shen2015minimum,klein2017opennmt}, headline generation~\cite{rush2015neural,chopra2016abstractive,shen2017recent}, text summarization~\cite{see2017get,nallapati2016abstractive}, and speech recognition~\cite{bahdanau2016end,graves2014towards,miao2015eesen}.
Inspired by the success of neural machine translation (NMT)~\cite{bahdanau2014neural}, Rush~\etal~\cite{rush2015neural} first introduced a neural attention seq2seq model with an attention based encoder and a Neural Network Language Model (NNLM) decoder to the abstractive sentence summarization task, which has achieved a significant performance improvement over conventional methods.
Chopra~\etal~\cite{chopra2016abstractive} further extended this model by replacing the feed-forward NNLM with a recurrent neural network (RNN).
The model is also equipped with a convolutional attention-based encoder and a 
RNN (Elman~\cite{elman1990finding} or LSTM~\cite{hochreiter1997long}) decoder, and outperforms other state-of-the-art models on commonly used benchmark datasets, i.e., the Gigaword corpus.
Nallapati~\etal~\cite{nallapati2016abstractive} introduced several novel elements to the RNN encoder-decoder architecture to address
critical problems in the abstractive text summarization, including using the following: (i) feature-rich encoder to capture keywords, (ii) a switching generator-pointer to model out-of-vocabulary (OOV) words, and (iii) the hierarchical attention to capture hierarchical document structures.
They also established benchmarks for these models on a CNN/Daily Mail dataset~\cite{hermann2015teaching,chen2016thorough}, which consists of pairs of news articles and multi-sentence highlights (summaries).
Before this dataset was introduced, many abstractive text summarization models have concentrated on compressing short documents to single sentence summaries~\cite{rush2015neural,chopra2016abstractive}.
For the task of summarizing long documents into multi-sentence summaries, these models have several shortcomings: 
1) They cannot accurately reproduce the salient information of source documents.
2) They cannot efficiently handle OOV words.
3) They tend to suffer from word- and sentence-level repetitions and generating unnatural summaries.
To tackle the first two challenges, See~\etal~\cite{see2017get} proposed a pointer-generator network that implicitly combines the abstraction with the extraction. 
This pointer-generator architecture can copy words from source texts via a pointer and generate novel words from a vocabulary via a generator.
With the pointing/copying mechanism~\cite{vinyals2015pointer,gu2016incorporating,gulcehre2016pointing,zeng2016efficient,cheng2016neural,miao2016language,song2018structure}, factual information can be reproduced accurately and OOV words can also be taken care in the summaries.
Many subsequent studies that achieved state-of-the-art performance have also demonstrated the effectiveness of the pointing/copying mechanism~\cite{paulus2017deep,celikyilmaz2018deep,jiang2018closed,gehrmann2018bottom}.
The third problem has been addressed by the coverage mechanism~\cite{see2017get}, intra-temporal and intra-decoder attention mechanisms~\cite{paulus2017deep}, and some other heuristic approaches, like forcing a decoder to never output the same trigram more than once during testing~\cite{paulus2017deep}.

\vspace{-2mm}
\subsection{Training Strategies}
There are two other non-trivial issues with the current seq2seq framework, i.e., \textit{exposure bias} and \textit{inconsistency of training and testing measurements}~\cite{bengio2015scheduled,venkatraman2015improving,ranzato2015sequence,keneshloo2018deep}.
Based on the neural probabilistic language model~\cite{bengio2003neural}, seq2seq models are usually trained by maximizing the likelihood of ground-truth tokens given their previous ground-truth tokens and hidden states (Teacher Forcing algorithm~\cite{bengio2015scheduled,williams1989learning}, see Fig.~\ref{fig:training}(b)).
However, at testing time (see Fig.~\ref{fig:training}(a)), previous ground-truth tokens are unknown, and they are replaced with tokens generated by the model itself.
Since the generated tokens have never been exposed to the decoder during training, the decoding error can accumulate quickly during the sequence generation.
This is known as \textit{exposure bias}~\cite{ranzato2015sequence}.
The other issue is the \textit{mismatch of measurements}.
Performance of seq2seq models is usually estimated with 
non-differentiable evaluation metrics, such as ROUGE~\cite{lin2004rouge} and BLEU~\cite{papineni2002bleu} scores, which are inconsistent with the log-likelihood function (cross-entropy loss) used in the training phase.
These problems are alleviated by the curriculum learning and reinforcement learning (RL) approaches.

\vspace{-1mm}
\subsubsection{Training with Curriculum and Reinforcement Learning Approaches}
Bengio~\etal~\cite{bengio2015scheduled} proposed a curriculum learning approach, known as \textit{scheduled sampling}, to slowly change the input of the decoder from ground-truth tokens to model generated ones. 
Thus, the proposed meta-algorithm bridges the gap between training and testing.
It is a practical solution for avoiding the exposure bias.
Ranzato~\etal~\cite{ranzato2015sequence} proposed a sequence level training algorithm, called 
MIXER (Mixed Incremental Cross-Entropy Reinforce), which consists of the cross entropy training, REINFORCE~\cite{williams1992simple} and curriculum learning~\cite{bengio2015scheduled}. 
REINFORCE can make use of any user-defined task specific reward (e.g., non-differentiable evaluation metrics), therefore, combining with curriculum learning, the proposed model is capable of addressing both issues of seq2seq models.
However, REINFORCE suffers from the high variance of gradient estimators and instability during training~\cite{bahdanau2016actor,rennie2017self,wu2016google}.
Bahdanau~\etal~\cite{bahdanau2016actor} proposed an actor-critic based RL method which has relatively lower variance for gradient estimators. 
In the actor-critic method, an additional critic network is trained to compute value functions given the policy from the actor network (a seq2seq model), and the actor network is trained based on the estimated value functions (assumed to be exact) from the critic network.
On the other hand, Rennie~\etal~\cite{rennie2017self} introduced a self-critical sequence training method (SCST) which has a lower variance compared to REINFORCE and does not need the second critic network.

\vspace{-1mm}
\subsubsection{Applications to Abstractive Text Summarization} 
RL algorithms for training seq2seq models have achieved success in a variety of language generation tasks, such as image captioning~\cite{rennie2017self}, machine translation~\cite{bahdanau2016actor}, and dialogue generation~\cite{li2016deep}.
Specific to the abstractive text summarization, 
Lin~\etal~\cite{ling2017coarse} introduced a coarse-to-fine attention framework for the purpose of summarizing long documents.
Their model parameters were learned with REINFORCE algorithm.
Zhang~\etal~\cite{zhang2017sentence} used REINFORCE algorithm and the curriculum learning strategy for the sentence simplification task.
Paulus~\etal~\cite{paulus2017deep} first applied the self-critic policy gradient algorithm to training their seq2seq model with the copying mechanism and obtained the state-of-the-art performance in terms of ROUGE scores~\cite{lin2004rouge}.
They proposed a mixed objective function that combines the RL loss with the traditional cross-entropy loss.
Thus, their method can both leverage the non-differentiable evaluation metrics and improve the readability.
Celikyilmaz~\etal~\cite{celikyilmaz2018deep} introduced a novel deep communicating agents method for abstractive summarization, where they also adopted the RL loss in their objective function.
Pasunuru~\etal~\cite{pasunuru2018multi} applied the self-critic policy gradient algorithm to train the pointer-generator network.
They also introduced two novel rewards (i.e., saliency and entailment rewards) in addition to ROUGE metric to keep the generated summaries salient and logically entailed.
Li~\etal~\cite{li2018actor} proposed a training framework based on the actor-critic method, where the actor network is an attention-based seq2seq model, and the critic network consists of a maximum likelihood estimator and a global summary quality estimator that is used to distinguish the generated and ground-truth summaries via a neural network binary classifier.
Chen~\etal~\cite{chen2018fast} proposed a compression-paraphrase multi-step procedure, for abstractive text summarization, which first extracts salient sentences from documents and then rewrites them.
In their model, they used an advantage actor-critic algorithm to optimize the sentence extractor for a better extraction strategy.
Keneshloo~\etal~\cite{keneshloo2018deep} conducted a comprehensive summary of various RL methods and their applications in training seq2seq models for different NLP tasks.
They also implemented these RL algorithms in an open source library (\url{https://github.com/yaserkl/RLSeq2Seq/}) constructed using the pointer-generator network~\cite{see2017get} as the base model.

\vspace{-1mm}
\subsection{Beyond RNN-based Seq2Seq Models}
Most of the prevalent seq2seq models that have attained state-of-the-art performance for sequence modeling and language generation tasks are RNN, especially long short-term memory (LSTM)~\cite{hochreiter1997long} and gated recurrent unit (GRU)~\cite{chung2014empirical}, based encoder-decoder models~\cite{sutskever2014sequence,bahdanau2014neural}.
Standard RNN models are difficult to train due to the vanishing and exploding gradients problems~\cite{bengio1994learning}. 
LSTM is a solution for vanishing gradients problem, but still does not address the exploding gradients issue.
This issue is recently solved using a gradient norm clipping strategy~\cite{pascanu2013difficulty}.
Another critical problem of RNN based models is the computation constraint for long sequences due to their inherent sequential dependence nature.
In other words, the current hidden state in a RNN is a function of previous hidden states.
Because of such dependence, RNN cannot be parallelized within a sequence along the time-step dimension (see Fig.~\ref{fig:basic_seq2seq}) during training and evaluation, and hence training them becomes major challenge for long sequences due to the computation time and memory constraints of GPUs~\cite{vaswani2017attention}.


Recently, it has been found that the convolutional neural network (CNN)~\cite{krizhevsky2012imagenet} based encoder-decoder models have the potential to alleviate the aforementioned problem, since they have better performance in terms of the following three considerations~\cite{kalchbrenner2016neural,bradbury2016quasi,gehring2017convolutional}.
1) A model can be parallelized during training and evaluation.
2) The computational complexity of the model is linear with respect to the length of sequences.
3) The model has short paths between pairs of input and output tokens, so that it can propagate gradient signals more efficiently~\cite{hochreiter2001gradient}.
Kalchbrenner~\etal~\cite{kalchbrenner2016neural} introduced a \textit{ByteNet} model which adopts the one-dimensional convolutional neural network of fixed depth to both the encoder and the decoder~\cite{oord2016wavenet}. 
The decoder CNN is stacked on top of the hidden representation of the encoder CNN, which ensures a shorter path between input and output.
The proposed ByteNet model has achieved state-of-the-art performance on a character-level machine translation task with parallelism and linear-time computational complexity~\cite{kalchbrenner2016neural}.
Bradbury~\etal~\cite{bradbury2016quasi} proposed a quasi-recurrent neural network (QRNN) encoder-decoder architecture, where both encoder and decoder are composed of convolutional layers and so-called `dynamic average pooling' layers~\cite{balduzzi2016strongly,bradbury2016quasi}.
The convolutional layers allow computations to be completely parallel across both mini-batches and sequence time-step dimensions, while they require less amount of time compared with computation demands for LSTM despite the sequential dependence still presents in the pooling layers~\cite{bradbury2016quasi}.
This framework has demonstrated to be effective by outperforming LSTM-based models on a character-level machine translation task with a significantly higher computational speed.
Recently, Gehring~\etal~\cite{gehring2016convolutional,gehring2017convolutional,fan2017controllable} attempted to build CNN based seq2seq models and apply them to large-scale benchmark datasets for sequence modeling.
In~\cite{gehring2016convolutional}, the authors proposed a convolutional encoder model, in which the encoder is composed of a succession of convolutional layers, and demonstrated its strong performance for machine translation.
They further constructed a convolutional seq2seq architecture by replacing the LSTM decoder with a CNN decoder and bringing in several novel elements, including gated linear units~\cite{dauphin2016language} and multi-step attention~\cite{gehring2017convolutional}.
The model also enables computations of all network elements parallelized, thus training and decoding can be much faster than the RNN models.
It also achieved competitive performance on several machine translation benchmark datasets.
More rencently, ConvS2S model~\cite{fan2018controllable,gehring2017convolutional} has been applied to the abstractive document summarization and outperforms the pointer-generator network~\cite{see2017get} on the CNN/Daily Mail dataset.

Vaswani~\etal~\cite{vaswani2017attention}
constructed a novel network architecture called  Transformer which only depends on feed-forward networks and a multi-head attention mechanism.
It has achieved superior performance in machine translation task with significantly less training time.
Currently, large Transformers \cite{vaswani2017attention,devlin2019bert,yang2019xlnet}, which are pre-trained on a massive text corpus with self-supervised objectives, have achieved superior results in a variety of downstream NLP tasks such as machine understanding \cite{devlin2019bert,liu2019multi}, question-answering \cite{clark2019electra,liu2019roberta}, and abstractive text summarization \cite{dong2019unified,zhang2019pegasus,lewis2019bart,raffel2019exploring,yan2020prophetnet,liu2019text}.
Zhang~\textit{et al.} \cite{zhang2019pegasus} demonstrated that their pre-trained encoder-decoder model can outperform previous state-of-the-art results \cite{narayan2018don,see2017get,grusky2018newsroom,fabbri2019multi,napoles2012annotated,koupaee2018wikihow,kim2019abstractive,sharma2019bigpatent,cohan2018discourse} on several datasets by fine-tuning with limited supervised examples, which shows that pre-trained models are promising candidates in zero-shot and low-resource summarization tasks.

\begin{figure*}[!t]
	\centering
	\vspace{-2mm}
	\includegraphics[width=0.9\textwidth,height=16em]{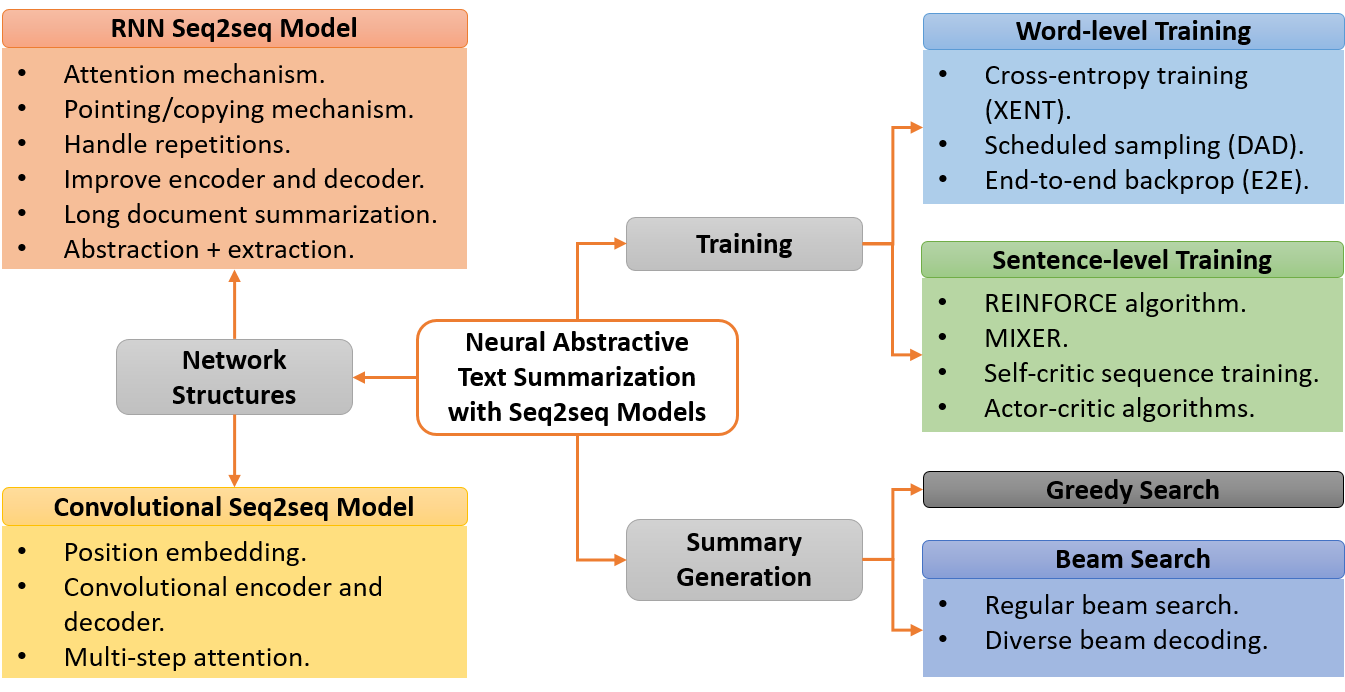}
	\vspace{-3mm}
	\caption{An overall taxonomy of topics on seq2seq models for neural abstractive text summarization.}
	\vspace{-5mm}
	\label{fig:taxonomy}
\end{figure*}

\vspace{-2mm}
\subsection{Other Studies}
So far, we primarily focused on the pointer-generator network, training neural networks with RL algorithms, CNN based seq2seq architectures, and Transformers. 
There are many other studies that aim to improve the performance of RNN seq2seq models for the task of abstractive text summarization from different perspectives and broaden their applications.

\vspace{-1mm}
\subsubsection{Network Structure and Attention}
The first way to boost the performance of seq2seq models is to design better network structures.
Zhou~\etal~\cite{zhou2017selective} introduced an information filter, namely, a selective gate network between the encoder and decoder.
This model can control the information flow from the encoder to the decoder via constructing a second level representation of the source texts with the gate network.
Zeng~\etal~\cite{zeng2016efficient} introduced a read-again mechanism to improve the quality of the representations of the source texts.
Tan~\etal~\cite{tan2017abstractive}
built a graph ranking model upon a hierarchical encoder-decoder framework, which enables the model to capture the salient information of the source documents and generate accurate, fluent and non-redundant summaries.
Xia~\etal~\cite{xia2017deliberation} proposed a deliberation network that passes the decoding process multiple times (deliberation process), to polish the sequences generated by the previous decoding process.
Li~\etal~\cite{li2017deep} incorporated a sequence of variational auto-encoders~\cite{kingma2013auto,rezende2014stochastic} into the decoder to capture the latent structure of the generated summaries.

\vspace{-1mm}
\subsubsection{Extraction + Abstraction}
Another way to improve the abstractive text summarization is to make use of the salient information from the extraction process.
Hsu~\etal~\cite{hsu2018unified}
proposed a unified framework
that takes advantage of both extractive and abstractive summarization using a novel attention mechanism, which is a combination of the sentence-level attention (based on the extractive summarization~\cite{nallapati2017summarunner}) and the word-level attention (based on the pointer-generator network~\cite{see2017get}), inspired by the intuition that words in less attended sentences should have lower attention scores.
Chen~\etal~\cite{chen2018fast} introduced a multi-step procedure, namely compression-paraphrase, for abstractive summarization, which first extracts salient sentences from documents and then rewrites them in order to get final summaries.
Li~\etal~\cite{li2018guiding} introduced a guiding generation model, where the
keywords in source texts is first retrieved with an extractive model~\cite{mihalcea2004textrank}.
Then, a guide network is applied to encode them to obtain the key information representations that will guide the summary generation process.

\vspace{-1mm}
\subsubsection{Long Documents}
Compared to short articles and texts with moderate lengths, there are many challenges that arise in long documents, such as difficulty in capturing the salient information~\cite{cohan2018discourse}.
Nallapati~\etal~\cite{nallapati2016abstractive} proposed a hierarchical attention model to capture hierarchical structures of long documents.
To make models scale-up to very long sequences, Ling~\etal~\cite{ling2017coarse} introduced a coarse-to-fine attention mechanism, which hierarchically reads and attends long documents (A document is split into many chunks of texts.).
By stochastically selecting chunks of texts during training, this approach can scale linearly with the number of chunks instead of the number of tokens.
Cohan~\etal~\cite{cohan2018discourse} proposed a discourse-aware attention model which has a similar idea to that of a hierarchical attention model.
Their model was applied to two large-scale datasets of scientific papers, i.e., arXiv and PubMed datasets.
Tan~\etal~\cite{tan2017abstractive} introduced a graph-based attention model which is built upon a hierarchical encoder-decoder framework where the pagerank algorithm~\cite{page1999pagerank} was used to calculate saliency scores of sentences.

\vspace{-1mm}
\subsubsection{Multi-Task Learning}
Multi-task learning has become a promising research direction for this problem since it allows seq2seq models to handle different tasks.
Pasunuru~\etal~\cite{pasunuru2017towards} introduced a multi-task learning framework, which incorporates knowledge from an entailment generation task into the abstractive text summarization task by sharing decoder parameters.
They further proposed a novel framework~\cite{guo2018soft} that is composed of two auxiliary tasks, i.e., question generation and
entailment generation, to improve their model for capturing the saliency and entailment for the abstractive text summarization.
In their model, different tasks share several encoder, decoder and attention layers.
Mccann~\etal~\cite{mccann2018natural} introduced a Natural Language Decathlon, a challenge that spans ten different tasks, including question-answering, machine translation, summarization, and so on.
They also proposed a multitask question answering network that can jointly learn all tasks without task-specific modules or parameters, since all tasks are mapped to the same framework of question-answering over a given context.

\vspace{-1mm}
\subsubsection{Beam Search}
Beam search algorithms have been commonly used in the decoding of different language generation tasks~\cite{wu2016google,see2017get}.
However, the generated candidate-sequences are usually lacking in diversity~\cite{gimpel2013systematic}. 
In other words, top-$K$ candidates are nearly identical, where $K$ is size of a beam.
Li~\etal~\cite{li2016diversity} replaced the log-likelihood objective function in the neural probabilistic language model~\cite{bengio2003neural} with Maximum Mutual Information (MMI)~\cite{bahl1986maximum} in their neural conversation models to remedy the problem.
This idea has also been applied to neural machine translation (NMT)~\cite{li2016mutual} to model the bi-directional dependency of source and target texts.
They further proposed a simple yet fast decoding algorithm that can generate diverse candidates and has shown performance improvement on the abstractive text summarization task~\cite{li2016simple}.
Vijayakumar~\etal~\cite{vijayakumar2016diverse} proposed generating diverse outputs by optimizing for a diversity-augmented objective function. 
Their method, referred to as Diverse Beam Search (DBS) algorithm, has been applied to image captioning, machine translation, and visual question-generation tasks.
Cibils~\etal~\cite{cibils2018diverse} introduced a meta-algorithm that first uses DBS to generate summaries, and then, picks candidates according to maximal marginal relevance~\cite{guo2010probabilistic} under the assumption that the most useful candidates should be close to the source document and far away from each other.
The proposed algorithm has boosted the performance of the pointer-generator network on CNN/Daily Mail dataset.

Despite many research papers that are published in the area of neural abstractive text summarization, there are few survey papers \cite{moratanch2016survey,rachabathuni2017survey,dalal2013survey} that provide a comprehensive study.
In this paper, we systematically review current advances of seq2seq models for the abstractive text summarization task from various perspectives, including network structures, training strategies, and sequence generation.
In addition to a literature survey, we also implemented some of these methods in an open-source library, namely NATS (\url{https://github.com/tshi04/NATS}).
Extensive set of experiments have been conducted on various benchmark text summarization datasets in order to examine the importance of different network components.
The main contributions of this paper can be summarized as follows:
\begin{itemize}[leftmargin=*]
\item Provide a comprehensive literature survey of current advances of seq2seq models with an emphasis on the abstractive text summarization.
\item Conduct a detailed review of the techniques used to tackle different challenges in RNN encoder-decoder architectures.
\item Review different strategies for training seq2seq models and approaches for generating summaries.
\item Provide an open-source library, which implements some of these models, and systematically investigate the effects of different network elements on the summarization performance.
\end{itemize}

The rest of this paper is organized as follows:
An overall taxonomy of topics on seq2seq models for neural abstractive text summarization is shown in Fig.~\ref{fig:taxonomy}.
A comprehensive list of papers published till date on the topic of neural abstractive text summarization have been summarized in Table~\ref{tab:paper2017} and~\ref{tab:paper2018}.
In Section~\ref{sec:rnn_encoder_decoder}, we introduce the basic seq2seq framework along with its extensions, including attention mechanism, pointing/copying mechanism, repetition handling, improving encoder or decoder, summarizing long documents and combining with extractive models.
Section~\ref{sec:training} summarizes different training strategies, including word-level training methods, such as cross-entropy training, and sentence-level training with RL algorithms.
In Section~\ref{sec:decoding}, we discuss generating summaries using the beam search algorithm and other diverse beam decoding algorithms.
In Section~\ref{sec:codes_exps}, we present details of our implementations and discuss our experimental results on the CNN/Daily Mail, Newsroom~\cite{grusky2018newsroom}, and Bytecup (\url{https://www.biendata.com/competition/bytecup2018/}) datasets.
We conclude this survey in Section~\ref{sec:conclusion}.

\vspace{-2mm}
\section{The RNN Encoder-Decoder Framework}
\label{sec:rnn_encoder_decoder}

In this section, we review different RNN-based encoder-decoder models for the neural abstractive text summarization.
We will start with the basic seq2seq framework and attention mechanism.
Then, we will describe more advanced network structures that can handle different challenges in the text summarization, such as repetition and out-of-vocabulary (OOV) words.
We will highlight various existing problems and proposed solutions.

\vspace{-2mm}
\subsection{Seq2seq Framework Basics}
\label{sec:seq2seq_framework}
A vanilla seq2seq framework for abstractive text summarization is composed of an encoder and a decoder.
The encoder reads a source article, denoted by $x=(x_1, x_2, ..., x_{J})$, and transforms it to hidden states $h^e=(h^e_1,h^e_2,...,h^e_{J})$; 
while the decoder takes these hidden states as the context input and outputs a summary $y=(y_1, y_2,...,y_{T})$.
Here, $x_i$ and $y_j$ are one-hot representations of the tokens in the source article and summary, respectively.
We use $J$ and $T$ to represent the number of tokens (document length) of the original source document and the summary, respectively.
A summarization task is defined as inferring a summary $y$ from a given source article $x$.

\begin{figure}[!tp]
	\centering
	\vspace{-4mm}
	\includegraphics[width=0.5\textwidth,height=12em]{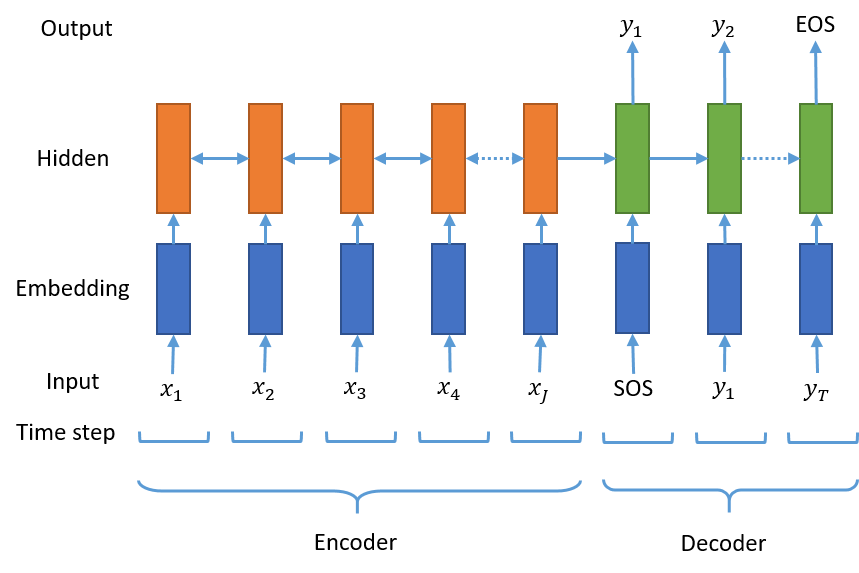}
	\vspace{-3mm}
	\caption{The basic seq2seq model. SOS and EOS represent the start and end of a sequence, respectively.}
	\vspace{-5mm}
	\label{fig:basic_seq2seq}
\end{figure}

Encoders and decoders can be feed-forward networks, CNN~\cite{gehring2016convolutional,gehring2017convolutional} or RNN. 
RNN architectures, especially long short term memory (LSTM)~\cite{hochreiter1997long} and gated recurrent unit (GRU)~\cite{cho2014learning}, have been most widely adopted for seq2seq models.
Fig.~\ref{fig:basic_seq2seq} shows a basic RNN seq2seq model with a bi-directional LSTM encoder and an LSTM decoder.
The bi-directional LSTM is considered since it usually gives better document representations compared to a forward LSTM.
The encoder reads a sequence of input tokens $x$ and turns them into a sequences of hidden states $h=(h_1, h_2, h_3, ..., h_{J})$. 
For the bi-directional LSTM, the input sequence is encoded as $\vec{h^e}$ and $\cev{h^e}$, where the right and left arrows denote the forward and backward temporal dependencies, respectively.
Superscript $e$ is the shortcut notation used to indicate that it is for the encoder.
During the decoding, the decoder takes the encoded representations of the source article (i.e., hidden and cell states $\vec{h^e_{J}}$, $\cev{h^e_{1}}$, $\vec{c^e_{J}}$, $\cev{c^e_{1}}$) as the input and generates the summary $y$.
In a simple encoder-decoder model, encoded vectors are used to initialize hidden and cell states of the LSTM decoder.
For example, we can initialize them as follows:
\begin{equation}
h^d_0=\tanh\left(W_\text{e2d}\big(\vec{h}^e_{J}\oplus\cev{h}^e_{1}\big)+b_\text{e2d}\right), c^d_0=\vec{c^e_{J}}\oplus\cev{c^e_{1}}
\end{equation}
Here, superscript $d$ denotes the decoder and $\oplus$ is a concatenation operator.
At each decoding step, we first update the hidden state $h^d_t$ conditioned on the previous hidden states and input tokens, i.e.,
$h_{t}^d=\text{LSTM}(h_{t-1}^d, E_{y_{t-1}})$.
Hereafter, we will not explicitly express the cell states in the input and output of LSTM, since only hidden states are passed to other parts of the model.
Then, the vocabulary distribution can be calculated with 
$P_{\text{vocab}, t} = \text{softmax}(W_\text{d2v}h_t^d+b_\text{d2v})$,
where $P_{\text{vocab}, t}$ is a vector whose dimension is the size of the vocabulary $\mathcal{V}$ and $\text{softmax}(v_t)=\frac{\exp(v_t)}{\sum_\tau\exp(v_\tau)}$ for each element $v_t$ of a vector $v$.
Therefore, the probability of generating the target token $w$ in the vocabulary $\mathcal{V}$ is denoted as $P_{\text{vocab}, t}(w)$.

This LSTM based encoder-decoder framework was the foundation of many neural abstractive text summarization models~\cite{nallapati2016abstractive,see2017get,paulus2017deep}.
However, there are many problems with this model. 
For example, the encoder is not well trained via back propagation through time~\cite{werbos1990backpropagation,sutskever2013training}, since the paths from encoder to the output are relatively far apart, which limits the propagation of gradient signals.
The accuracy and human-readability of generated summaries is also very low with a lot of OOV words\footnote{In the rest of this paper, we will use $<$unk$>$, i.e., unknown words, to denote OOV words.} and repetitions.
The rest of this section will discuss different models that were proposed in the literature to resolve these issues for producing better quality summaries.

\vspace{-2mm}
\subsection{Attention Mechanism}
\label{sec:attention}

\begin{figure}[!tp]
	\centering
	\vspace{-3mm}
	\begin{subfigure}[]{0.48\linewidth}
    \includegraphics[height=12em,width=\linewidth]{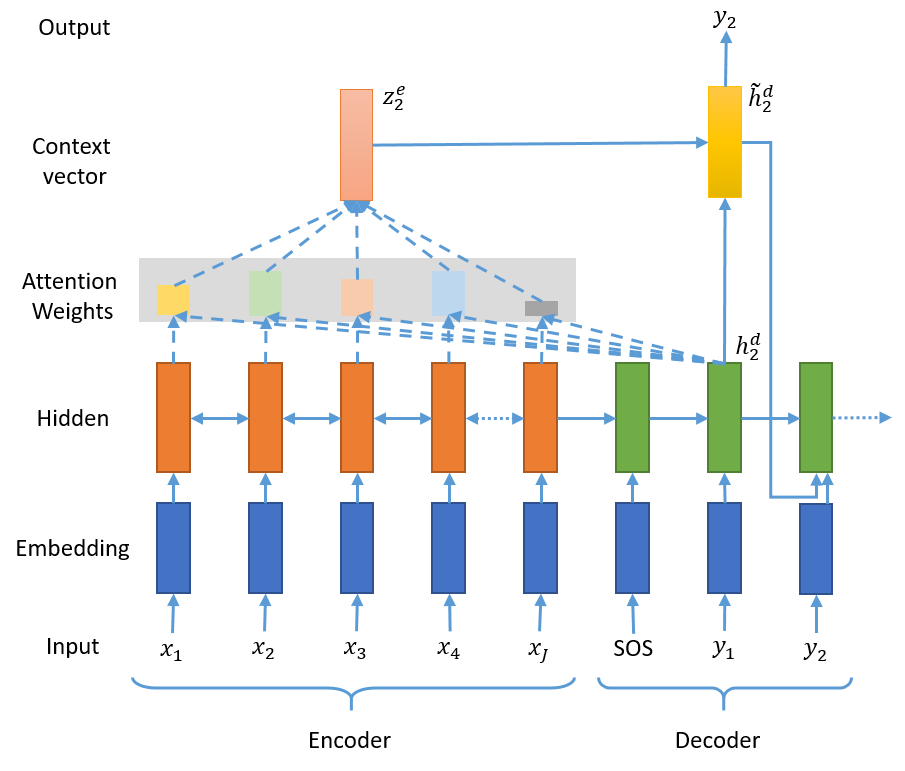}
    \caption{}
    \end{subfigure}
    \begin{subfigure}[]{0.48\linewidth}
    \includegraphics[height=12em,width=\linewidth]{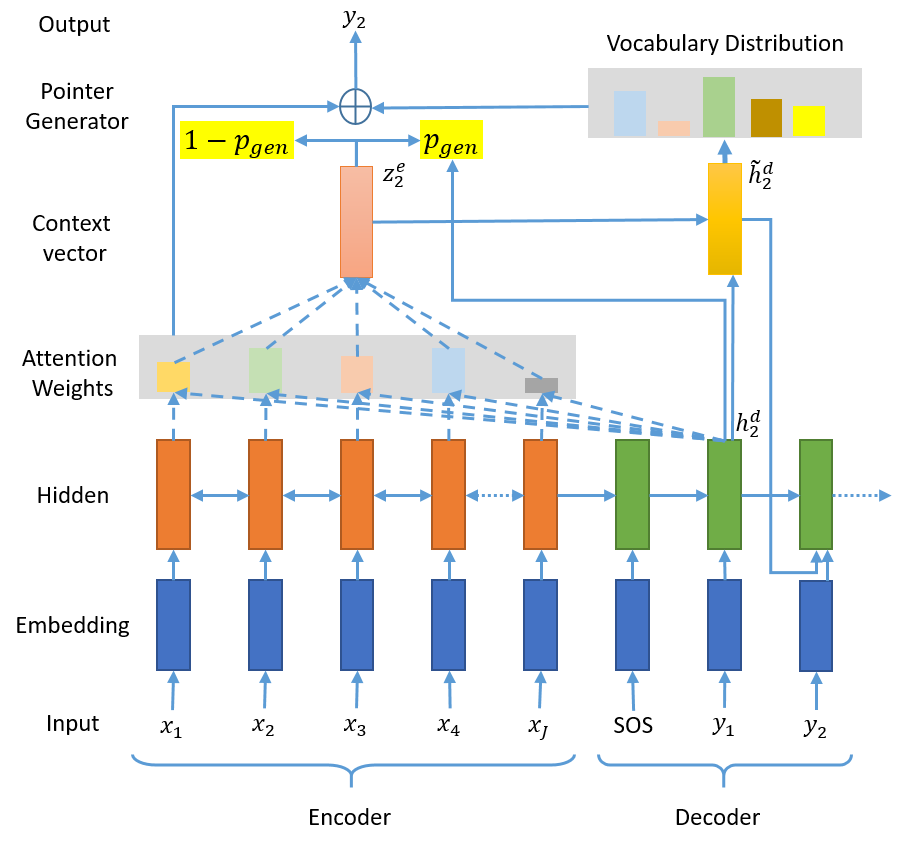}
    \caption{}
    \end{subfigure}
    \vspace{-4mm}
	\caption{(a) An attention-based seq2seq model. (b) The pointer-generator network.}
	\vspace{-5mm}
	\label{fig:attn_pg}
\end{figure}

The attention mechanism has achieved great success and is commonly used in seq2seq models for different natural language processing (NLP) tasks \cite{hu2018introductory}, such as machine translation~\cite{bahdanau2014neural,luong2015effective}, image captioning~\cite{xu2015show}, and neural abstractive text summarization~\cite{nallapati2016abstractive,see2017get,paulus2017deep}.
In an attention based encoder-decoder architecture (shown in Fig.~\ref{fig:attn_pg}(a)), the decoder not only takes the encoded representations (i.e., final hidden and cell states) of the source article as input, but also selectively focuses on parts of the article at each decoding step.
For example, suppose we want to compress the source input ``Kylian Mbappe scored two goals in four second-half minutes to send France into the World Cup quarter-finals with a thrilling 4-3 win over Argentina on Saturday." to its short version ``France beat Argentina 4-3 to enter quarter-finals.". 
When generating the token ``beat", the decoder may need to attend ``a thrilling 4-3 win" than other parts of the text. 
This attention can be achieved by an alignment mechanism~\cite{bahdanau2014neural}, which first computes the attention distribution of the source tokens and then lets the decoder know where to attend to produce a target token.
In the encoder-decoder framework depicted in Fig.~\ref{fig:basic_seq2seq} and~\ref{fig:attn_pg}(a), given all the hidden states of the encoder, i.e., $h^e=(h^e_1, h^e_2, ..., h^e_{J})$ and $h^e_j=\vec{h^e_j}\oplus\cev{h^e_j}$,
and the current decoder hidden state $h^d_t$, the attention distribution $\alpha^e_t$ over the source tokens is calculated as follows:
\begin{equation}
    \alpha^e_{tj} = \frac{\exp(s^e_{tj})}{\sum_{k=1}^{J}\exp(s^e_{tk})},
    \label{eqn:attn_weight}
\end{equation}
where the alignment score $s^e_{tj} = s(h^e_j, h^d_t)$ is obtained by the content-based score function, which has three alternatives as suggested in~\cite{luong2015effective}:
\begin{equation}
s(h^e_j, h^d_t)=
\begin{cases}
(h_j^e)^\top h_t^d & \text{dot} \\
(h_j^e)^\top W_\text{align}h_t^d & \text{general} \\
(v_\text{align})^\top\tanh\left(W_\text{align}\big(h_j^e\oplus h_t^d\big)+b_\text{align}\right) & \text{concat}
\end{cases}
\label{eqn:align}
\end{equation}
It should be noted that the number of additional parameters for `dot', `general' and `concat' approaches are $0$, $|h_j^e|\times|h_t^d|$ and $((|h^e_j|+|h_t^d|)\times|v_\text{align}|+2\times|v_\text{align}|)$, respectively. 
Here $|\cdot|$ represents the dimension of a vector.
The `general' and `concat' are commonly used score functions in the abstractive text summarization~\cite{see2017get,paulus2017deep}.
One of the drawbacks of `dot' method is that it requires $h^e_j$ and $h^d_t$ to have the same dimension.
With the attention distribution, we can naturally define the source side context vector for the target word as
\begin{equation}
z_t^e=\sum_{j=1}^{J}\alpha^e_{tj}h_j^e
\label{eqn:attn_context_vector}
\end{equation}
Together with the current decoder hidden state $h^d_t$, we get the attention hidden state~\cite{luong2015effective}
\begin{equation}
\tilde{h}_t^d=W_z\big(z_t^e\oplus h_t^d\big)+b_z
\label{eqn:attn_hidden}
\end{equation}
Finally, the vocabulary distribution is calculated by
\begin{equation}
P_{\text{vocab}, t} = \text{softmax}\left(W_\text{d2v}\tilde{h}_t^d+b_\text{d2v}\right)
\label{eqn:prob_vocab}
\end{equation}
When $t>1$, the decoder hidden state $h_{t+1}^d$ is updated by
\begin{equation}
h_{t+1}^d=\text{LSTM}\left(h_{t}^d, E_{y_{t}}\oplus \tilde{h}_{t}^d\right)
\label{eqn:atten_decode_hidden}
\end{equation}
where the input is concatenation of $E_{y_{t}}$ and $\tilde{h}_{t}^d$.

\vspace{-2mm}
\subsection{Pointing/Copying Mechanism}

The pointing/copying mechanism~\cite{vinyals2015pointer} represents a class of approaches that generate target tokens by directly copying them from input sequences based on their attention weights. 
It can be naturally applied to the abstractive text summarization since summaries and articles can share the same vocabulary~\cite{see2017get}.
More importantly, it is capable of dealing with out-of-vocabulary (OOV) words~\cite{nallapati2016abstractive,gu2016incorporating,gulcehre2016pointing,see2017get}.
A variety of studies have shown a boost in performance after incorporating the pointing/copying mechanism into the seq2seq framework~\cite{see2017get,paulus2017deep,celikyilmaz2018deep}.
In this section, we review several alternatives of this mechanism for the abstractive text summarization.

\vspace{-1mm}
\subsubsection{Pointer Softmax~\cite{gulcehre2016pointing}}
The basic architecture of pointer softmax is described as follows.
It consists of three fundamental components:
short-list softmax, location softmax and switching network.
At decoding step $t$, a short-list softmax $P_{\text{vocab},t}$ calculated by Eq.~(\ref{eqn:prob_vocab}) is used to predict target tokens in the vocabulary.
The location softmax gives locations of tokens that will be copied from the source article $x$ to the target $y_t$ based on attention weights $\alpha^e_t$.
With these two components, a switching network is designed to
determine whether to predict a token from the vocabulary or copy one from the source article if it is an OOV token.
The switching network is a multilayer perceptron (MLP) with a sigmoid activation function, which estimates the probability $p_{\text{gen},t}$ of generating tokens from the vocabulary based on the context vector $z_t^e$ and hidden state $h_t^d$ with
\begin{equation}
p_{\text{gen},t}=\sigma(W_{s,z}z_t^e+W_{s,h}h_t^d+b_s)
\end{equation}
where $p_{\text{gen},t}$ is a scalar and $\sigma(a)=\frac{1}{1+\exp{(-a)}}$ is a sigmoid activation function.
The final probability of producing the target token $y_t$ is given by the concatenation of vectors $p_{\text{gen},t} P_{\text{vocab},t}$ and $(1-p_{\text{gen},t})\alpha^e_t$. 

\vspace{-1mm}
\subsubsection{Switching Generator-Pointer~\cite{nallapati2016abstractive}}
Similar to the switching network in pointer softmax~\cite{gulcehre2016pointing}, the switching generator-pointer is also equipped with a `switch', which determines whether to generate a token from the vocabulary or point to one in the source article at each decoding step.
The switch is explicitly modeled by
\begin{equation}
p_{\text{gen},t}=\sigma(W_{s,z}z_t^e+W_{s,h}h_t^d+W_{s,E}E_{y_{t-1}}+b_s)
\label{eqn:pgen}
\end{equation}
If the switch is turned on, the decoder produces a word from the vocabulary with the distribution $P_{\text{vocab},t}$ (see Eq.~(\ref{eqn:prob_vocab})).
Otherwise, the decoder generates a pointer based on the attention distribution $\alpha_t^e$ (see Eq.~(\ref{eqn:attn_weight})), i.e.,
$p_j=\arg\max_{j\in \{1, 2, ...,J\}}\alpha^e_{tj}$, where $p_j$ is the position of the token in the source article.
When a pointer is activated, embedding of the pointed token $E_{x_j}$ will be used as an input for the next decoding step.

\vspace{-1mm}
\subsubsection{CopyNet~\cite{gu2016incorporating}}
CopyNet has a differentiable network architecture and can be easily trained in an end-to-end manner.
In this framework, the probability of generating a target token is a combination of the probabilities of two modes, i.e. generate-mode and copy-mode.
First, CopyNet represents unique tokens in the vocabulary and source sequence by $\mathcal{V}$ and $\mathcal{X}$, respectively, and builds an extended vocabulary $\mathcal{V}_\text{ext}=\mathcal{V}\cup\mathcal{X}\cup\text{$<$unk$>$}$.
Then, the vocabulary distribution over the extended vocabulary is calculated by
\begin{equation}
P_{\mathcal{V}_\text{ext}}(y_t)=P_{g}(y_t) + P_{c}(y_t)
\end{equation}
where $P_{g}$ and $P_{c}$ are also defined on $\mathcal{V}_\text{ext}$, i.e.,
\begin{equation}
P_{g}(y_t)=\begin{cases}
\frac{1}{Z}\exp{\psi_g({y_t})} & y_t\in\mathcal{V}\cup\text{$<$unk$>$}\\
0& \text{otherwise}
\end{cases}
\text{, }
P_{c}(y_t)=\begin{cases}
\frac{1}{Z}\sum_{j:x_j=y_t}\exp{\psi_c(x_j)}&y_t\in\mathcal{X}\\
0 & \text{otherwise}
\end{cases}
\end{equation}
Here, $Z$ is a normalization factor shared by both the above equations.
$\psi_g(y_t)$ is calculated with 
$\psi_g(y_t) = W_\text{d2v} \tilde{h}_t^d+b_\text{d2v}$ and
$\psi_c(x_j)$ is obtained by Eq.~(\ref{eqn:align}).

\vspace{-1mm}
\subsubsection{Pointer-Generator Network~\cite{see2017get}}
\label{sec:pointer-generator}
Pointer-generator network also has a differentiable network architecture (see Fig.~\ref{fig:attn_pg}(b)).
Similar to CopyNet~\cite{gu2016incorporating}, the vocabulary distribution over an extended vocabulary $\mathcal{V}_\text{ext}$ is calculated by
\begin{equation}
P_{\mathcal{V}_\text{ext}}(y_t)=p_{\text{gen},t}P_{g}(y_t)+(1-p_{\text{gen},t})P_{c}(y_t)
\end{equation}
where $p_{\text{gen},t}$ is obtained by Eq.~(\ref{eqn:pgen}).
Vocabulary distribution $P_{g}(y_t)$ and attention distribution $P_{c}(y_t)$ are defined as follows:
\begin{equation}
P_{g}(y_t)=\begin{cases}
P_{\text{vocab},t}(y_t) & y_t\in\mathcal{V}\cup\text{$<$unk$>$}\\
0& \text{otherwise}
\end{cases}
\text{, }
P_{c}(y_t)=\begin{cases}
\sum_{j:x_j=y_t}\alpha^e_{tj} & y_t\in\mathcal{X}\\
0 & \text{otherwise}
\end{cases}
\end{equation}

The pointer-generator network has been used as the base model for many abstractive text summarization models (see Table~\ref{tab:paper2017} and \ref{tab:paper2018}).
Finally, it should be noted that $p_{\text{gen},t}\in(0,1)$ in CopyNet and pointer-generator network can be viewed as a ``soft-switch" to choose between generation and copying, which is different from ``hard-switch" (i.e., $p_{\text{gen},t}=0,1$) in pointer softmax and switching generator-pointer~\cite{gulcehre2016pointing,nallapati2016abstractive,paulus2017deep}.

\vspace{-1mm}
\subsection{Repetition Handling}

One of the critical challenges for attention based seq2seq models is that the generated sequences have repetitions, since the attention mechanism tends to ignore the past alignment information~\cite{tu2016modeling,sankaran2016temporal}.
For summarization and headline generation tasks, model generated summaries suffer from both word-level and sentence-level repetitions.
The latter is specific to summaries which consist of several sentences~\cite{nallapati2016abstractive,see2017get,paulus2017deep}, such as those in CNN/Daily Mail dataset~\cite{nallapati2016abstractive} and Newsroom dataset~\cite{grusky2018newsroom}.
In this section, we review several approaches that have been proposed to overcome the repetition problem.

\vspace{-1mm}
\subsubsection{Temporal Attention~\cite{nallapati2016abstractive,paulus2017deep}}
Temporal attention method was originally proposed to deal with the attention deficiency problem in neural machine translation (NMT)~\cite{sankaran2016temporal}.
Nallapati~\etal~\cite{nallapati2016abstractive} have found that it can also overcome the problem of repetition when generating multi-sentence summaries, since it prevents the model from attending the same parts of a source article by tracking the past attention weights.
More formally, given the attention score $s^e_{tj}$ in Eq. (\ref{eqn:align}), we can first define a temporal attention score as~\cite{paulus2017deep}:
\begin{equation}
s^\text{temp}_{tj}=
\begin{cases}
\exp{(s^{e}_{tj})} & \text{if } t=1 \\
\frac{\exp{(s^e_{tj})}}{\sum_{k=1}^{t-1}\exp{(s^e_{kj})}} & \text{otherwise}
\end{cases}
\label{eqn:temporal}
\end{equation}
Then, the attention distribution and context vector (see Eq.~(\ref{eqn:attn_context_vector})) are calculated with
\begin{equation}
\alpha^\text{temp}_{tj}=\frac{s^\text{temp}_{tj}}{\sum_{k=1}^{J}s^\text{temp}_{tk}},\ 
z_t^e=\sum_{j=1}^{J}\alpha^\text{temp}_{tj}h_j^e.
\end{equation}
It can be seen from Eq. (\ref{eqn:temporal}) that, at each decoding step, the input tokens which have been highly attended will have a lower attention score via the normalization in time dimension.
As a result, the decoder will not repeatedly attend the same part of the source article.

\vspace{-1mm}
\subsubsection{Intra-decoder Attention~\cite{paulus2017deep}}
Intra-decoder attention is another technique to handle the repetition problem for long-sequence generations.
Compared to the regular attention based models, it allows a decoder to not only attend tokens in a source article but also keep track of the previously decoded tokens in a summary, so that the decoder will not repeatedly produce the same information.

For $t>1$, intra-decoder attention scores, denoted by $s^d_{t\tau}$, can be calculated in the same manner as the attention score $s^e_{tj}$ \footnote{We have to replace $h^e_j$ with $h^d_\tau$ in Eq.~(\ref{eqn:align}), where $\tau\in\{1,..., t-1\}$.}.
Then, the attention weight for each token is expressed as
\begin{equation}
\alpha^d_{t\tau}=\frac{\exp(\text{s}^d_{t\tau})}{\sum_{k=1}^{t-1}\exp(\text{s}^d_{tk})}
\end{equation}
With attention distribution, we can calculate the decoder-side context vector by taking linear combination of the decoder hidden states, i.e., $h_{<t}^d$, as
$z_t^d=\sum_{\tau=1}^{t-1}\alpha^d_{t\tau}h^d_{\tau}$.
The decoder-side and encoder-side context vector will be both used to calculate the vocabulary distribution.

\vspace{-1mm}
\subsubsection{Coverage~\cite{see2017get}}
\label{sec:coverage}
The coverage model was first proposed for the NMT task~\cite{tu2016modeling} to address the problems of the standard attention mechanism which tends to ignore the past alignment information.
Recently, \citeauthor{see2017get}~\cite{see2017get} introduced the coverage mechanism to the abstractive text summarization task.
In their model, they first defined a coverage vector $u_t^e$ as the sum of attention distributions of the previous decoding steps, i.e.,
$u^e_t=\sum_{j}^{t-1}\alpha^e_{tj}$.
Thus, it contains the accumulated attention information on each token in the source article during the previous decoding steps.
The coverage vector will then be used as an additional input to calculate the attention score
\begin{equation}
s_{tj}^e=(v_\text{align})^\top\tanh\left(W_\text{align}\big(h_j^e\oplus h_t^d\oplus u_t^e\big)+b_\text{align}\right)
\end{equation}
As a result, the attention at current decoding time-step is aware of the attention during the previous decoding steps.
Moreover, they defined a novel coverage loss to ensure that the decoder does not repeatedly attend the same locations when generating multi-sentence summaries.
Here, the coverage loss is defined as
\begin{equation}
\text{covloss}_t=\sum_j\min(\alpha^e_{tj},u^e_{tj}),
\end{equation}
which is upper bounded by $1$.

\subsubsection{Distraction~\cite{chen2016distraction}}
The coverage mechanism has also been used in~\cite{chen2016distraction} (known as \textit{distraction}) for the document summarization task.
In addition to the distraction mechanism over the attention, they also proposed a distraction mechanism over the encoder context vectors.
Both mechanisms are used to prevent the model from attending certain regions of the source article repeatedly.
Formally, given the context vector at current decoding step $z^e_t$ and all historical context vectors $(z^e_1, z^e_2,...,z^e_{t-1})$ (see Eq. (\ref{eqn:attn_context_vector})), the distracted context vector $z^{e,\text{dist}}_t$ is defined by
$z^{e,\text{dist}}_t = \tanh(W_{\text{dist},z}z^e_t-W_{\text{hist},z}\sum_j^{t-1}z^e_j)$,
where both $W_{\text{dist},z}$ and $W_{\text{hist},z}$ are diagonal parameter matrices.

\vspace{-1mm}
\subsection{Improving Encoded Representations}

Although LSTM and bi-directional LSTM encoders\footnote{GRU and bi-directional GRU are also often seen in abstractive summarization papers.} have been commonly used in the seq2seq models for the abstractive text summarization~\cite{nallapati2016abstractive,see2017get,paulus2017deep}, representations of the source articles are still believed to be sub-optimal.
In this section, we review some approaches that aim to improve the encoding process.


\vspace{-2mm}
\subsubsection{Selective Encoding~\cite{zhou2017selective}}
The selective encoding model was proposed for the abstractive sentence summarization task~\cite{zhou2017selective}.
Built upon an attention based encoder-decoder framework, it introduces a selective gate network into the encoder for the purpose of distilling salient information from source articles.
A second layer representation, namely, distilled representation, of a source article is constructed over the representation of the first LSTM layer (a bi-directional GRU encoder in this work.).
Formally, the distilled representation of each token in the source article is defined as
\begin{equation}
    h^{e}_{\text{sel},j} = \text{gate}_{\text{sel},j}\times h^e_j
\end{equation}
where $\text{gate}_{\text{sel},j}$ denotes the selective gate for token $x_j$ and is calculated as follows:
\begin{equation}
    \text{gate}_{\text{sel},j} = \sigma(W_{\text{sel}, h}h^e_j+W_{\text{sel},\text{sen}}h^e_\text{sen}+b_\text{sel})
\end{equation}
where $h^e_\text{sen}=\vec{h^e_{J}}\oplus \cev{h^e_{1}}$.
The distilled representations are then used for the decoding.
Such a gate network can control information flow from an encoder to a decoder and can also select salient information, therefore, it boosts the performance of the sentence summarization task~\cite{zhou2017selective}.

\vspace{-1mm}
\subsubsection{Read-Again Encoding~\cite{zeng2016efficient}}
Intuitively, read-again mechanism is motivated by human readers who read an article several times before writing a summary.
To simulate this cognitive process, a read-again encoder reads a source article twice and outputs two-level representations.
In the first read, an LSTM encodes tokens and the article as $(h^{e,1}_1, h^{e,1}_2, ..., h^{e,1}_{J})$ and $h^{e,1}_\text{sen}=h^{e,1}_{J}$, respectively.
In the second read, we use another LSTM to encode the source text based on the outputs of the first read.
Formally, the encoder hidden state of the second read $h_j^{e,2}$ is updated by
\begin{equation}
h_j^{e,2}=\text{LSTM}(h_{j-1}^{e,2}, E_{x_{j}}\oplus h^{e,1}_j\oplus h^{e,1}_\text{sen})
\end{equation}
The hidden states $(h^{e,2}_1, h^{e,2}_2, ..., h^{e,2}_{J})$ of the second read will be passed into decoders for summary generation.

\vspace{-2mm}
\subsection{Improving Decoder}

\vspace{-1mm}
\subsubsection{Embedding Weight Sharing~\cite{paulus2017deep}}
\label{sec:weight-sharing}
Sharing the embedding weights with the decoder is a practical approach that can boost the performance since it allows us to reuse the semantic and syntactic information in an embedding matrix during summary generation~\cite{inan2016tying,paulus2017deep}.
Suppose the embedding matrix is represented by $W_\text{emb}$, we can formulate the matrix used in the summary generation (see Eq. (\ref{eqn:prob_vocab})) as
$W_\text{d2v}=\tanh(W_\text{emb}^\top\cdot W_\text{proj})$.
By sharing model weights, the number of parameters is significantly less than a standard model since the number of parameters for $W_\text{proj}$ is $|h^e_j|\times|h^d_t|$, while that for $W_\text{d2v}$ is $|h^d_t|\times|\mathcal{V}|$, where $|h|$ represents the dimension of vector $h$ and $|\mathcal{V}|$ denotes size of the vocabulary.

\vspace{-1mm}
\subsubsection{Deep Recurrent Generative Decoder (DRGD)~\cite{li2017deep}}
Conventional encoder-decoder models calculate hidden states and attention weights in an entirely deterministic fashion, which limits the capability of representations and results in low quality summaries.
Incorporating variational auto-encoders (VAEs)~\cite{kingma2013auto,rezende2014stochastic} into the encoder-decoder framework provides a practical solution for this problem.
Inspired by the variational RNN proposed in~\cite{chung2015recurrent} to model the highly structured sequential data, Li~\etal~\cite{li2017deep} introduced a seq2seq model with DRGD that aims to capture latent structure information of summaries and improve the summarization quality.
This model employs GRU as the basic recurrent model for both encoder and decoder.
However, to be consistent with this survey paper, we will explain their ideas using LSTM instead.

There are two LSTM layers to calculate the decoder hidden state $h_t^d$.
At the decoding step $t$, the first layer hidden state $h^{d,1}_t$ is updated by $h^{d,1}_t=\text{LSTM}^{1}(h^{d,1}_{t-1}, E_{y_{t-1}})$.
Then, the attention weights $\alpha^e_{tj}$ and the context vector $z^{d,1}_t$ are calculated with the encoder hidden state $h^e$ and the first layer decoder hidden state $h^{d,1}_t$ using Eqs. (\ref{eqn:attn_weight}), (\ref{eqn:align}) and (\ref{eqn:attn_context_vector}).
For the second layer, the hidden state $h^{d,2}_t$ is updated with $h^{d,2}_t=\text{LSTM}^2(h^{d,2}_{t-1}, E_{y_{t-1}}\oplus z^{d,1}_t)$.
Finally, the decoder hidden state is obtained by $h^d_t=h^{d,1}_t\oplus h^{d,2}_t$, where $h^d$ is also referred to as the deterministic hidden state.

VAE is incorporated into the decoder to capture latent structure information of summaries which is represented by a multivariate Gaussian distribution.
By using a reparameterization trick~\cite{rezende2014stochastic,doersch2016tutorial}, latent variables can be first expressed as $\xi_t = \mu_t + \eta_t\otimes\epsilon$,
where the noise variable $\epsilon\sim\mathcal{N}(0,I)$, and Gaussian parameters $\mu_t$ and $\eta_t$ in the network are calculated by
\begin{equation}
\mu_t = W_{\text{vae},\mu} h^\text{enc}_t + b_{\text{vae},\mu},
\log(\eta_t^2) = W_{\text{vae},\eta} h^\text{enc}_t + b_{\text{vae},\eta}
\end{equation}
where $h^\text{enc}_t$ is a hidden vector of the encoding process of the VAE and defined as
\begin{equation}
    h^\text{enc}_t = \sigma(W_{\text{enc},\xi}\xi_{t-1}+W_{\text{enc},y}E_{y_{t-1}}+W_{\text{enc},h}h^d_{t-1}+b_\text{enc})
\end{equation}
With the latent structure variables $\xi_t$, the output hidden states $h^\text{dec}_t$ can be formulated as
\begin{equation}
    h^\text{dec}_t = \tanh(W_{\text{dec},\xi}\xi_t+W_{\text{dec},h}h^{d,2}_t+b_\text{dec})
\end{equation}
Finally, the vocabulary distribution is calculated by
$P_{\text{vocab}, t} = \text{softmax}(W_\text{d2v}h_t^\text{dec}+b_\text{d2v})$.

We primarily focused on the network structure of DRGD in this section.
The details of VAE and its derivations can be found in~\cite{kingma2013auto,li2017deep,rezende2014stochastic,doersch2016tutorial}.
In DRGD, VAE is incorporated into the decoder of a seq2seq model, more recent works have also used VAE in the attention layer~\cite{bahuleyan2017variational} and for the sentence compression task~\cite{miao2016language}.

\vspace{-1mm}
\subsection{Summarizing Long Document}

Compared to sentence summarization, the abstractive summarization for very long documents has been relatively less investigated.
Recently, attention based seq2seq models with pointing/copying mechanism have shown their power in summarizing long documents with 400 and 800 tokens~\cite{see2017get,paulus2017deep}.
However, performance improvement primarily attributes to copying and repetition/redundancy avoiding techniques~\cite{nallapati2016abstractive,see2017get,paulus2017deep}.
For very long documents, we need to consider several important factors to generate high quality summaries, such as saliency (i.e., significance), fluency, coherence and novelty~\cite{tan2017abstractive}.
Usually, seq2seq models combined with the beam search decoding algorithm can generate fluent and human-readable sentences.
In this section, we review models that aim to improve the performance of long document summarization from the perspective of saliency.

Seq2seq models for long document summarization usually consists of an encoder with a hierarchical architecture which is used to capture the hierarchical structure of the source documents.
The top-level salient information includes the important sentences~\cite{nallapati2016abstractive,tan2017abstractive}, chunks of texts~\cite{ling2017coarse}, sections~\cite{cohan2018discourse}, and paragraphs~\cite{celikyilmaz2018deep}, 
while the lower-level salient information represents keywords.
Hereafter, we will use the term `chunk' to represent the top-level information.
The hierarchical encoder first encodes tokens in a chunk for the chunk representation, and then, encodes different chunks in a document for the document representation.
In this paper, we only consider the single-layer forward LSTM\footnote{The deep communicating agents model~\cite{celikyilmaz2018deep} , which requires multiple layers of bi-directional LSTM, falls out of the scope of this survey.} for both word and chunk encoders.

Suppose, the hidden states of chunk $i$ and word $j$ in this chunk are represented by $h^\text{chk}_i$ and $h^\text{wd}_{ij}$.
At decoding step $t$, we can calculate word-level attention weight $\alpha^{\text{wd},t}_{ij}$ for the current decoder hidden state $h^d_t$ with
$\alpha^{\text{wd},t}_{ij}=\frac{\exp(s^{\text{wd},t}_{ij})}{\sum_{k,l}\exp(s^{\text{wd},t}_{kl})}$
At the same time, we can also calculate chunk-level attention weight $\alpha^{\text{chk},t}_{i}$ by
$\alpha^{\text{chk},t}_{i}=\frac{\exp(s^{\text{chk},t}_{i})}{\sum_{k}\exp(s^{\text{chk},t}_{k})}$,
where both alignment scores $s^{\text{wd},t}_{ij}=s^{\text{wd}}(h^\text{wd}_{ij},h^d_t)$ and $s^{\text{chk},t}_{i}=s^{\text{chk}}(h^\text{chk}_{i},h^d_t)$ can be calculated using Eq.~(\ref{eqn:align}).
In this section, we will review four different models that are based on the hierarchical encoder for the task of long document text summarization.


\vspace{-1mm}
\subsubsection{Hierarchical Attention~\cite{nallapati2016abstractive}}

The intuition behind a hierarchical attention is that 
words in less important chunks should be less attended.
Therefore, with chunk-level attention distribution $\alpha^{\text{chk},t}$ and word-level attention distribution $\alpha^{\text{wd},t}$, we first calculate re-scaled word-level attention distribution by
\begin{equation}
\alpha^{\text{scale},t}_{ij}=\frac{\alpha^{\text{chk},t}_{i}\alpha^{\text{wd},t}_{ij}}{\sum_{k,l}\alpha^{\text{chk},t}_{k}\alpha^{\text{wd},t}_{kl}}
\label{eqn:hier_attn_scale}
\end{equation}
This re-scaled attention will then be used to calculate the context vector using Eq. (\ref{eqn:attn_context_vector}), i.e.,
$z_t^e=\sum_{i,j}\alpha^{\text{scale},t}_{ij}h^\text{wd}_{ij}$.
It should be noted that such hierarchical attention framework is different from the hierarchical attention network proposed in~\cite{yang2016hierarchical}, where the chunk representation is obtained using
$z^{\text{wd},t}_i=\sum_j\alpha^{\text{wd},t}_{ij}h^\text{wd}_{ij}$
instead of the last hidden state of the word-encoder.

\subsubsection{Discourse-Aware Attention~\cite{cohan2018discourse}}
The idea of the discourse-aware attention is similar to that of the hierarchical attention giving Eq.~(\ref{eqn:hier_attn_scale}).
The main difference between these two attention models is that the re-scaled attention distribution in the discourse-aware attention is calculated by 
\begin{equation}
    \alpha^{\text{scale},t}_{ij}=\frac{\exp(\alpha^{\text{chk},t}_i s^{\text{wd},t}_{ij})}{\sum_{k,l}\exp(\alpha^{\text{chk},t}_k s^{\text{wd},t}_{kl})},
\end{equation}
where $s^{\text{wd},t}_{ij}$ is an alignment score instead of an attention weight.

\vspace{-1mm}
\subsubsection{Coarse-to-Fine Attention~\cite{ling2017coarse}}
The coarse-to-fine (C2F) attention was proposed for computational efficiency.
Similar to the hierarchical attention~\cite{nallapati2016abstractive}, the proposed model also has both chunk-level attention and word-level attention.
However, instead of using word-level hidden states in all chunks for calculating the context vector, the C2F attention method first samples a chunk $i$ from the chunk-level attention distribution, and then calculates the context vector using
$z_t^e=\sum_{j}\alpha^{\text{scale},t}_{ij}h^\text{wd}_{ij}$.
At the test time, the stochastic sampling of the chunks will be replaced by a greedy search.

\vspace{-1mm}
\subsubsection{Graph-based Attention~\cite{tan2017abstractive}}
The aforementioned hierarchical attention mechanism implicitly captures the chunk-level salient information, where the importance of a chunk is determined solely by its attention weight.
In contrast, the graph-based attention framework allows us to calculate the saliency scores explicitly using the pagerank algorithm~\cite{page1999pagerank,haveliwala2002topic} on a graph
whose vertices and edges are chunks of texts and their similarities, respectively.
Formally, at the decoding time-step $t$, saliency scores for all input chunks are obtained by
$f^t=(1-\lambda)(I-\lambda W^\text{adj}(t)D_\text{adj}^{-1}(t))^{-1}\chi_{\mathcal{T}}$,
where adjacent matrix $W^\text{adj}$ (similarity of chunks) is calculated by
$W^\text{adj}_{ij}=h^\text{chk}_iW_\text{par}h^\text{chk}_j$.
$D_\text{adj}$ is a diagonal matrix with its $(i,i)$-element equal to the sum of the $i^\text{th}$ column of $W^\text{adj}$.
$\lambda$ is a damping factor.
The vector $\chi_\mathcal{T}$ is defined as 
$\chi_{\mathcal{T},i}=\begin{cases}\frac{1}{|\mathcal{T}|} & i\in \mathcal{T}\\ 0 & \text{otherwise}\end{cases}$,
where $\mathcal{T}$ is a topic (see~\cite{haveliwala2002topic,tan2017abstractive} for more details).
Finally, the graph-based attention distribution over a chunk can be obtained by
\begin{equation}
    \alpha^{\text{chk},t}_{i} = \frac{\max(f_i^t-f_i^{t-1}, 0)}{\sum_k(\max(f_k^t-f_k^{t-1}, 0))}
\end{equation}
where $f^0_i$ is initialized with $0$.
It can be seen that the graph-based attention mechanism will focus on chunks that rank higher than the previous decoding step, i.e., $f_i^t>f_i^{t-1}$.
Therefore, it provides an efficient way to select salient information from source documents.

\begin{table}[!tp]
    \vspace{-3mm}
	\caption{An overview of different seq2seq models for the neural abstractive text summarization (2015-2017).}
	\vspace{-3mm}
	\centering
	\label{tab:paper2017}
    \resizebox{1.\linewidth}{!}{
	\begin{tabular}{|c|l|>{\raggedright\arraybackslash}p{13em}|>{\raggedright\arraybackslash}p{13em}|>{\raggedright\arraybackslash}p{4em}|>{\raggedright\arraybackslash}p{5em}|>{\raggedright\arraybackslash}p{8em}|>{\raggedright\arraybackslash}p{5em}|}
		\hline
         \bf Year & \bf Reference & \bf Highlights & \bf Framework & \bf Training & \bf Optimizer & \bf Datasets & \bf Metrics \\\hline
         
         & Rush~\etal~\cite{rush2015neural} & Attention Based Summarization (ABS) & Bag-of-words, Convolution, Attention $\rightarrow$ Neural Network Language Model (NNLM) & XENT & SGD & DUC, Gigaword & ROUGE \\\cline{2-8}
         2015 & lopyrev~\etal~\cite{lopyrev2015generating} & Simple Attention & LSTM $\rightarrow$ LSTM & XENT & RMSProp & Gigaword & BLEU \\\cline{2-8}
         & Ranzato~\etal~\cite{ranzato2015sequence} & Sequence-level Training & Elman, LSTM $\rightarrow$ Elman, LSTM & XENT, DAD, E2E, MIXER & SGD & Gigaword & ROUGE, BLEU\\\hline
         
         & Chopra~\etal~\cite{chopra2016abstractive} & Recurrent Attentive Summarizer & Convolution Encoder, Attentive Encoder $\rightarrow$ Elman, LSTM & XENT & SGD & DUC, Gigaword & ROUGE \\\cline{2-8}
         & Nallapati~\etal~\cite{nallapati2016abstractive} & Switch Generator-Pointer, Temporal-Attention, Hierarchical-Attention & RNN, Feature-rich Encoder $\rightarrow$ RNN & XENT & Adadelta & DUC, Gigaword, CNN/DM & ROUGE \\\cline{2-8}
         & Miao~\etal~\cite{miao2016language} & Auto-encoding Sentence Compression, Forced-Attention Sentence Compression, Pointer Network & Encoder $\rightarrow$ Compressor $\rightarrow$ Decoder & XENT+RL & Adam & Gigaword & ROUGE\\\cline{2-8}
         2016 & Chen~\etal~\cite{chen2016distraction} & Distraction & GRU $\rightarrow$ GRU & XENT & Adadelta & CNN, LCSTS & ROUGE\\\cline{2-8}
         & Gulcehre~\etal~\cite{gulcehre2016pointing} & Pointer softmax & GRU $\rightarrow$ GRU & XENT & Adadelta & Gigaword & ROUGE \\\cline{2-8}
         & Gu~\etal~\cite{gu2016incorporating} & CopyNet & GRU $\rightarrow$ GRU & XENT & SGD & LCSTS & ROUGE \\\cline{2-8}
         & Zeng~\etal~\cite{zeng2016efficient} & Read-again, Copy Mechanism & LSTM/GRU/Hierarchical read-again encoder $\rightarrow$ LSTM & XENT & SGD & DUC, Gigaword & ROUGE\\\cline{2-8}
         & Takase~\etal~\cite{takase2016neural} & Abstract Meaning Representation (AMR) based on ABS. & Attention-based AMR encoder $\rightarrow$ NNLM & XENT & SGD & DUC, Gigaword & ROUGE \\\hline
         
         & See~\etal~\cite{see2017get} & Pointer-Generator Network, Coverage & LSTM $\rightarrow$ LSTM & XENT & Adadelta & CNN/DM & ROUGE, METER \\\cline{2-8}
         & Paulus~\etal~\cite{paulus2017deep} & A Deep Reinforced Model, Intra-temporal and Intra-decoder Attention, Weight Sharing & LSTM $\rightarrow$ LSTM & XENT + RL & Adam & CNN/DM & ROUGE, Human\\\cline{2-8}
         & Zhou~\etal~\cite{zhou2017selective} & Selective Encoding, Abstractive Sentence Summarization & GRU $\rightarrow$ GRU & XENT & SGD & DUC, Gigaword, MSR-ATC &  ROUGE \\\cline{2-8}
         & Xia~\etal~\cite{xia2017deliberation} & Deliberation Networks & LSTM $\rightarrow$ LSTM & XENT & Adadelta & Gigaword & ROUGE \\\cline{2-8}
         & Nema~\etal~\cite{nema2017diversity} & Query-based, Diversity based Attention & GRU query encoder, document encoder $\rightarrow$ GRU & XENT & Adam & Debatepedia & ROUGE\\\cline{2-8}
         & Tan~\etal~\cite{tan2017abstractive} & Graph-based Attention & Hierarchical Encoder $\rightarrow$ LSTM & XENT & Adam & CNN/DM, CNN, DailyMail & ROUGE\\\cline{2-8}
         & Ling~\etal~\cite{ling2017coarse} & Coarse-to-fine Attention & LSTM $\rightarrow$ LSTM & RL & SGD & CNN/DM & ROUGE, PPL\\\cline{2-8}
         2017 & Zhang~\etal~\cite{zhang2017sentence} & Sentence Simplification, Reinforcement Learning & LSTM $\rightarrow$ LSTM & RL & Adam & Newsela, WikiSmall, WikiLarge & BLEU, FKGL, SARI \\\cline{2-8}
         & Li~\etal~\cite{li2017deep} & Deep Recurrent Generative Decoder (DRGD) & GRU $\rightarrow$ GRU, VAE & XENT, VAE & Adadelta & DUC, LCSTS & ROUGE \\\cline{2-8}
         & Liu~\etal~\cite{liu2017generative} & Adversarial Training & Pointer-Generator Network & GAN & Adadelta & CNN/DM & ROUGE, Human \\\cline{2-8}
         & Pasunuru~\etal~\cite{pasunuru2017towards} & Multi-Task with Entailment Generation & LSTM document encoder and premise Encoder $\rightarrow$ LSTM Summary and Entailment Decoder & Hybrid-Objective & Adam & DUC, Gigaword, SNLI & ROUGE, METEOR, BLEU, CIDEr-D\\\cline{2-8}
         & Gehring~\etal~\cite{gehring2017convolutional}& Convolutional Seq2seq, Position Embeddings, Gated Linear Unit, Multi-step Attention  & CNN $\rightarrow$ CNN & XENT & Adam & DUC, Gigaword & ROUGE\\\cline{2-8}
         & Fan~\etal~\cite{fan2017controllable} & Convolutional Seq2seq, Controllable  & CNN $\rightarrow$ CNN & XENT & Adam & DUC, CNNDM & ROUGE, Human \\\hline
	\end{tabular}}
	\vspace{-5mm}
\end{table}

\begin{table}[!tp]
    \vspace{-2mm}
	\caption{An overview of different seq2seq models for the neural abstractive text summarization (2018).}
	\vspace{-3mm}
	\centering
	\label{tab:paper2018}
    \resizebox{1.\linewidth}{!}{
	\begin{tabular}{|c|l|>{\raggedright\arraybackslash}p{13em}|>{\raggedright\arraybackslash}p{13em}|>{\raggedright\arraybackslash}p{4em}|>{\raggedright\arraybackslash}p{5em}|>{\raggedright\arraybackslash}p{8em}|>{\raggedright\arraybackslash}p{5em}|}
		\hline
         \bf Year & \bf Reference & \bf Highlights & \bf Framework & \bf Training & \bf Optimizer & \bf Datasets & \bf Metrics \\\hline

         & Celikyilmaz~\etal~\cite{celikyilmaz2018deep}& Deep Communicating Agents, Semantic Cohesion Loss & LSTM $\rightarrow$ LSTM & Hybrid-Objective & Adam & CNN/DM,NYT & ROUGE, Human\\\cline{2-8}
         & Chen~\etal~\cite{chen2018fast} & Reinforce-Selected Sentence Rewriting & LSTM Encoder $\rightarrow$ Extractor $\rightarrow$ Abstractor & XENT + RL & SGD  & DUC, CNN/DM & ROUGE, Human\\\cline{2-8}
         & Hsu~\etal~\cite{hsu2018unified} & Abstraction + Extraction, Inconsistency Loss & Extractor: GRU. Abstractor:  Pointer-generator Network & Hybrid-Objective + RL & Adadelta & CNN/DM & ROUGE, Human \\\cline{2-8}
         & Li~\etal~\cite{li2018actor} & Actor-Critic & GRU $\rightarrow$ GRU & RL & Adadelta & DUC, CNN/DM, LCSTS & ROUGE \\\cline{2-8}
         & Li~\etal~\cite{li2018guiding} & Abstraction + Extraction, Key Information Guide Network (KIGN) & KIGN: LSTM. Framework: Pointer-Generator Network & XENT & Adadelta & CNN/DM & ROUGE \\\cline{2-8}
         & Lin~\etal~\cite{lin2018global}& Global Encoding, Convolutional Gated Unit & LSTM $\rightarrow$ LSTM & XENT & Adam & Gigaword, LCSTS & ROUGE \\\cline{2-8}
         & Pasunuru~\etal~\cite{pasunuru2018multi} & Multi-Reward Optimization for RL: ROUGE, Saliency and Entailment. & LSTM $\rightarrow$ LSTM & RL & Adam & DUC, CNN/DM, SNLI, MultiNLI, SQuAD & ROUGE, Human \\\cline{2-8}
         & Song~\etal~\cite{song2018structure} & Structured-Infused Copy Mechanisms & Pointer-Generator Network & Hybrid-Objective & Adam & Gigaword & ROUGE, Human \\\cline{2-8}
         2018 &\citeauthor{cohan2018discourse}~\cite{cohan2018discourse} & Discourse Aware Attention & Hierarchical RNN LSTM Encoder $\rightarrow$ LSTM & XENT & Adagrad & PubMed, arXiv & ROUGE\\\cline{2-8}
         & Guo~\etal~\cite{guo2018soft} & Multi-Task Summarization with Entailment and Question Generation & Multi-Task Encoder-Decoder Framework & Hybrid-Objective & Adam & DUC, Gigaword, SQuAD, SNLI & ROUGE, METEOR \\\cline{2-8}
         & Cibils~\etal~\cite{cibils2018diverse} & Diverse Beam Search, Plagiarism and Extraction Scores &  Pointer-Generator Network & XENT & Adagrad & CNN/DM & ROUGE \\\cline{2-8}
         & Wang~\etal~\cite{wang2018reinforced} & Topic Aware Attention & CNN $\rightarrow$ CNN & RL & - &Gigaword, CNN/DM, LCSTS & ROUGE \\\cline{2-8}
         & \citeauthor{kryscinski2018improving}~\cite{kryscinski2018improving}  & Improve Abstraction & LSTM Encoder $\rightarrow$ Decoder: Contextual Model and Language Model & XENT + RL & Asynchronous Gradient Descent Optimizer & CNN/DM &  ROUGE, Novel n-gram Test, Human \\\cline{2-8}
         & Gehrmann~\etal~\cite{gehrmann2018bottom} & Bottom-up Attention, Abstraction + Extraction & Pointer-Generator Network & Hybrid-Objective & Adagrad & CNN/DM, NYT & ROUGE, $\%$Novel \\\cline{2-8}
         & Zhang~\etal~\cite{zhang2018learning} & Learning to Summarize Radiology Findings & Pointer-Generator Network + Background Encoder & XENT & Adam & Radiology Reports & ROUGE \\\cline{2-8}
         & Jiang~\etal~\cite{jiang2018closed} & Closed-book Training & Pointer-Generator Network + Closed-book Decoder & Hybrid-Objective + RL & Adam & DUC, CNN/DM &  ROUGE, METEOR \\\cline{2-8}
         & Chung~\etal~\cite{chung2018main} & Main Pointer Generator & Pointer-Generator Network + Document Encoder & XENT & Adadelta & CNN/DM & ROUGE \\\cline{2-8}
         & Chen~\etal~\cite{chen2018iterative} & Iterative Text Summarization & GRU encoder, GRU decoder, iterative unit & Hybrid-Objective & Adam & DUC, CNN/DM & ROUGE
         \\\hline
	\end{tabular}}
	\vspace{-5mm}
\end{table}

\vspace{-1mm}
\subsection{Extraction + Abstraction}

Extractive summarization approaches usually show a better performance comparing to the abstractive approaches~\cite{see2017get,nallapati2016abstractive,zhou2018neural} especially with respect to ROUGE measures.
One of the advantages of the extractive approaches is that they can summarize source articles by extracting salient snippets and sentences directly from these documents~\cite{nallapati2017summarunner}, while abstractive approaches rely on word-level attention mechanism to determine the most relevant words to the target words at each decoding step.
In this section, we review several studies that have attempted to improve the performance of the abstractive summarization by combining them with extractive models.

\vspace{-1mm}
\subsubsection{Extractor + Pointer-Generator Network~\cite{hsu2018unified}}
This model proposes a unified framework that tries to leverage the sentence-level salient information from an extractive model and 
incorporate them into an abstractive model (a pointer-generator network).
More formally, inspired by the hierarchical attention mechanism~\cite{nallapati2016abstractive}, they replaced the attention distribution $\alpha^e_t$ in the abstractive model with a scaled version $\alpha^{\text{scale}}_t$, where the attention weights are expressed by
$\alpha^{\text{scale}}_{tj}=\frac{\alpha^{\text{extra}}_{tj}\alpha^{\text{wd}}_{tj}}{\sum_{k}\alpha^{\text{extra}}_{tk}\alpha^{\text{wd}}_{tk}}$.
Here, $\alpha^\text{extra}_{tj}$ is the sentence-level salient score of the sentence at word position $j$ and decoding step $t$.
Different from~\cite{nallapati2016abstractive}, the salient scores (sentence-level attention weights) are obtained from another deep neural network known as extractor~\cite{hsu2018unified}.

During training, in addition to cross-entropy and coverage loss used in the pointer-generator network, this paper also proposed two other losses, i.e., \textit{extractor loss} and \textit{inconsistency loss}.
The \textit{extractor loss} is used to train the extractor and is defined by 
$L_\text{ext}=-\frac{1}{N}\sum_{n=1}^N g_n\log\beta_n+(1-g_n)\log(1-\beta_n)$,
where $g_n$ is the ground truth label for the $n^\text{th}$ sentence and $N$ is the total number of sentences.
The \textit{inconsistency loss} is expressed as
$L_\text{inc}=-\frac{1}{T}\sum_{t=1}^T\log(\frac{1}{|\mathcal{K}|}\sum_{j\in\mathcal{K}}\alpha^e_{tj}\alpha^\text{extra}_{tj})$,
where $\mathcal{K}$ is the set of the top-$k$ attended words and $T$ is the total number of words in a summary.
Intuitively, the inconsistency loss is used to ensure that the sentence-level attentions in the extractive model and word-level attentions in the abstractive model are consistent with each other. 
In other words, when word-level attention weights are high, the corresponding sentence-level attention weights should also be high.

\vspace{-1mm}
\subsubsection{Key-Information Guide Network (KIGN)~\cite{li2018guiding}}

This approach uses a guiding generation mechanism that leverages the key (salient) information, i.e., keywords, to guide decoding process.
This is a two-step procedure.
First, keywords are extracted from source articles using the TextRank algorithm~\cite{mihalcea2004textrank}.
Second, a KIGN encodes the key information and incorporates them into the decoder to guide the generation of summaries.
Technically speaking, we can use a bi-directional LSTM to encode the key information and the output vector is the concatenation of hidden states, i.e.,
$h^\text{key}=\vec{h^\text{key}_N}\oplus\cev{h^\text{key}_1}$, where $N$ is the length of the key information sequence.
Then, the alignment mechanism is modified as
$s^e_{tj}=(v_\text{align})^\top\tanh(W_\text{align}^eh^e_j+W_\text{align}^d h^d_t+W_\text{align}^\text{key}h^\text{key})$.
Similarly, the soft-switch in the pointer-generator network is calculated using
$p_{\text{gen},t}=\sigma(W_{s,z}z_t^e+W_{s,h}h_t^d+W_{s,\text{key}}h^\text{key}+b_s)$.

\vspace{-1mm}
\subsubsection{Reinforce-Selected Sentence Rewriting~\cite{chen2018fast}}

Most models introduced in this survey are built upon the encoder-decoder framework~\cite{nallapati2016abstractive,see2017get,paulus2017deep}, in which the encoder reads source articles and turns them into vector representations, and the decoder takes the encoded vectors as input and generates summaries.
Unlike these models, the reinforce-selected sentence rewriting model~\cite{chen2018fast} consists of two seq2seq models.
The first one is an extractive model (\textit{extractor}) which is designed to extract salient sentences from a source article, while the second is an abstractive model (\textit{abstractor}) which paraphrases and compresses the extracted sentences into a short summary.
The abstractor network is a standard attention-based seq2seq model with the copying mechanism for handling OOV words.
For the extractor network, an encoder first uses a CNN to encode tokens and obtains representations of sentences, and then it uses an LSTM to encode the sentences and represent a source document.
With the sentence-level representations, the decoder (another LSTM) is designed to recurrently extract salient sentences from the document using the pointing mechanism~\cite{vinyals2015pointer}.
This model has achieved the state-of-the-art performance on CNN/Daily Mail dataset and was demonstrated to be computationally more efficient than the pointer-generator network~\cite{see2017get}.

\vspace{-1mm}
\section{Training Strategies}
\label{sec:training}

In this section, we review different strategies to train the seq2seq models for abstractive text summarization.
As discussed in~\cite{ranzato2015sequence}, there are two categories of training methodologies, i.e., word-level and sequence-level training.
The commonly used teacher forcing algorithm~\cite{bengio2015scheduled,williams1989learning} and cross-entropy training~\cite{bengio2003neural,berger1996maximum} belong to the first category, while different RL-based algorithms~\cite{ranzato2015sequence,rennie2017self,bahdanau2016actor} fall into the second.
We now discuss the basic ideas of different training algorithms and their applications to seq2seq models for the text summarization.
A comprehensive survey of deep RL for seq2seq models can be found in~\cite{keneshloo2018deep}.

\vspace{-1mm}
\subsection{Word-Level Training}

The word-level training for language models represents methodologies that try to optimize predictions of the next token~\cite{ranzato2015sequence}.
For example, in the abstractive text summarization, given a source article $x$, a seq2seq model generates a summary $y$ with the probability $P_\theta(y|x)$, where $\theta$ represents model parameters (e.g.,  weights $W$ and bias $b$).
In a neural language model~\cite{bengio2003neural}, this probability can be expanded to
\begin{equation}
P_\theta(y|x)=\prod_{t=1}^{T}P_\theta(y_t|y_{<t},x),
\label{eqn:language_model}
\end{equation}
where each multiplier $P_\theta(y_t|y_{<t},x)$, known as likelihood, is a conditional probability of the next token $y_t$ given all previous ones denoted by $y_{<t}=(y_1, y_2, ..., y_{t-1})$.
Intuitively, the text generation process can be described as follows.
Starting with a special token `SOS' (start of sequence), the model generates a token $y_t$ at a time $t$ with the probability $P_\theta(y_t|y_{<t},x)=P_{\text{vocab},t}(y_t)$.
This token can be obtained by a sampling method or a greedy search, i.e., $y_t=\arg\max_{y_t} P_{\text{vocab},t}$ (see Fig.~\ref{fig:training}(a)).
The generated token will then be fed into the next decoding step.
The generation is stopped when the model outputs `EOS' (end of sequence) token or when the length reaches a user defined maximum threshold.
In this section, we review different approaches for learning model parameters, i.e., $\theta$.
We will start with the commonly used end-to-end training approach, i.e., cross-entropy training, and then move on to two different methods for avoiding the problem of exposure bias.

\vspace{-1mm}
\subsubsection{Cross-Entropy Training (XENT)~\cite{ranzato2015sequence}}

\begin{figure}[!tp]
	\centering
	\vspace{-4mm}
	\begin{subfigure}[]{0.32\linewidth}
    \includegraphics[width=\linewidth, height=11em]{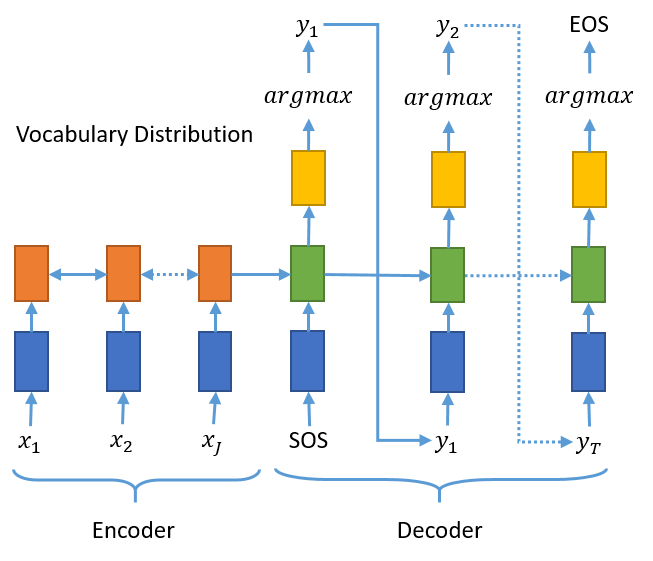}
    \caption{}
    \end{subfigure}
    \begin{subfigure}[]{0.32\linewidth}
    \includegraphics[width=\linewidth, height=11em]{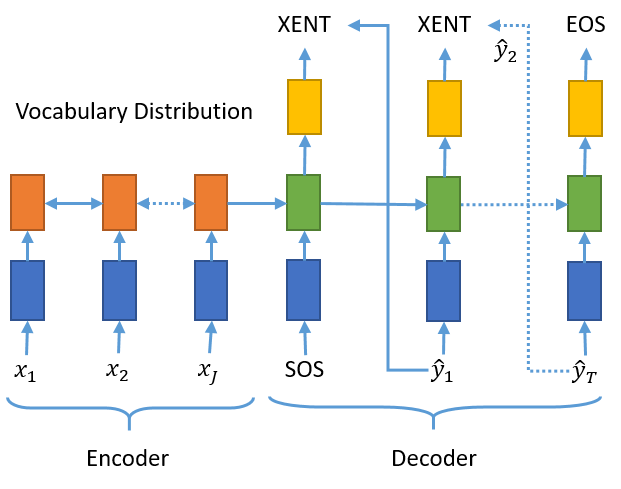}
    \caption{}
    \end{subfigure}
    \begin{subfigure}[]{0.32\linewidth}
    \includegraphics[width=\linewidth, height=11em]{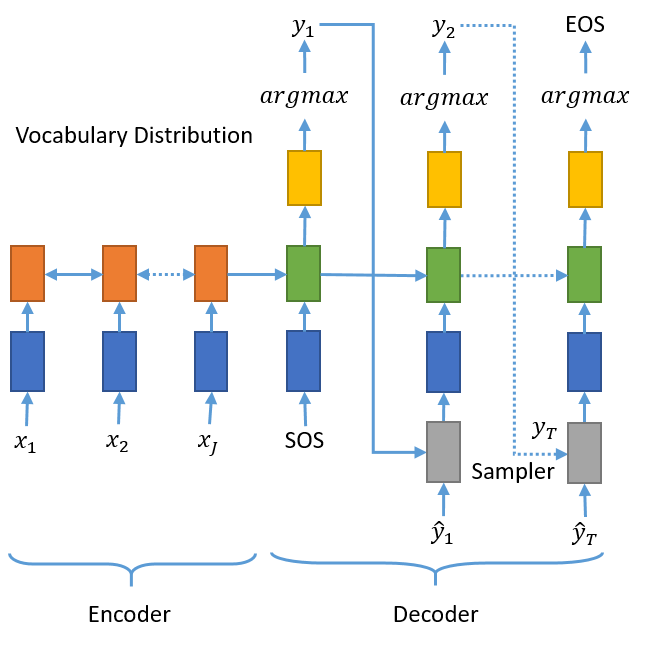}
    \caption{}
    \end{subfigure}
    \vspace{-4mm}
	\caption{(a) Generation process with a greedy search. (b) Training with the teacher forcing algorithm. (c) Illustration of the scheduled sampling.}
	\vspace{-5mm}
	\label{fig:training}
\end{figure}

To learn model parameters $\theta$, XENT maximizes the log-likelihood of observed sequences (ground-truth) $\hat{y}_t=(\hat{y}_1,\hat{y}_2,...,\hat{y}_{T})$, i.e.,
\begin{equation}
\log P_\theta(\hat{y}|x)=\sum_{t=1}^{T}\log P_\theta(\hat{y}_t|\hat{y}_{<t},x)
\end{equation}
which is equivalent to minimizing the cross entropy (XENT) loss, 
$\text{loss}_\text{XENT}=-\log P_\theta(\hat{y}|x)$.
We show this training strategy in Fig.~\ref{fig:training}(b).
The algorithm is also known as the teacher forcing algorithm~\cite{bengio2015scheduled,williams1989learning}. 
During training, it uses observed tokens (ground-truth) as input and aims to improve the probability of the next observed token at each decoding step.
However, during testing, it relies on predicted tokens from the previous decoding step.
This is the major difference between training and testing (see Fig.~\ref{fig:training}(a) and (b)).
Since the predicted tokens may not be the observed ones, this discrepancy will be accumulated over time and thus yields summaries that are very different from ground-truth summaries.
This problem is known as exposure bias~\cite{ranzato2015sequence,venkatraman2015improving,bengio2015scheduled}.

\vspace{-1mm}
\subsubsection{Scheduled Sampling~\cite{ranzato2015sequence,venkatraman2015improving,bengio2015scheduled}}

Scheduled sampling algorithm, also known as Data As Demonstrator (DAD)~\cite{ranzato2015sequence,venkatraman2015improving}, has been proposed to solve the exposure bias problem.
As shown in Fig.~\ref{fig:training}(c), during training, the input at each decoding step comes from a sampler which can decide whether it is a model generated token $y_t$ from the last step or an observed token $\hat{y}_t$ from training data.
The sampling is based on a Bernoulli distribution
\begin{equation}
    P_\text{dad}(y) = p_\text{dad}^{I(y=\hat{y}_t)}\cdot(1-p_\text{dad})^{I(y=y_t)}
\end{equation}
where $p_\text{dad}$ is the probability of using a token from training data and $I(y=y_t)$ is a binary indicator function.
In the scheduled sampling algorithm, $p_\text{dad}$ is an annealing/scheduling function and decreases with training time from $1$ to $0$.
As suggested by Bengio~\etal~\cite{bengio2015scheduled}, scheduling function can take different forms, e.g.,
\begin{equation}
    p_\text{dad}=\begin{cases}
    1-\alpha k& \text{linear decay}\\
    \alpha^{k} & \text{exponential decay}\\
    \frac{\alpha}{\alpha+\exp(k/\alpha)} & \text{inverse sigmoid decay}\\
    \end{cases}
\end{equation}
where $k$ is training step and $\alpha$ is a parameter that guarantees $p_\text{dad}\in[0,1]$.
This strategy is often referred to as a curriculum learning algorithm~\cite{bengio2015scheduled,zhang2017sentence,ranzato2015sequence}.

The main intuition behind this algorithm is that, at the beginning stage, a model with random parameters cannot generates relevant/correct tokens, therefore, a decoder takes ground-truth tokens from training data as input.
As the training proceeds, the model gradually reduces the probability of using ground-truth tokens.
By the end of the training, the model assumes that it has been well trained and can generate reasonable tokens, thus, the decoder can completely rely on its own predictions~\cite{bengio2015scheduled}.

\vspace{-1mm}
\subsubsection{End-To-End Backprop (E2E)~\cite{ranzato2015sequence}}

This algorithm is another method that exposes a model to its own predictions during training.
At each decoding step, it still uses XENT to train the model parameters.
However, the input is neither a ground truth token nor a model generated token.
Instead, it is a fusion of top-$k$ tokens from the last decoding step, where $k$ is a hyper-parameter.
More specifically, the model first samples the top-$k$ tokens, denoted as $(y^{\text{samp}_1}_t,y^{\text{samp}_2}_t, ..., y^{\text{samp}_k}_t)$, from the vocabulary distribution $P_{\text{vocab},t}$.
Then, it can re-scale their probabilities as follows:
\begin{equation}
    P_{\text{samp},t}(y^{\text{samp}_i}_t)=\frac{P_{\text{vocab},t}(y^{\text{samp}_i}_t)}{\sum_jP_{\text{vocab},t}(y^{\text{samp}_j}_t)}
\end{equation}
and obtain a vector in the embedding space by
$E_{\text{samp},t} = \sum_{i=1}^kP_{\text{samp},t}(y^{\text{samp}_i}_t)E_{y^{\text{samp}_i}_t}$.
This fused vector will be served as the input for the next decoding step.
It should be noted that E2E also makes use of DAD in practice~\cite{ranzato2015sequence}, where a sampler is used to determine whether to take fused vectors or embeddings of ground-truth tokens as input.

\vspace{-2mm}
\subsection{Sequence-Level Training}

The sequence-level training with deep RL algorithms has recently received a lot of popularity in the area of neural abstractive text summarization~\cite{ranzato2015sequence,zhang2017sentence,paulus2017deep,keneshloo2018deep,pasunuru2018multi,li2018actor,keneshloo2019deep} due to its ability to incorporate any user-defined metrics, including non-differentiable ROUGE scores, to train neural networks.
In this section, we review several different policy gradient algorithms that have been used to train abstractive text summarization models.
For actor-critic algorithms and related work, the readers are encouraged to go through these publications~\cite{keneshloo2018deep,li2018actor,chen2018fast}.

In RL setting, generating a sequence of tokens in a summary can be considered as a sequential decision making process, where an encoder-decoder model is viewed as an agent, which first reads a source article $x$ and initializes its internal state (hidden and cell states for LSTM).
At the decoding step $t$, it updates the state and takes an action $y^\pi_t\in\mathcal{V}$ (i.e., picking a token from the vocabulary) according to a policy $\pi=P_\theta(y_t|y_{<t},x)$.
Here, the vocabulary is viewed as an action space.
By the end of the decoding, it will produce a sequence of actions $(y^\pi_1, y^\pi_2,...,y^\pi_{T})$ and observe a reward $R(y^\pi_1, y^\pi_2,...,y^\pi_{T})$, which is usually ROUGE scores~\cite{lin2004rouge} in the context of text summarization.
Then, RL algorithms will be used to update the agent by comparing the action sequence based on current policy with the optimal action sequence (i.e., the ground-truth summary).
In this section, we will start with the commonly used REINFORCE for training seq2seq models.
Then, we will introduce the MIXER algorithm (which can improve the convergence rate and stability of the training) and the self-critic sequence training approach (which shows low variance for the gradient estimator). 

\subsubsection{REINFORCE~\cite{williams1992simple,ranzato2015sequence,zhang2017sentence,ling2017coarse}}
The goal of REINFORCE is to find parameters that maximize the expected rewards.
Therefore, the loss function is defined as negative expected reward, i.e.,
$L_{\theta}=-\mathbb{E}_{(y^\pi_{<T+1})\sim\pi_\theta}R(y^\pi_{<T+1})$,
where $y^\pi_{<T+1}$ represents $y^\pi_1, y^\pi_2,...,y^\pi_{T}$.
The above equation can also be rewritten as
\begin{equation}
    L_{\theta}=-\sum_{y^\pi_{<T+1}\in\mathcal{Y}}\pi_\theta(y^\pi_{<T+1})R(y^\pi_{<T+1})
\end{equation}
where $\mathcal{Y}$ represents a set that contains all possible sequences. 
To optimize policy with respect to model parameters, we take the derivative of the loss function and obtain
\begin{equation}
\nabla_\theta L_{\theta}
=-\sum_{y^\pi_{<T+1}\in\mathcal{Y}}\nabla_\theta\pi_\theta(y^\pi_{<T+1})R(y^\pi_{<T+1})
=-\sum_{y^\pi_{<T+1}\in\mathcal{Y}}\pi_\theta(y^\pi_{<T+1})\nabla_\theta\log\pi_\theta(y^\pi_{<T+1})R(y^\pi_{<T+1})
\end{equation}
In the abstractive text summarization task, the policy is expressed as $\pi_\theta=P_\theta(y^\pi_{<T+1}|x)$ and according to Eq. (\ref{eqn:language_model}), the above equation can be expressed as follows:
\begin{equation}
\aligned
\nabla_\theta L_{\theta}
&=-\sum_{y^\pi_{<T+1}\in\mathcal{Y}}P_\theta(y^\pi_{<T+1}|x)\cdot\big[\sum_{t=1}^{T}\nabla_\theta\log P_\theta(y^\pi_t|y^\pi_{<t},x)\big]
\cdot R(y^\pi_{<T+1})\\
&=-\mathbb{E}_{y^\pi_1\sim P_\theta(y^\pi_1|x)}
\mathbb{E}_{y^\pi_2\sim P_\theta(y^\pi_2|y^\pi_1,x)}\cdots
\mathbb{E}_{y^\pi_{T}\sim P_\theta(y^\pi_{T}|y^\pi_{<T},x)}
\big[\sum_{t=1}^{T}\nabla_\theta\log P_\theta(y^\pi_t|y^\pi_{<t},x)\big]\cdot R(y^\pi_{<T+1})\\
&=-\mathbb{E}_{y^\pi_{<T+1}\sim P_\theta(y^\pi_{<T+1}|x)}
\big[\sum_{t=1}^{T}\nabla_\theta\log P_\theta(y^\pi_t|y^\pi_{<t},x)\big]
\cdot R(y^\pi_{<T+1})\\
\endaligned
\label{eqn:reinforce_gradient}
\end{equation}
The reward $R(y^\pi_{<T+1})$ will be back-propagated to every node of the computational graph via the above gradient estimator.
With the gradient, the model parameters are updated by
\begin{equation}
    \theta\leftarrow\theta+\alpha\nabla_\theta L_{\theta}
\end{equation}
where $\alpha$ is the learning rate.

As it can be seen from Eq. (\ref{eqn:reinforce_gradient}), computing gradient requires us to sample all sequences, which is not practical due to the presence of $|\mathcal{V}|^{T}$ possible number of sequences.
Instead, REINFORCE approximates the expectation with a single sample, thus, the gradient is expressed as follows:
\begin{equation}
\aligned
\nabla_\theta L_{\theta}
&\approx-\sum_{t=1}^{T}\nabla_\theta\log P_\theta(y^\pi_t|y^\pi_{<t},x)\cdot R(y^\pi_{<T+1})\\
\endaligned
\end{equation}
One of the problems associated with this method is high variance of gradient estimator, because it makes use of only one sample to train the model.
A practical solution to alleviate this problem is introducing a baseline reward~\cite{weaver2001optimal,xu2015show,rennie2017self} denoted by $b$ to the gradient, i.e.,
\begin{equation}
\nabla_\theta L_{\theta}
=-\mathbb{E}_{y^\pi_{<T+1}\sim P_\theta(y^\pi_{<T+1}|x)}
\big[\sum_{t=1}^{T}\nabla_\theta\log P_\theta(y^\pi_t|y^\pi_{<t},x)\big]\cdot(R(y^\pi_{<T+1})-b))
\end{equation}
The baseline $b$ is arbitrary function but should not depend on $y^\pi_{<T+1}$~\cite{rennie2017self,sutton1998reinforcement}.
In this way, it will not change the expectation of the gradient since 
$\mathbb{E}_{y^\pi_{<T+1}\sim P_\theta(y^\pi_{<T+1}|x)}
\big[\sum_{t=1}^{T}\nabla_\theta\log P_\theta(y^\pi_t|y^\pi_{<t},x)\big]\cdot b=0$.
The complete derivations of the above equation can be found in \cite{rennie2017self}.
In practice, the gradient with the baseline is approximated with
\begin{equation}
\nabla_\theta L_{\theta}
\approx-(R(y^\pi_{<T+1})-b)\sum_{t=1}^{T}\nabla_\theta\log P_\theta(y^\pi_t|y^\pi_{<t},x)
\label{eqn:reinforce_baseline}
\end{equation}
Better ways of sampling a sequence and different approaches to calculate the baseline can be found in~\cite{keneshloo2018deep,xu2015show,zaremba2015reinforcement,ranzato2015sequence}.

\vspace{-1mm}
\subsubsection{MIXER~\cite{ranzato2015sequence}}

Training seq2seq models using REINFORCE may suffer from slow convergence and can also fail due to the large action space and poor initialization (which refers to randomly initialize parameters and start with random policy).
To alleviate this problem, 
Ranzato~\etal~\cite{ranzato2015sequence} modified REINFORCE by incorporating the idea of curriculum learning strategy and proposed a MIXER algorithm.
In this algorithm, they first trained a seq2seq model for $N$-epochs to ensure RL starts with a better policy.
Afterwards, in each batch and for each sequence, they used the cross entropy loss for the first $T-\Delta$ steps and REINFORCE for the remaining $\Delta$ steps, where $\Delta$ is an integer number.
Training was continued for another $N$-epochs, where $N$ is also an integer number.
Then, they increased REINFORCE steps to $2\Delta$ and continued training for another $N$-epochs.
This process will repeat until the whole sequence is trained by REINFORCE.
This algorithm has shown a better performance for greedy generation compared to XENT, DAD and E2E in the task of abstractive text summarization.

\vspace{-1mm}
\subsubsection{Self-Critic Sequence Training (SCST)~\cite{paulus2017deep,celikyilmaz2018deep,pasunuru2018multi,rennie2017self}}
The main idea of SCST is to use testing time inference algorithm as the baseline function in REINFORCE.
Suppose the greedy search (see Fig.~\ref{fig:training}(a)) is used to sample actions during testing.
Then, at each training iteration, the model generates two action sequences, in which the first one $y^{\pi,\text{greedy}}_{<T+1}$ is from greedy search while the second one $y^\pi_{<T+1}$ is sampled from a distribution $P_\theta(y^\pi_{<T+1}|x)$.
According to SCST, baseline $b$ is defined as reward $R(y^{\pi,\text{greedy}}_{<T+1})$ to the first sequence.
Therefore, the gradient of the loss function in SCST is expressed as
\begin{equation}
\nabla_\theta L_{\theta}
\approx-(R(y^\pi_{<T+1})-R(y^{\pi,\text{greedy}}_{<T+1}))\sum_{t=1}^{T}\nabla_\theta\log P_\theta(y^\pi_t|y^\pi_{<t},x),
\end{equation}
according to Eq. (\ref{eqn:reinforce_baseline}). 
The SCST has shown low variance and can be effectively optimized with mini-batch SGD compared to REINFORCE~\cite{rennie2017self}.
It has also been demonstrated to be effective in improving the performance of seq2seq models for the task of abstractive text summarization~\cite{paulus2017deep}.
In this work, the authors used the following RL loss to train their model.
\begin{equation}
    L_{\text{RL}}
\approx-(R(y^\pi_{<T+1})-R(y^{\pi,\text{greedy}}_{<T+1}))\sum_{t=1}^{T}\log P_\theta(y^\pi_t|y^\pi_{<t},x)
\end{equation}
Although the model performs better than those trained with XENT in terms of ROUGE scores, human-readability of generated summaries is low.
To alleviate this problem, the authors also defined a mixed loss function of RL and XENT, i.e., $L_{\text{MIXED}}=\gamma L_\text{RL} + (1-\gamma)L_\text{XENT}$, where $\gamma\in(0,1)$ is a hyper-parameter. 
The model trained with the mixed loss can achieve better human-readability and ROUGE scores are still better than those obtained with XENT.
They also used scheduled sampling to reducing exposure bias, in which the scheduling function is a constant ($p_\text{dad}=0.75$).

We have reviewed different RNN encoder-decoder architectures and training strategies in the last two sections.
Now, we are at the position to generate summaries for given source articles.


\vspace{-2mm}
\section{Summary Generation}
\label{sec:decoding}

Generally speaking, the goal of summary generation is to find an optimal sequence $y^*_{<T+1}$ such that
\begin{equation}
y^*_{<T+1}=\arg\max_{y_{<T+1}\in \mathcal{Y}}\log P_\theta(y_{<T+1}|x)
=\arg\max_{y_{<T+1}\in \mathcal{Y}}\sum_{t=1}^{T}\log P_\theta(y_t|y_{<t},x)
\label{eqn:max_log}
\end{equation}
where $\mathcal{Y}$ represents a set that contains all possible sequences (summaries).
However, since it has $|\mathcal{V}|^{T}$ elements, the exact inference is intractable in practice~\cite{rush2015neural}.
Here, $\mathcal{V}$ represents the output vocabulary.
In this section, we review the beam search algorithm and its extensions for approximating the exact inference.

\vspace{-2mm}
\subsection{Greedy and Beam Search}

As shown in Fig.~\ref{fig:training}(a), we can generate a sub-optimal sequence with greedy search, i.e.,
\begin{equation}
    y^*_t=\arg\max_{y_t\in\mathcal{V}}\log P_\theta(y_t|y_{<t},x)
\end{equation}
at each decoding step $t$.
Although greedy search is computationally efficient, human-readability of generated summaries is low.

\begin{algorithm*}[!tp]
	\SetAlgoLined
	\SetKw{Initialize}{Initialize}
	\KwIn{Source article $x$, beam size $B$, summary length $T$, model parameters $\theta$\;}
	\KwOut{$B$-best summaries\;}
	\BlankLine
	\Initialize:\\
    Output sequences $Q^\text{seq}=\text{[SOS]}_{B\times T}$\;
	Accumulated probabilities $Q^\text{prob}=[1.0]_{B\times 1}$\;
	The last decoded tokens $Q^\text{word}=\text{[SOS]}_{B\times 1}$\;
    States (hidden and cell states for LSTM) $Q^\text{states}=[0.0]_{B\times |h^d_t|}$\;
	Context vectors $Q^\text{ctx}=[0.0]_{B\times |z^e_t|}$\;
	\BlankLine
	\BlankLine
	Compute $(h^e_1,h^e_2,...,h^e_{J})$ with encoder\;
	Update $Q^\text{states}$ with encoder states\;
    
    \For{t=$1$, $T$}{
    Initialize candidates $Q^\text{cand,seq}$, $Q^\text{cand,prob}$, $Q^\text{cand,word}$, $Q^\text{cand,states}$, $Q^\text{cand,ctx}$ by repeating $Q^\text{seq}$, $Q^\text{prob}$, $Q^\text{word}$, $Q^\text{states}$ and $Q^\text{ctx}$ $B$ times, respectively\;
    \For{b=$1$, $B$}{
    Compute $P_\theta(y^\text{cand}_{t,b}|y_{<t,b},x)$ using decoder LSTM cell with input $(h^e_1,h^e_2,...,h^e_{J})$, $Q^\text{word}_b$, $Q^\text{states}_b$ and $Q^\text{ctx}_b$\;
    Select the top-$B$ candidate words $y^\text{cand}_{t,b,b'}$, where $b'=1,2,...,B$\;
    Select corresponding probability $P_\theta(y^\text{cand}_{t,b,b'}|y_{<t,b},x)$, hidden states $h^d_{t,b,b'}$, cell states $c^d_{t,b,b'}$ and context vector $z^e_{t,b,b'}$\;
    Update elements of $Q^\text{cand,seq}_{b',b,t}$, $Q^\text{cand,word}_{b',b}$ with $y^\text{cand}_{t,b,b'}$\;
    Update elements of $Q^\text{cand,states}_{b',b}$ with $h^d_{t,b,b'}$ and $c^d_{t,b,b'}$\;
    Update elements of $Q^\text{cand,ctx}_{b',b}$ with $z^e_{t,b,b'}$\;
    Update $Q^\text{cand,prob}_{b',b}$ with Eq.(\ref{eqn:beam_update_prob})\;
    }
    Flatten $Q^\text{cand,prob}$ and choose $B$ best hypotheses\;
    Update $Q^\text{seq}_t$, $Q^\text{prob}$, $Q^\text{word}$, $Q^\text{states}$, $Q^\text{ctx}$ with corresponding candidates.
    }
	\caption{Beam search algorithm for decoding seq2seq models.}
	\label{alg:beam_search}
\end{algorithm*}

Beam search algorithm is a compromise between greedy search and exact inference and has been commonly employed in different language generation tasks~\cite{rush2015neural,luong2015effective,see2017get}.
Beam search is a graph-search algorithm that generates sequences from left to right by retaining only $B$ top scoring (top-$B$) sequence-fragments at each decoding step.
More formally, we denote decoded top-$B$ sequence fragments, also known as hypotheses \cite{rush2015neural}, at time-step $t-1$ as $y_{<t,1},y_{<t,2},...,y_{<t,B}$ and their scores as $S^\text{bm}_{<t,1},S^\text{bm}_{<t,2},...,S^\text{bm}_{<t,B}$.
For each fragment $y_{<t,b}$, we first calculate $P_\theta(y^\text{cand}_{t,b}|y_{<t,b},x)$, which determines $B$ most probable words $y^\text{cand}_{t,b,1},y^\text{cand}_{t,b,2},...,y^\text{cand}_{t,b,B}$ to expand it.
The score for each expanded fragment, i.e., new hypotheses, $y^\text{cand}_{<t+1,b,b'}$ can then be updated with either
\begin{equation}
    S^\text{cand}_{t,b,b'}=S^\text{bm}_{<t,b}\times P_\theta(y^\text{cand}_{t,b,b'}|y_{<t,b},x)
    \label{eqn:beam_update_prob}
\end{equation}
where $S^\text{bm}_{<t,b}$ is initialized with $1$, or
$S^\text{cand}_{t,b,b'}=S^\text{bm}_{<t,b}+\log P_\theta(y^\text{cand}_{t,b,b'}|y_{<t,b},x)$,
where $S^\text{bm}_{<t,b}$ is initialized with $0$.
Here, $b$ and $b'$ are labels of a current hypothesis and a word candidate, respectively.
This yields $B\times B$ expanded fragments, i.e., new hypotheses, in which only the top-$B$ of them along with their scores are retained for the next decoding step.
This procedure will be repeated until `EOS' token is generated.
In Algorithm~\ref{alg:beam_search}, we show pseudo-codes of a beam search algorithm for generating summaries with Seq2Seq models given the beam size of $B$ and batch size of $1$.

\vspace{-2mm}
\subsection{Diversity-Promoting Algorithms}

Despite widespread applications, beam search algorithm suffered from lacking of diversity within a beam~\cite{gimpel2013systematic,li2016simple,vijayakumar2016diverse,krause2017hierarchical}.
In other words, the top-$B$ hypotheses may differ by just a couple tokens at the end of sequences, which not only limits applications of summarization systems but also wastes computational resources~\cite{li2016diversity,vijayakumar2016diverse}.
Therefore, it is important to promote the diversity of generated sequences.
Ippolito~\etal~\cite{ippolito2019comparison} performed an extensive analysis of different post-training decoding algorithms that aim to increase the diversity during decoding.
They have also shown the power of oversampling (i.e., first sampling additional candidates, and then, filtering them to the desired number.) in improving the diversity without sacrificing performance.
In this section, we briefly introduce some studies that aim to increase the diversity of the beam search algorithm for abstractive summarization models.

\vspace{-1mm}
\subsubsection{Maximum Mutual Information (MMI)~\cite{li2016diversity,li2016mutual,li2016simple}}
The MMI based methods were originally proposed for neural conversation models and then applied to other tasks, such as machine translation and summarization~\cite{li2016mutual,li2016simple}.
The basic intuition here is that a desired model should not only take into account the dependency of a target on a source, but also should consider the likelihood of the source for a given target, which is achieved by replacing the log-likelihood of the target, i.e., $\log P_\theta(y|x)$ in Eq.~(\ref{eqn:max_log}), with pairwise mutual information of the source and target, defined by $\log\frac{P_\theta(y,x)}{P_\theta(x)P_\theta(y)}$.
During the training, model parameters are learned by maximizing mutual information.
When generating sequences, the objective is expressed as follows:
\begin{equation}
y^*=\arg\max_{y\in \mathcal{Y}}\log \frac{P_\theta(y,x)}{P_\theta(x)P_\theta(y)}
=\arg\max_{y\in \mathcal{Y}}\left(\log P_\theta(y|x)-\log P_\theta(y)\right)\\
\label{eqn:max_mmi}
\end{equation}
However, it is obvious that calculating $P_\theta(y)$ is intractable.
Thus, several approximation methods have been proposed in the literature to alleviate this problem \cite{li2016diversity,li2016mutual}.
These approximation method can be summarized by the following three steps:
\begin{itemize}[leftmargin=*]
\item Train two seq2seq models, one for $P_{\theta_1}(y|x)$ and the other for $P_{\theta_2}(x|y)$, where $\theta_1$ and $\theta_2$ are the model parameters.

\item Generate a diverse $N$-best list of sequences based on $P_{\theta_1}(y|x)$.
To achieve this goal, the method for calculating the scores for beam search algorithm has been modified as
\begin{equation}
S^\text{beam}_{k,k'}=b_{<t,k}+\log P_\theta(y_{t,k'}|y_{<t,k},x) - \gamma k'
\label{eqn:beam_diverse}
\end{equation}
By adding the last term $\gamma k'$, the model explicitly encourages hypotheses from different parents, i.e., different $k$, which results in more diverse results.
Therefore, parameter $\gamma$ is also known as the diversity rate which indicates the degree of diversity integrated into beam search algorithm~\cite{li2016simple}.

\item Re-rank $N$-best list by linearly combining $P_{\theta_1}(y|x)$ and $P_{\theta_2}(x|y)$.
The ranking score for each candidate in the $N$-best list is defined as follows:
\begin{equation}
S_\text{rank}(y)=\log P_{\theta_1}(y|x)+\lambda\log P_{\theta_2}(x|y)+\beta\Omega(y)
\end{equation}
where $\Omega(y)$ is a task-specific auxiliary term. 
$\lambda$ and $\beta$ are parameters that can be learned using minimum error rate training~\cite{och2003minimum} on the development dataset.
\end{itemize}

\vspace{-1mm}
\subsubsection{Diverse Beam Search (DBS)~\cite{vijayakumar2016diverse,cibils2018diverse}}
DBS is another approach that aims to increase the diversity of standard beam search algorithm.
It first partitions the hypotheses into $G$ groups.
Then, at each decoding step, it sequentially performs a beam search on each group based on a dissimilarity augmented scoring function
\begin{equation}
S^\text{cand}_{t,b,b'}=S^\text{cand}_{<t,b}+\log P_\theta(y^{\text{cand},g}_{t,b,b'}|y^g_{<t,b},x)
+\lambda_g\Delta(y^{\text{cand},g}_{<t+1,b,b'};y^\text{1}_{<t+1},...,y^\text{g-1}_{<t+1})
\end{equation}
where $\lambda_g\geq 0$ is a parameter.
$\Delta(y^{\text{cand},g}_{<t+1,b,b'}; y^\text{1}_{<t+1},...,y^\text{g-1}_{<t+1})$ represents a diversity function, which measures the dissimilarity or distance between candidate $y^{\text{cand},g}_{<t+1,b,b'}$ in group $g$ and sequences in groups from $1$ to $g-1$.
Standard beam search is applied to group 1.
Intuitively, the generated sequences in different groups are very different from each other due to the penalty of diversity functions.
In~\cite{cibils2018diverse}, DBS has been combined with pointer-generator network~\cite{see2017get} to improve the diversity of the model produced summaries.

\vspace{-2mm}
\section{Implementations and Experiments}
\label{sec:codes_exps}

Apart from a comprehensive literature survey and a detailed review of different techniques for network structures, training strategies and summary generations, we have also developed an open-source library, namely, NATS (\url{https://github.com/tshi04/NATS}), based on RNN seq2seq framework for abstractive text summarization \cite{shi2019leafnats}.
In this section, we first introduce the details of our implementations and then systematically experiment with different network elements and hyper-parameters on three public available datasets, i.e., CNN/Daily Mail, Newsroom, and Bytecup.

\vspace{-2mm}
\subsection{Implementations}

The NATS is equipped with following important features:
\begin{itemize}[leftmargin=*]
\item {\bf Attention based seq2seq framework}.
We implemented the attention based seq2seq model shown in Fig.~\ref{fig:attn_pg}(a).
Encoder and decoder can be chosen to be either LSTM or GRU.
The attention scores can be calculated with one of three alignment methods given in Eq.~(\ref{eqn:align}).

\item {\bf Pointer-generator network}.
Based on the attention based seq2seq framework, we implemented pointer-generator network discussed in Section~\ref{sec:pointer-generator}.

\item {\bf Intra-temporal attention mechanism}. 
The temporal attention can work with all three alignment methods.

\item {\bf Intra-decoder attention mechanism}.
The alignment method for intra-decoder attention is the same as that for the attention mechanism.

\item {\bf Coverage mechanism}. To handle the repetition problem, we implemented the coverage mechanism discussed in Section~\ref{sec:coverage}.
If coverage is switched off, coverage loss will be set to 0.

\item {\bf Weight sharing mechanism}.
As discussed in Section~\ref{sec:weight-sharing}, weight sharing mechanism can boost the performance using significantly fewer parameters.

\item {\bf Beam search algorithm}.
We implemented an efficient beam search algorithm that can also handle the case when the batch size $>1$.

\item {\bf Unknown words replacement}.
Similar to~\cite{chen2016distraction}, we implemented a heuristic unknown words replacement technique to boost the performance.
Theoretically, a pointer-generator network may generate OOV words even with the copying mechanism, because $<$unk$>$ is still in the extended vocabulary.
Thus, after the decoding is completed, we manually check $<$unk$>$ in summaries and replace them with words in source articles using attention weights.
This meta-algorithm can be used for any attention-based seq2seq model.
\end{itemize}

\vspace{-2mm}
\subsection{Datasets}

\vspace{-1mm}
\subsubsection{CNN/Daily Mail Dataset}
CNN/Daily Mail dataset (https://github.com/abisee/cnn-dailymail)
consists of more than $300K$ news articles and each of them is paired with several highlights, known as multi-sentence summaries~\cite{nallapati2016abstractive, see2017get}.
We have summarized the basic statistics of the dataset in Table~\ref{tab:data_stat}.
There are primarily two versions of this dataset.
The first version anonymizes name entities~\cite{nallapati2016abstractive}, while the second one keeps the original texts~\cite{see2017get}.
In this paper, we used the second version and obtained processed data from See~\etal~\cite{see2017get} (\url{https://github.com/JafferWilson/Process-Data-of-CNN-DailyMail}).

\vspace{-1mm}
\subsubsection{Newsroom Dataset}
The Cornell Newsroom dataset (\url{https://summari.es/})~\cite{grusky2018newsroom} was recently released and consists of 1.3 million article-summary pairs, out of which 1.2 million of them are publicly available for training and evaluating summarization systems.
We first used newsroom library (\url{https://github.com/clic-lab/newsroom}) to scrape and extract the raw data.
Then, texts were tokenized with SpaCy package.
We developed a data processing tool to tokenize texts and prepare input for NATS.
In this survey, we created two datasets for text summarization and headline generation, respectively.
The basic statistics of them are shown in Table~\ref{tab:data_stat}.

\begin{table}[!tp]
    \vspace{-4mm}
	\caption{Basic statistics of the CNN/Daily Mail dataset.}
	\vspace{-3mm}
	\centering
	\resizebox{0.9\linewidth}{!}{
	\begin{tabular}{|l|c|c|c|c|c|c|c|c|c|}
		\hline
		& \multicolumn{3}{c|}{\bf CNN/Daily Mail} 
		& \multicolumn{3}{c|}{\bf Newsroom}
		& \multicolumn{3}{c|}{\bf Bytecup}
		\\\hline
		& \bf Train & \bf Dev & \bf Test & \bf Train & \bf Dev & \bf Test & \bf Train & \bf Dev & \bf Test
		\\\hline
		$\#$ pairs 
		& 287,227 & 13,368 & 11,490
		& 992,985 & 108,612 & 108,655
		& 892,734 & 111,592 & 111,592
		\\\hline
		Article Length 
		& 751 & 769 & 778
		& 773 & 766 & 765
		& 640 & 640 & 639
		\\\hline
		Headline Length 
		& - & - & -
		& 10 & 10 & 10
		& 12 & 12 & 12
		\\\hline
		Summary Length 
		& 55 & 61 & 58
		& 31 & 31 & 31
		& - & - & -
		\\\hline
	\end{tabular}}
	\vspace{-5mm}
    \label{tab:data_stat}
\end{table}

\vspace{-1mm}
\subsubsection{Bytecup Dataset}
Byte Cup 2018 International Machine Learning Contest (\url{https://www.biendata.com/competition/bytecup2018/}) released a new dataset, which will be referred to as Bytecup dataset in this survey, for the headline generation task.
It consists of 1.3 million pieces of articles, out of which 1.1 million are released for training.
In our experiments, we create training, development and testing sets (0.8/0.1/0.1) based on this training dataset.
Texts are tokenized using Stanford CoreNLP package and prepared with our data processing tool.
The basic statistics of the dataset are shown in Table~\ref{tab:data_stat}.

\vspace{-2mm}
\subsection{Parameter Settings}

In all our experiments, we set the dimension of word embeddings and hidden states (for both encoder and decoder) as 128 and 256, respectively.
During training, the embeddings are learned from scratch.
Adam~\cite{kingma2014adam} with hyper-parameter $\beta_1=0.9$, $\beta_2=0.999$ and $\epsilon=10^{-8}$ is used for stochastic optimization.
Learning rate is fixed to $0.0001$ and mini-batches of size 16 are used. 
Gradient clipping is also used with a maximum gradient norm of 2.0.
For all datasets, the vocabulary consists of $50K$ words and is shared between source and target.
For the CNN/Daily Mail dataset, we truncate source articles to $400$ tokens and limit the length of summaries to $100$ tokens.
For the Newsroom dataset, source articles, summaries and headlines are truncated to $400$, $50$ and $20$, respectively.
For the Bytecup dataset, lengths of source articles and headlines are also limited to $400$ and $20$ tokens, respectively.
During training, we run 35 epochs for the CNN/Daily Mail dataset and 20 epochs for the Newsroom and Bytecup dataset.
During testing, we set the size of a beam to 5.

\vspace{-2mm}
\subsection{ROUGE Evaluations}

Recall-Oriented Understudy for Gisting Evaluation (ROUGE) scores were first introduced in~\cite{lin2004rouge} and have become standard metrics for evaluating abstractive text summarization models.
They determine the quality of summarization by counting the number of overlapping units (i.e., $n$-grams, word sequences and word pairs) between  machine generated and golden-standard (human-written) summaries~\cite{lin2004rouge}.
Within all different ROUGE measures, ROUGE-1 (unigram), ROUGE-2 (bigram) and ROUGE-L (longest common subsequence) have been most widely used for single-document abstractive summarization~\cite{see2017get}.
In this paper, different models are evaluated using pyrouge (\url{https://pypi.python.org/pypi/pyrouge/0.1.0}) package, which provides precision, recall and F-score for these measures.

\vspace{-2mm}
\subsection{Experiments on CNN/Daily Mail dataset}

In the past few years, the CNN/Daily Mail dataset has become a standard benchmark dataset used for evaluating the performance of different summarization models that can generate multi-sentence summaries for relatively longer documents~\cite{see2017get,nallapati2017summarunner,zhang2018neural,nallapati2016abstractive,paulus2017deep}.
In our experiments, we systematically investigated the effects of six network components in the seq2seq framework on summarization performance, including (i) alignment methods in the attention mechanism, (ii) pointing mechanism, (iii) intra-temporal attention, (iv) intra-decoder attention, (v) weight sharing and (vi) coverage mechanism.

Our experimental results are shown in Table~\ref{tab:cnndm-exp}.
To effectively represent different models, ID of each model consists of a letter followed by five binary-indicators, corresponding to the six important components.
The letters `G', `D' and `C' denote alignment methods `general', `dot' and `concat', respectively.
$1$ and $0$ indicates if a component is switched on or off, respectively.
At first, it can be clearly seen that the performance of three basic attention based models (i.e., G00000, D00000, C00000) 
are close to each other.
In these tests, we still keep the OOV tokens in generated summaries when performing the ROUGE evaluations, which results in relatively lower ROUGE precision scores.
Therefore, ROUGE F-scores may be lower than those reported in the literature~\cite{nallapati2016abstractive,see2017get}.
Comparing G10000 with G00000, and C10000 with C00000, we find that pointing mechanism significantly improves the performance of attention based seq2seq models.
By analyzing summaries, we observed that most of the tokens are copied from source articles, which results in summaries that are similar to the ones generated by the extractive models\footnote{The extractive models attempt to extract sentences from the source articles.}.
As discussed in Section~\ref{sec:pointer-generator}, another advantage of pointing mechanism is that it can effectively handle OOV tokens.

\begin{table*}[!tp]
    \vspace{-4mm}
	\caption{ROUGE scores on the CNN/Daily Mail dataset in our experiments.}
	\vspace{-3mm}
	\centering
	\label{tab:cnndm-exp}
    \resizebox{\linewidth}{!}{
	\begin{tabular}{|c|l|>{\centering\arraybackslash}p{4.5em}|>{\centering\arraybackslash}p{4.5em}|>{\centering\arraybackslash}p{4.5em}|>{\centering\arraybackslash}p{4.5em}|>{\centering\arraybackslash}p{4.5em}|c|c|c|}
		\hline
		\bf Model ID & \bf Attention & \bf Pointer-generator & \bf Intra-temporal & \bf Intra-decoder & \bf Weight sharing & \bf Coverage & \bf R-1 & \bf R-2 & \bf R-L\\\hline
        G00000 & general & - & - & - & - & - & 27.62 & 10.15 & 25.81\\\hline
        G10000 & general & \checkmark & - & - & - & - & 33.85 & 14.08 & 31.33\\\hline
        G11000 & general & \checkmark & \checkmark & - & - & - & 36.78 & 15.82 & 34.05 \\\hline
        G11010 & general & \checkmark & \checkmark & - & \checkmark & - & 37.07 & 15.93 & 34.22 \\\hline
        G11100 & general & \checkmark & \checkmark & \checkmark & - & - & 36.74 & 15.78 & 33.98 \\\hline
        G11110 & general & \checkmark & \checkmark & \checkmark & \checkmark & - & \bf 37.60 & \bf 16.27 & \bf 34.77 \\\hline
        D00000 & dot & - & - & - & - & - & 27.60 & 10.19 & 25.81\\\hline
        D11100 & dot & \checkmark & \checkmark & \checkmark & - & - & 37.00 & 15.87 & 34.24\\\hline
        D11110 & dot & \checkmark & \checkmark & \checkmark &  \checkmark & - & \bf 37.65 & \bf 16.25 & \bf 34.85\\\hline
        C00000 & concat & - & - & - & - & - & 27.61 & 10.16 & 25.78 \\\hline
        C10000 & concat & \checkmark & - & - & - & - & 35.55 & 15.18 & 32.84 \\\hline
        C10001 & concat & \checkmark & - & - & - & \checkmark & 38.64 & 16.70 & 35.63 \\\hline
        C10100 & concat & \checkmark & - & \checkmark & - & - & 36.51 & 15.75 & 33.70 \\\hline
        C10101 & concat & \checkmark & - & \checkmark & - & \checkmark & \bf 39.23 & \bf 17.28 & 36.02 \\\hline
        C10110 & concat & \checkmark & - & \checkmark & \checkmark & - & 36.46 & 15.68 & 33.69 \\\hline
        C10111 & concat & \checkmark & - & \checkmark & \checkmark & \checkmark & 39.14 & 17.13 & \bf 36.04 \\\hline
	\end{tabular}}
	\vspace{-5mm}
\end{table*}

The remaining four components are tested upon the pointer-generator network.
By comparing G11000 and G10000, we see that the intra-temporal attention increases almost 3 ROUGE points. 
This might be because of its capability of reducing repetitions.
However, most of our models that combine intra-temporal attention with `concat' failed during training after a few epochs.
Thus, we did not report these results.
As to intra-decoder attention, we observe from G11000, G11010, G11100 and G11110 that it does not boost the performance of the model before adding weight sharing mechanism.
However, in the case of `concat', the models with intra-decoder attention have a better performance.
Weight sharing mechanism does not always boost the performance of the models (according to the comparison of C10100 and C10110).
However, as aforementioned, models that adopt weight sharing mechanism have much fewer parameters.
Finally, we find the coverage mechanism can significantly boost performance by at least 2 ROUGE points, which is consistent with the results presented in~\cite{see2017get}.
It should be noted that the coverage mechanism can only work with the `concat' attention mechanism according to Section~\ref{sec:coverage}.

\vspace{-2mm}
\subsection{Experiments on Newsroom and Bytecup Datasets}

We also tested NATS toolkit on the Newsroom dataset, which was released recently.
In our experiments, we tokenized the raw data with three different packages and generated three versions of the dataset for the task of text summarization and three versions for the task of headline generation.
Experimental results obtained with the models G11110 and C10110 on the released testing set~\cite{grusky2018newsroom} are shown in Table~\ref{tab:newsroom_byte}.
It can be observed that G11110 performs better than C10110 on CNN/Daily Mail data from Table~\ref{tab:cnndm-exp}, however, C10110 achieves better ROUGE scores in both text summarization and headline generation tasks on Newsroom dataset.
Finally, we summarize our results for the Bytecup headline generation dataset in Table~\ref{tab:newsroom_byte}.
C10110 achieves slightly better scores than G11110.

\begin{table}[!tp]
    \vspace{-4mm}
	\caption{ROUGE scores on Newsroom and Bytecup datasets.}
	\vspace{-3mm}
	\centering
	\label{tab:newsroom_summary}
	\resizebox{0.7\linewidth}{!}{
	\begin{tabular}{|c|c|c|c|c|c|c|c|c|c|}
		\hline
		& \multicolumn{3}{c|}{\bf Newsroom-Summary}
		& \multicolumn{3}{c|}{\bf Newsroom-Title}
		& \multicolumn{3}{c|}{\bf Bytecup}
		\\\hline
	    \bf Model & \bf R-1 & \bf R-2 & \bf R-L & \bf R-1 & \bf R-2 & \bf R-L & \bf R-1 & \bf R-2 & \bf R-L 
	    \\\hline
        G11110 
        & 39.11 & 27.54 & 35.99 
        & 25.23 & 11.46 & 23.72
        & 39.04 & 22.72 & 35.90
        \\\hline
        C10110 
        & 39.36 & 27.86 & 36.35 
        & 26.56 & 12.31 & 25.04
        & 39.13 & 22.98 & 36.14
        \\\hline
	\end{tabular}}
	\vspace{-5mm}
	\label{tab:newsroom_byte}
\end{table}

\vspace{-1mm}
\section{Conclusion and Future Directions}
\label{sec:conclusion}

Being one of the most successful applications of seq2seq models, neural abstractive text summarization has become a prominent research topic that has gained a lot of attention from both industry and academia.
In this paper, we provided a comprehensive survey on the recent advances of seq2seq models for the task of abstractive text summarization.
This work primarily focuses on the challenges associated with neural network architectures, model parameter inference mechanisms and summary generation procedures, and the solutions of different models and algorithms.
We also provided a taxonomy of these topics and an overview of different seq2seq models for the abstractive text summarization.
As part of this survey, we developed an open source toolkit, namely, NATS, which is equipped with several important features, including attention, pointing mechanism, repetition handling, and beam search.
In our experiments, we first summarized the experimental results of different seq2seq models in the literature on the widely used CNN/Daily Mail dataset. We also conducted extensive experiments on this dataset using NATS to examine the effectiveness of different neural network components.
Finally, we established benchmarks for two recently released datasets, i.e., Newsroom and Bytecup.

Despite advances of Seq2Seq models in the abstractive text summarization task, there are still many research challenges that are worth pursing in the future.
(1) \textit{Large Transformers}. 
These large-scale models are first pre-trained on massive text corpora with self-supervised objectives and then fine-tuned on downstream tasks.
They have achieved state-of-the-art performance on a variety of summarization benchmark datasets \cite{dong2019unified,zhang2019pegasus,lewis2019bart,raffel2019exploring,yan2020prophetnet,liu2019text}.
In \cite{zhang2019pegasus}, the pre-trained encoder-decoder model can outperform previous state-of-the-art results on several datasets by fine-tuning with limited supervised examples, which shows that pre-trained models are promising candidates for zero-shot and low-resource summarization.
(2) \textit{Reinforcement Learning (RL)}.
RL-based training strategies can incorporate any user-defined metrics, including non-differentiable ones, as rewards to train summarization models \cite{paulus2017deep,huang2020knowledge,pasunuru2018multi,keneshloo2018deep}.
These metrics can be ROUGE \cite{lin2004rouge}, BERTScore \cite{zhang2019bertscore}, or saliency and entailment rewards \cite{pasunuru2018multi} inferred from the Natural Language Inference task \cite{bowman2015large}.
Therefore, we may improve current models with RL by leveraging external resources and characteristics of different datasets.
(3) \textit{Summary Generation}. 
Most Seq2Seq-based summarization models rely on beam-search algorithm to generate summaries.
Recently, sampling-based approaches have achieved success in open-ended language generation \cite{holtzman2019curious} since they can increase the diversity of the generated texts. It is a promising research direction that can potentially increase the novelty of the generated summaries without sacrificing their quality.
(4) \textit{Datasets}.
Seq2Seq-based summarization models are widely trained and evaluated on News corpora \cite{hermann2015teaching,nallapati2016abstractive,grusky2018newsroom,narayan2018don}.
However, the journalistic writing style promotes the leading paragraphs of most news articles as summaries \cite{koupaee2018wikihow}, which causes the models to favor extraction rather than abstraction \cite{gehrmann2018bottom,grusky2018newsroom}.
To alleviate this problem, researchers have introduced several datasets from other domains 
\cite{cohan2018discourse,sharma2019bigpatent,koupaee2018wikihow,kim2019abstractive}.
In the future, many new datasets will likely be released to build better abstractive summarization systems.
(5) \textit{Evaluation}. Most automatic evaluation protocols, e.g., ROUGE and BERTScore \cite{zhang2019bertscore}, are not sufficient to evaluate the overall quality of generated summaries \cite{maynez2020faithfulness,kryscinski2019neural}.
We still have to access some critical features, like \textit{factual correctness} \cite{kryscinski2019neural}, \textit{fluency}, and \textit{relevance} \cite{chen2018fast}, of generated summaries by human experts.
Thus, a future research direction along this line is building better evaluation systems that go beyond current metrics to capture the most important features which agree with humans.
For example, some attempts have been made for generic text generation
\cite{sellam2020bleurt,zhao2019moverscore}.

\vspace{-1mm}
\begin{acks}
This work was supported in part by the US National Science Foundation grants IIS-1619028, IIS-1707498 and IIS-1838730.
\end{acks}

\bibliographystyle{ACM-Reference-Format}
\bibliography{ref}


\begin{thebibliography}{160}


\ifx \showCODEN    \undefined \def \showCODEN     #1{\unskip}     \fi
\ifx \showDOI      \undefined \def \showDOI       #1{#1}\fi
\ifx \showISBNx    \undefined \def \showISBNx     #1{\unskip}     \fi
\ifx \showISBNxiii \undefined \def \showISBNxiii  #1{\unskip}     \fi
\ifx \showISSN     \undefined \def \showISSN      #1{\unskip}     \fi
\ifx \showLCCN     \undefined \def \showLCCN      #1{\unskip}     \fi
\ifx \shownote     \undefined \def \shownote      #1{#1}          \fi
\ifx \showarticletitle \undefined \def \showarticletitle #1{#1}   \fi
\ifx \showURL      \undefined \def \showURL       {\relax}        \fi
\providecommand\bibfield[2]{#2}
\providecommand\bibinfo[2]{#2}
\providecommand\natexlab[1]{#1}
\providecommand\showeprint[2][]{arXiv:#2}

\bibitem[\protect\citeauthoryear{Allahyari, Pouriyeh, Assefi, Safaei, Trippe,
  Gutierrez, and Kochut}{Allahyari et~al\mbox{.}}{2017}]%
        {allahyari2017text}
\bibfield{author}{\bibinfo{person}{Mehdi Allahyari}, \bibinfo{person}{Seyedamin
  Pouriyeh}, \bibinfo{person}{Mehdi Assefi}, \bibinfo{person}{Saeid Safaei},
  \bibinfo{person}{Elizabeth~D Trippe}, \bibinfo{person}{Juan~B Gutierrez},
  {and} \bibinfo{person}{Krys Kochut}.} \bibinfo{year}{2017}\natexlab{}.
\newblock \showarticletitle{Text summarization techniques: A brief survey}.
\newblock \bibinfo{journal}{{\em arXiv preprint arXiv:1707.02268\/}}
  (\bibinfo{year}{2017}).
\newblock


\bibitem[\protect\citeauthoryear{Bahdanau, Brakel, Xu, Goyal, Lowe, Pineau,
  Courville, and Bengio}{Bahdanau et~al\mbox{.}}{2016a}]%
        {bahdanau2016actor}
\bibfield{author}{\bibinfo{person}{Dzmitry Bahdanau}, \bibinfo{person}{Philemon
  Brakel}, \bibinfo{person}{Kelvin Xu}, \bibinfo{person}{Anirudh Goyal},
  \bibinfo{person}{Ryan Lowe}, \bibinfo{person}{Joelle Pineau},
  \bibinfo{person}{Aaron Courville}, {and} \bibinfo{person}{Yoshua Bengio}.}
  \bibinfo{year}{2016}\natexlab{a}.
\newblock \showarticletitle{An actor-critic algorithm for sequence prediction}.
\newblock \bibinfo{journal}{{\em arXiv preprint arXiv:1607.07086\/}}
  (\bibinfo{year}{2016}).
\newblock


\bibitem[\protect\citeauthoryear{Bahdanau, Cho, and Bengio}{Bahdanau
  et~al\mbox{.}}{2014}]%
        {bahdanau2014neural}
\bibfield{author}{\bibinfo{person}{Dzmitry Bahdanau},
  \bibinfo{person}{Kyunghyun Cho}, {and} \bibinfo{person}{Yoshua Bengio}.}
  \bibinfo{year}{2014}\natexlab{}.
\newblock \showarticletitle{Neural machine translation by jointly learning to
  align and translate}.
\newblock \bibinfo{journal}{{\em arXiv preprint arXiv:1409.0473\/}}
  (\bibinfo{year}{2014}).
\newblock


\bibitem[\protect\citeauthoryear{Bahdanau, Chorowski, Serdyuk, Brakel, and
  Bengio}{Bahdanau et~al\mbox{.}}{2016b}]%
        {bahdanau2016end}
\bibfield{author}{\bibinfo{person}{Dzmitry Bahdanau}, \bibinfo{person}{Jan
  Chorowski}, \bibinfo{person}{Dmitriy Serdyuk}, \bibinfo{person}{Philemon
  Brakel}, {and} \bibinfo{person}{Yoshua Bengio}.}
  \bibinfo{year}{2016}\natexlab{b}.
\newblock \showarticletitle{End-to-end attention-based large vocabulary speech
  recognition}. In \bibinfo{booktitle}{{\em Acoustics, Speech and Signal
  Processing (ICASSP), 2016 IEEE International Conference on}}. IEEE,
  \bibinfo{pages}{4945--4949}.
\newblock


\bibitem[\protect\citeauthoryear{Bahl, Brown, De~Souza, and Mercer}{Bahl
  et~al\mbox{.}}{1986}]%
        {bahl1986maximum}
\bibfield{author}{\bibinfo{person}{Lalit Bahl}, \bibinfo{person}{Peter Brown},
  \bibinfo{person}{Peter De~Souza}, {and} \bibinfo{person}{Robert Mercer}.}
  \bibinfo{year}{1986}\natexlab{}.
\newblock \showarticletitle{Maximum mutual information estimation of hidden
  Markov model parameters for speech recognition}. In \bibinfo{booktitle}{{\em
  Acoustics, Speech, and Signal Processing, IEEE International Conference on
  ICASSP'86.}}, Vol.~\bibinfo{volume}{11}. IEEE, \bibinfo{pages}{49--52}.
\newblock


\bibitem[\protect\citeauthoryear{Bahuleyan, Mou, Vechtomova, and
  Poupart}{Bahuleyan et~al\mbox{.}}{2018}]%
        {bahuleyan2017variational}
\bibfield{author}{\bibinfo{person}{Hareesh Bahuleyan}, \bibinfo{person}{Lili
  Mou}, \bibinfo{person}{Olga Vechtomova}, {and} \bibinfo{person}{Pascal
  Poupart}.} \bibinfo{year}{2018}\natexlab{}.
\newblock \showarticletitle{Variational Attention for Sequence-to-Sequence
  Models}. In \bibinfo{booktitle}{{\em COLING}}.
\newblock


\bibitem[\protect\citeauthoryear{Balduzzi and Ghifary}{Balduzzi and
  Ghifary}{2016}]%
        {balduzzi2016strongly}
\bibfield{author}{\bibinfo{person}{David Balduzzi} {and}
  \bibinfo{person}{Muhammad Ghifary}.} \bibinfo{year}{2016}\natexlab{}.
\newblock \showarticletitle{Strongly-typed recurrent neural networks}. In
  \bibinfo{booktitle}{{\em Proceedings of the 33rd International Conference on
  International Conference on Machine Learning-Volume 48}}. JMLR. org,
  \bibinfo{pages}{1292--1300}.
\newblock


\bibitem[\protect\citeauthoryear{Bengio, Vinyals, Jaitly, and Shazeer}{Bengio
  et~al\mbox{.}}{2015}]%
        {bengio2015scheduled}
\bibfield{author}{\bibinfo{person}{Samy Bengio}, \bibinfo{person}{Oriol
  Vinyals}, \bibinfo{person}{Navdeep Jaitly}, {and} \bibinfo{person}{Noam
  Shazeer}.} \bibinfo{year}{2015}\natexlab{}.
\newblock \showarticletitle{Scheduled sampling for sequence prediction with
  recurrent neural networks}. In \bibinfo{booktitle}{{\em Advances in Neural
  Information Processing Systems}}. \bibinfo{pages}{1171--1179}.
\newblock


\bibitem[\protect\citeauthoryear{Bengio, Ducharme, Vincent, and Jauvin}{Bengio
  et~al\mbox{.}}{2003}]%
        {bengio2003neural}
\bibfield{author}{\bibinfo{person}{Yoshua Bengio}, \bibinfo{person}{R{\'e}jean
  Ducharme}, \bibinfo{person}{Pascal Vincent}, {and} \bibinfo{person}{Christian
  Jauvin}.} \bibinfo{year}{2003}\natexlab{}.
\newblock \showarticletitle{A neural probabilistic language model}.
\newblock \bibinfo{journal}{{\em Journal of machine learning research\/}}
  \bibinfo{volume}{3}, \bibinfo{number}{Feb} (\bibinfo{year}{2003}),
  \bibinfo{pages}{1137--1155}.
\newblock


\bibitem[\protect\citeauthoryear{Bengio, Simard, and Frasconi}{Bengio
  et~al\mbox{.}}{1994}]%
        {bengio1994learning}
\bibfield{author}{\bibinfo{person}{Yoshua Bengio}, \bibinfo{person}{Patrice
  Simard}, {and} \bibinfo{person}{Paolo Frasconi}.}
  \bibinfo{year}{1994}\natexlab{}.
\newblock \showarticletitle{Learning long-term dependencies with gradient
  descent is difficult}.
\newblock \bibinfo{journal}{{\em IEEE transactions on neural networks\/}}
  \bibinfo{volume}{5}, \bibinfo{number}{2} (\bibinfo{year}{1994}),
  \bibinfo{pages}{157--166}.
\newblock


\bibitem[\protect\citeauthoryear{Berger, Pietra, and Pietra}{Berger
  et~al\mbox{.}}{1996}]%
        {berger1996maximum}
\bibfield{author}{\bibinfo{person}{Adam~L Berger}, \bibinfo{person}{Vincent
  J~Della Pietra}, {and} \bibinfo{person}{Stephen A~Della Pietra}.}
  \bibinfo{year}{1996}\natexlab{}.
\newblock \showarticletitle{A maximum entropy approach to natural language
  processing}.
\newblock \bibinfo{journal}{{\em Computational linguistics\/}}
  \bibinfo{volume}{22}, \bibinfo{number}{1} (\bibinfo{year}{1996}),
  \bibinfo{pages}{39--71}.
\newblock


\bibitem[\protect\citeauthoryear{Bhatia and Jaiswal}{Bhatia and
  Jaiswal}{2016}]%
        {bhatia2016automatic}
\bibfield{author}{\bibinfo{person}{Neelima Bhatia} {and}
  \bibinfo{person}{Arunima Jaiswal}.} \bibinfo{year}{2016}\natexlab{}.
\newblock \showarticletitle{Automatic text summarization and it's methods-a
  review}. In \bibinfo{booktitle}{{\em Cloud System and Big Data Engineering
  (Confluence), 2016 6th International Conference}}. IEEE,
  \bibinfo{pages}{65--72}.
\newblock


\bibitem[\protect\citeauthoryear{Bowman, Angeli, Potts, and Manning}{Bowman
  et~al\mbox{.}}{2015}]%
        {bowman2015large}
\bibfield{author}{\bibinfo{person}{Samuel Bowman}, \bibinfo{person}{Gabor
  Angeli}, \bibinfo{person}{Christopher Potts}, {and}
  \bibinfo{person}{Christopher~D Manning}.} \bibinfo{year}{2015}\natexlab{}.
\newblock \showarticletitle{A large annotated corpus for learning natural
  language inference}. In \bibinfo{booktitle}{{\em Proceedings of the 2015
  Conference on Empirical Methods in Natural Language Processing}}.
  \bibinfo{pages}{632--642}.
\newblock


\bibitem[\protect\citeauthoryear{Bradbury, Merity, Xiong, and Socher}{Bradbury
  et~al\mbox{.}}{2016}]%
        {bradbury2016quasi}
\bibfield{author}{\bibinfo{person}{James Bradbury}, \bibinfo{person}{Stephen
  Merity}, \bibinfo{person}{Caiming Xiong}, {and} \bibinfo{person}{Richard
  Socher}.} \bibinfo{year}{2016}\natexlab{}.
\newblock \showarticletitle{Quasi-recurrent neural networks}.
\newblock \bibinfo{journal}{{\em arXiv preprint arXiv:1611.01576\/}}
  (\bibinfo{year}{2016}).
\newblock


\bibitem[\protect\citeauthoryear{Celikyilmaz, Bosselut, He, and
  Choi}{Celikyilmaz et~al\mbox{.}}{2018}]%
        {celikyilmaz2018deep}
\bibfield{author}{\bibinfo{person}{Asli Celikyilmaz}, \bibinfo{person}{Antoine
  Bosselut}, \bibinfo{person}{Xiaodong He}, {and} \bibinfo{person}{Yejin
  Choi}.} \bibinfo{year}{2018}\natexlab{}.
\newblock \showarticletitle{Deep Communicating Agents for Abstractive
  Summarization}. In \bibinfo{booktitle}{{\em Proceedings of the 2018
  Conference of the North American Chapter of the Association for Computational
  Linguistics: Human Language Technologies, Volume 1 (Long Papers)}},
  Vol.~\bibinfo{volume}{1}. \bibinfo{pages}{1662--1675}.
\newblock


\bibitem[\protect\citeauthoryear{Chen, Bolton, and Manning}{Chen
  et~al\mbox{.}}{2016a}]%
        {chen2016thorough}
\bibfield{author}{\bibinfo{person}{Danqi Chen}, \bibinfo{person}{Jason Bolton},
  {and} \bibinfo{person}{Christopher~D Manning}.}
  \bibinfo{year}{2016}\natexlab{a}.
\newblock \showarticletitle{A Thorough Examination of the {CNN/Daily Mail}
  Reading Comprehension Task}. In \bibinfo{booktitle}{{\em Proceedings of the
  54th Annual Meeting of the Association for Computational Linguistics (Volume
  1: Long Papers)}}, Vol.~\bibinfo{volume}{1}. \bibinfo{pages}{2358--2367}.
\newblock


\bibitem[\protect\citeauthoryear{Chen, Zhu, Ling, Wei, and Jiang}{Chen
  et~al\mbox{.}}{2016b}]%
        {chen2016distraction}
\bibfield{author}{\bibinfo{person}{Qian Chen}, \bibinfo{person}{Xiaodan Zhu},
  \bibinfo{person}{Zhenhua Ling}, \bibinfo{person}{Si Wei}, {and}
  \bibinfo{person}{Hui Jiang}.} \bibinfo{year}{2016}\natexlab{b}.
\newblock \showarticletitle{Distraction-based neural networks for modeling
  documents}. In \bibinfo{booktitle}{{\em Proceedings of the Twenty-Fifth
  International Joint Conference on Artificial Intelligence}}. AAAI Press,
  \bibinfo{pages}{2754--2760}.
\newblock


\bibitem[\protect\citeauthoryear{Chen, Gao, Tao, Song, Zhao, and Yan}{Chen
  et~al\mbox{.}}{2018}]%
        {chen2018iterative}
\bibfield{author}{\bibinfo{person}{Xiuying Chen}, \bibinfo{person}{Shen Gao},
  \bibinfo{person}{Chongyang Tao}, \bibinfo{person}{Yan Song},
  \bibinfo{person}{Dongyan Zhao}, {and} \bibinfo{person}{Rui Yan}.}
  \bibinfo{year}{2018}\natexlab{}.
\newblock \showarticletitle{Iterative Document Representation Learning Towards
  Summarization with Polishing}. In \bibinfo{booktitle}{{\em Proceedings of the
  2018 Conference on Empirical Methods in Natural Language Processing}}.
  \bibinfo{pages}{4088--4097}.
\newblock


\bibitem[\protect\citeauthoryear{Chen and Bansal}{Chen and Bansal}{2018}]%
        {chen2018fast}
\bibfield{author}{\bibinfo{person}{Yen-Chun Chen} {and} \bibinfo{person}{Mohit
  Bansal}.} \bibinfo{year}{2018}\natexlab{}.
\newblock \showarticletitle{Fast Abstractive Summarization with
  Reinforce-Selected Sentence Rewriting}. In \bibinfo{booktitle}{{\em
  Proceedings of the 56th Annual Meeting of the Association for Computational
  Linguistics (Volume 1: Long Papers)}}. \bibinfo{publisher}{Association for
  Computational Linguistics}, \bibinfo{pages}{675--686}.
\newblock


\bibitem[\protect\citeauthoryear{Cheng and Lapata}{Cheng and Lapata}{2016}]%
        {cheng2016neural}
\bibfield{author}{\bibinfo{person}{Jianpeng Cheng} {and}
  \bibinfo{person}{Mirella Lapata}.} \bibinfo{year}{2016}\natexlab{}.
\newblock \showarticletitle{Neural Summarization by Extracting Sentences and
  Words}. In \bibinfo{booktitle}{{\em Proceedings of the 54th Annual Meeting of
  the Association for Computational Linguistics (Volume 1: Long Papers)}},
  Vol.~\bibinfo{volume}{1}. \bibinfo{pages}{484--494}.
\newblock


\bibitem[\protect\citeauthoryear{Cho, van Merrienboer, Gulcehre, Bahdanau,
  Bougares, Schwenk, and Bengio}{Cho et~al\mbox{.}}{2014}]%
        {cho2014learning}
\bibfield{author}{\bibinfo{person}{Kyunghyun Cho}, \bibinfo{person}{Bart van
  Merrienboer}, \bibinfo{person}{Caglar Gulcehre}, \bibinfo{person}{Dzmitry
  Bahdanau}, \bibinfo{person}{Fethi Bougares}, \bibinfo{person}{Holger
  Schwenk}, {and} \bibinfo{person}{Yoshua Bengio}.}
  \bibinfo{year}{2014}\natexlab{}.
\newblock \showarticletitle{Learning Phrase Representations using {RNN}
  Encoder--Decoder for Statistical Machine Translation}. In
  \bibinfo{booktitle}{{\em Proceedings of the 2014 Conference on Empirical
  Methods in Natural Language Processing (EMNLP)}}.
  \bibinfo{pages}{1724--1734}.
\newblock


\bibitem[\protect\citeauthoryear{Chopra, Auli, and Rush}{Chopra
  et~al\mbox{.}}{2016}]%
        {chopra2016abstractive}
\bibfield{author}{\bibinfo{person}{Sumit Chopra}, \bibinfo{person}{Michael
  Auli}, {and} \bibinfo{person}{Alexander~M Rush}.}
  \bibinfo{year}{2016}\natexlab{}.
\newblock \showarticletitle{Abstractive sentence summarization with attentive
  recurrent neural networks}. In \bibinfo{booktitle}{{\em Proceedings of the
  2016 Conference of the North American Chapter of the Association for
  Computational Linguistics: Human Language Technologies}}.
  \bibinfo{pages}{93--98}.
\newblock


\bibitem[\protect\citeauthoryear{Chung, Gulcehre, Cho, and Bengio}{Chung
  et~al\mbox{.}}{2014}]%
        {chung2014empirical}
\bibfield{author}{\bibinfo{person}{Junyoung Chung}, \bibinfo{person}{Caglar
  Gulcehre}, \bibinfo{person}{Kyunghyun Cho}, {and} \bibinfo{person}{Yoshua
  Bengio}.} \bibinfo{year}{2014}\natexlab{}.
\newblock \showarticletitle{Empirical evaluation of gated recurrent neural
  networks on sequence modeling}. In \bibinfo{booktitle}{{\em NIPS 2014
  Workshop on Deep Learning, December 2014}}.
\newblock


\bibitem[\protect\citeauthoryear{Chung, Kastner, Dinh, Goel, Courville, and
  Bengio}{Chung et~al\mbox{.}}{2015}]%
        {chung2015recurrent}
\bibfield{author}{\bibinfo{person}{Junyoung Chung}, \bibinfo{person}{Kyle
  Kastner}, \bibinfo{person}{Laurent Dinh}, \bibinfo{person}{Kratarth Goel},
  \bibinfo{person}{Aaron~C Courville}, {and} \bibinfo{person}{Yoshua Bengio}.}
  \bibinfo{year}{2015}\natexlab{}.
\newblock \showarticletitle{A recurrent latent variable model for sequential
  data}. In \bibinfo{booktitle}{{\em Advances in neural information processing
  systems}}. \bibinfo{pages}{2980--2988}.
\newblock


\bibitem[\protect\citeauthoryear{Chung, Xu, Liu, and Ouyang}{Chung
  et~al\mbox{.}}{2018}]%
        {chung2018main}
\bibfield{author}{\bibinfo{person}{Tong~Lee Chung}, \bibinfo{person}{Bin Xu},
  \bibinfo{person}{Yongbin Liu}, {and} \bibinfo{person}{Chunping Ouyang}.}
  \bibinfo{year}{2018}\natexlab{}.
\newblock \showarticletitle{Main Point Generator: Summarizing with a Focus}. In
  \bibinfo{booktitle}{{\em International Conference on Database Systems for
  Advanced Applications}}. Springer, \bibinfo{pages}{924--932}.
\newblock


\bibitem[\protect\citeauthoryear{Cibils, Musat, Hossman, and Baeriswyl}{Cibils
  et~al\mbox{.}}{2018}]%
        {cibils2018diverse}
\bibfield{author}{\bibinfo{person}{Andr{\'e} Cibils}, \bibinfo{person}{Claudiu
  Musat}, \bibinfo{person}{Andreea Hossman}, {and} \bibinfo{person}{Michael
  Baeriswyl}.} \bibinfo{year}{2018}\natexlab{}.
\newblock \showarticletitle{Diverse Beam Search for Increased Novelty in
  Abstractive Summarization}.
\newblock \bibinfo{journal}{{\em arXiv preprint arXiv:1802.01457\/}}
  (\bibinfo{year}{2018}).
\newblock


\bibitem[\protect\citeauthoryear{Clark, Luong, Le, and Manning}{Clark
  et~al\mbox{.}}{2019}]%
        {clark2019electra}
\bibfield{author}{\bibinfo{person}{Kevin Clark}, \bibinfo{person}{Minh-Thang
  Luong}, \bibinfo{person}{Quoc~V Le}, {and} \bibinfo{person}{Christopher~D
  Manning}.} \bibinfo{year}{2019}\natexlab{}.
\newblock \showarticletitle{ELECTRA: Pre-training Text Encoders as
  Discriminators Rather Than Generators}. In \bibinfo{booktitle}{{\em
  International Conference on Learning Representations}}.
\newblock


\bibitem[\protect\citeauthoryear{Cohan, Dernoncourt, Kim, Bui, Kim, Chang, and
  Goharian}{Cohan et~al\mbox{.}}{2018}]%
        {cohan2018discourse}
\bibfield{author}{\bibinfo{person}{Arman Cohan}, \bibinfo{person}{Franck
  Dernoncourt}, \bibinfo{person}{Doo~Soon Kim}, \bibinfo{person}{Trung Bui},
  \bibinfo{person}{Seokhwan Kim}, \bibinfo{person}{Walter Chang}, {and}
  \bibinfo{person}{Nazli Goharian}.} \bibinfo{year}{2018}\natexlab{}.
\newblock \showarticletitle{A Discourse-Aware Attention Model for Abstractive
  Summarization of Long Documents}. In \bibinfo{booktitle}{{\em Proceedings of
  the 2018 Conference of the North American Chapter of the Association for
  Computational Linguistics: Human Language Technologies, Volume 2 (Short
  Papers)}}, Vol.~\bibinfo{volume}{2}. \bibinfo{pages}{615--621}.
\newblock


\bibitem[\protect\citeauthoryear{Dalal and Malik}{Dalal and Malik}{2013}]%
        {dalal2013survey}
\bibfield{author}{\bibinfo{person}{Vipul Dalal} {and} \bibinfo{person}{Latesh~G
  Malik}.} \bibinfo{year}{2013}\natexlab{}.
\newblock \showarticletitle{A survey of extractive and abstractive text
  summarization techniques}. In \bibinfo{booktitle}{{\em Emerging Trends in
  Engineering and Technology (ICETET), 2013 6th International Conference on}}.
  IEEE, \bibinfo{pages}{109--110}.
\newblock


\bibitem[\protect\citeauthoryear{Das and Martins}{Das and Martins}{2007}]%
        {das2007survey}
\bibfield{author}{\bibinfo{person}{Dipanjan Das} {and}
  \bibinfo{person}{Andr{\'e}~FT Martins}.} \bibinfo{year}{2007}\natexlab{}.
\newblock \showarticletitle{A survey on automatic text summarization}.
\newblock \bibinfo{journal}{{\em Literature Survey for the Language and
  Statistics II course at CMU\/}}  \bibinfo{volume}{4} (\bibinfo{year}{2007}),
  \bibinfo{pages}{192--195}.
\newblock


\bibitem[\protect\citeauthoryear{Dauphin, Fan, Auli, and Grangier}{Dauphin
  et~al\mbox{.}}{2017}]%
        {dauphin2016language}
\bibfield{author}{\bibinfo{person}{Yann~N Dauphin}, \bibinfo{person}{Angela
  Fan}, \bibinfo{person}{Michael Auli}, {and} \bibinfo{person}{David
  Grangier}.} \bibinfo{year}{2017}\natexlab{}.
\newblock \showarticletitle{Language Modeling with Gated Convolutional
  Networks}. In \bibinfo{booktitle}{{\em International Conference on Machine
  Learning}}. \bibinfo{pages}{933--941}.
\newblock


\bibitem[\protect\citeauthoryear{Devlin, Chang, Lee, and Toutanova}{Devlin
  et~al\mbox{.}}{2019}]%
        {devlin2019bert}
\bibfield{author}{\bibinfo{person}{Jacob Devlin}, \bibinfo{person}{Ming-Wei
  Chang}, \bibinfo{person}{Kenton Lee}, {and} \bibinfo{person}{Kristina
  Toutanova}.} \bibinfo{year}{2019}\natexlab{}.
\newblock \showarticletitle{BERT: Pre-training of Deep Bidirectional
  Transformers for Language Understanding}. In \bibinfo{booktitle}{{\em
  Proceedings of the 2019 Conference of the North American Chapter of the
  Association for Computational Linguistics: Human Language Technologies,
  Volume 1 (Long and Short Papers)}}. \bibinfo{pages}{4171--4186}.
\newblock


\bibitem[\protect\citeauthoryear{Doersch}{Doersch}{2016}]%
        {doersch2016tutorial}
\bibfield{author}{\bibinfo{person}{Carl Doersch}.}
  \bibinfo{year}{2016}\natexlab{}.
\newblock \showarticletitle{Tutorial on variational autoencoders}.
\newblock \bibinfo{journal}{{\em arXiv preprint arXiv:1606.05908\/}}
  (\bibinfo{year}{2016}).
\newblock


\bibitem[\protect\citeauthoryear{Dong, Yang, Wang, Wei, Liu, Wang, Gao, Zhou,
  and Hon}{Dong et~al\mbox{.}}{2019}]%
        {dong2019unified}
\bibfield{author}{\bibinfo{person}{Li Dong}, \bibinfo{person}{Nan Yang},
  \bibinfo{person}{Wenhui Wang}, \bibinfo{person}{Furu Wei},
  \bibinfo{person}{Xiaodong Liu}, \bibinfo{person}{Yu Wang},
  \bibinfo{person}{Jianfeng Gao}, \bibinfo{person}{Ming Zhou}, {and}
  \bibinfo{person}{Hsiao-Wuen Hon}.} \bibinfo{year}{2019}\natexlab{}.
\newblock \showarticletitle{Unified language model pre-training for natural
  language understanding and generation}. In \bibinfo{booktitle}{{\em Advances
  in Neural Information Processing Systems}}. \bibinfo{pages}{13042--13054}.
\newblock


\bibitem[\protect\citeauthoryear{Elman}{Elman}{1990}]%
        {elman1990finding}
\bibfield{author}{\bibinfo{person}{Jeffrey~L Elman}.}
  \bibinfo{year}{1990}\natexlab{}.
\newblock \showarticletitle{Finding structure in time}.
\newblock \bibinfo{journal}{{\em Cognitive science\/}} \bibinfo{volume}{14},
  \bibinfo{number}{2} (\bibinfo{year}{1990}), \bibinfo{pages}{179--211}.
\newblock


\bibitem[\protect\citeauthoryear{Fabbri, Li, She, Li, and Radev}{Fabbri
  et~al\mbox{.}}{2019}]%
        {fabbri2019multi}
\bibfield{author}{\bibinfo{person}{Alexander~Richard Fabbri},
  \bibinfo{person}{Irene Li}, \bibinfo{person}{Tianwei She},
  \bibinfo{person}{Suyi Li}, {and} \bibinfo{person}{Dragomir Radev}.}
  \bibinfo{year}{2019}\natexlab{}.
\newblock \showarticletitle{Multi-News: A Large-Scale Multi-Document
  Summarization Dataset and Abstractive Hierarchical Model}. In
  \bibinfo{booktitle}{{\em Proceedings of the 57th Annual Meeting of the
  Association for Computational Linguistics}}. \bibinfo{pages}{1074--1084}.
\newblock


\bibitem[\protect\citeauthoryear{Fan, Grangier, and Auli}{Fan
  et~al\mbox{.}}{2017}]%
        {fan2017controllable}
\bibfield{author}{\bibinfo{person}{Angela Fan}, \bibinfo{person}{David
  Grangier}, {and} \bibinfo{person}{Michael Auli}.}
  \bibinfo{year}{2017}\natexlab{}.
\newblock \showarticletitle{Controllable Abstractive Summarization}.
\newblock \bibinfo{journal}{{\em arXiv preprint arXiv:1711.05217\/}}
  (\bibinfo{year}{2017}).
\newblock


\bibitem[\protect\citeauthoryear{Fan, Grangier, and Auli}{Fan
  et~al\mbox{.}}{2018}]%
        {fan2018controllable}
\bibfield{author}{\bibinfo{person}{Angela Fan}, \bibinfo{person}{David
  Grangier}, {and} \bibinfo{person}{Michael Auli}.}
  \bibinfo{year}{2018}\natexlab{}.
\newblock \showarticletitle{Controllable Abstractive Summarization}. In
  \bibinfo{booktitle}{{\em Proceedings of the 2nd Workshop on Neural Machine
  Translation and Generation}}. \bibinfo{pages}{45--54}.
\newblock


\bibitem[\protect\citeauthoryear{Gambhir and Gupta}{Gambhir and Gupta}{2017}]%
        {gambhir2017recent}
\bibfield{author}{\bibinfo{person}{Mahak Gambhir} {and} \bibinfo{person}{Vishal
  Gupta}.} \bibinfo{year}{2017}\natexlab{}.
\newblock \showarticletitle{Recent automatic text summarization techniques: a
  survey}.
\newblock \bibinfo{journal}{{\em Artificial Intelligence Review\/}}
  \bibinfo{volume}{47}, \bibinfo{number}{1} (\bibinfo{year}{2017}),
  \bibinfo{pages}{1--66}.
\newblock


\bibitem[\protect\citeauthoryear{Gehring, Auli, Grangier, and Dauphin}{Gehring
  et~al\mbox{.}}{2017a}]%
        {gehring2016convolutional}
\bibfield{author}{\bibinfo{person}{Jonas Gehring}, \bibinfo{person}{Michael
  Auli}, \bibinfo{person}{David Grangier}, {and} \bibinfo{person}{Yann
  Dauphin}.} \bibinfo{year}{2017}\natexlab{a}.
\newblock \showarticletitle{A Convolutional Encoder Model for Neural Machine
  Translation}. In \bibinfo{booktitle}{{\em Proceedings of the 55th Annual
  Meeting of the Association for Computational Linguistics (Volume 1: Long
  Papers)}}, Vol.~\bibinfo{volume}{1}. \bibinfo{pages}{123--135}.
\newblock


\bibitem[\protect\citeauthoryear{Gehring, Auli, Grangier, Yarats, and
  Dauphin}{Gehring et~al\mbox{.}}{2017b}]%
        {gehring2017convolutional}
\bibfield{author}{\bibinfo{person}{Jonas Gehring}, \bibinfo{person}{Michael
  Auli}, \bibinfo{person}{David Grangier}, \bibinfo{person}{Denis Yarats},
  {and} \bibinfo{person}{Yann~N Dauphin}.} \bibinfo{year}{2017}\natexlab{b}.
\newblock \showarticletitle{Convolutional Sequence to Sequence Learning}. In
  \bibinfo{booktitle}{{\em International Conference on Machine Learning}}.
  \bibinfo{pages}{1243--1252}.
\newblock


\bibitem[\protect\citeauthoryear{Gehrmann, Deng, and Rush}{Gehrmann
  et~al\mbox{.}}{2018}]%
        {gehrmann2018bottom}
\bibfield{author}{\bibinfo{person}{Sebastian Gehrmann},
  \bibinfo{person}{Yuntian Deng}, {and} \bibinfo{person}{Alexander Rush}.}
  \bibinfo{year}{2018}\natexlab{}.
\newblock \showarticletitle{Bottom-Up Abstractive Summarization}. In
  \bibinfo{booktitle}{{\em Proceedings of the 2018 Conference on Empirical
  Methods in Natural Language Processing}}. \bibinfo{pages}{4098--4109}.
\newblock


\bibitem[\protect\citeauthoryear{Gimpel, Batra, Dyer, and Shakhnarovich}{Gimpel
  et~al\mbox{.}}{2013}]%
        {gimpel2013systematic}
\bibfield{author}{\bibinfo{person}{Kevin Gimpel}, \bibinfo{person}{Dhruv
  Batra}, \bibinfo{person}{Chris Dyer}, {and} \bibinfo{person}{Gregory
  Shakhnarovich}.} \bibinfo{year}{2013}\natexlab{}.
\newblock \showarticletitle{A systematic exploration of diversity in machine
  translation}. In \bibinfo{booktitle}{{\em Proceedings of the 2013 Conference
  on Empirical Methods in Natural Language Processing}}.
  \bibinfo{pages}{1100--1111}.
\newblock


\bibitem[\protect\citeauthoryear{Graves and Jaitly}{Graves and Jaitly}{2014}]%
        {graves2014towards}
\bibfield{author}{\bibinfo{person}{Alex Graves} {and} \bibinfo{person}{Navdeep
  Jaitly}.} \bibinfo{year}{2014}\natexlab{}.
\newblock \showarticletitle{Towards end-to-end speech recognition with
  recurrent neural networks}. In \bibinfo{booktitle}{{\em International
  Conference on Machine Learning}}. \bibinfo{pages}{1764--1772}.
\newblock


\bibitem[\protect\citeauthoryear{Grusky, Naaman, and Artzi}{Grusky
  et~al\mbox{.}}{2018}]%
        {grusky2018newsroom}
\bibfield{author}{\bibinfo{person}{Max Grusky}, \bibinfo{person}{Mor Naaman},
  {and} \bibinfo{person}{Yoav Artzi}.} \bibinfo{year}{2018}\natexlab{}.
\newblock \showarticletitle{Newsroom: A Dataset of 1.3 Million Summaries with
  Diverse Extractive Strategies}. In \bibinfo{booktitle}{{\em Proceedings of
  the 2018 Conference of the North American Chapter of the Association for
  Computational Linguistics: Human Language Technologies, Volume 1 (Long
  Papers)}}, Vol.~\bibinfo{volume}{1}. \bibinfo{pages}{708--719}.
\newblock


\bibitem[\protect\citeauthoryear{Gu, Lu, Li, and Li}{Gu et~al\mbox{.}}{2016}]%
        {gu2016incorporating}
\bibfield{author}{\bibinfo{person}{Jiatao Gu}, \bibinfo{person}{Zhengdong Lu},
  \bibinfo{person}{Hang Li}, {and} \bibinfo{person}{Victor~OK Li}.}
  \bibinfo{year}{2016}\natexlab{}.
\newblock \showarticletitle{Incorporating Copying Mechanism in
  Sequence-to-Sequence Learning}. In \bibinfo{booktitle}{{\em Proceedings of
  the 54th Annual Meeting of the Association for Computational Linguistics
  (Volume 1: Long Papers)}}, Vol.~\bibinfo{volume}{1}.
  \bibinfo{pages}{1631--1640}.
\newblock


\bibitem[\protect\citeauthoryear{Gulcehre, Ahn, Nallapati, Zhou, and
  Bengio}{Gulcehre et~al\mbox{.}}{2016}]%
        {gulcehre2016pointing}
\bibfield{author}{\bibinfo{person}{Caglar Gulcehre}, \bibinfo{person}{Sungjin
  Ahn}, \bibinfo{person}{Ramesh Nallapati}, \bibinfo{person}{Bowen Zhou}, {and}
  \bibinfo{person}{Yoshua Bengio}.} \bibinfo{year}{2016}\natexlab{}.
\newblock \showarticletitle{Pointing the Unknown Words}. In
  \bibinfo{booktitle}{{\em Proceedings of the 54th Annual Meeting of the
  Association for Computational Linguistics (Volume 1: Long Papers)}},
  Vol.~\bibinfo{volume}{1}. \bibinfo{pages}{140--149}.
\newblock


\bibitem[\protect\citeauthoryear{Guo, Pasunuru, and Bansal}{Guo
  et~al\mbox{.}}{2018}]%
        {guo2018soft}
\bibfield{author}{\bibinfo{person}{Han Guo}, \bibinfo{person}{Ramakanth
  Pasunuru}, {and} \bibinfo{person}{Mohit Bansal}.}
  \bibinfo{year}{2018}\natexlab{}.
\newblock \showarticletitle{Soft Layer-Specific Multi-Task Summarization with
  Entailment and Question Generation}. In \bibinfo{booktitle}{{\em Proceedings
  of the 56th Annual Meeting of the Association for Computational Linguistics
  (Volume 1: Long Papers)}}. \bibinfo{publisher}{Association for Computational
  Linguistics}, \bibinfo{pages}{687--697}.
\newblock


\bibitem[\protect\citeauthoryear{Guo and Sanner}{Guo and Sanner}{2010}]%
        {guo2010probabilistic}
\bibfield{author}{\bibinfo{person}{Shengbo Guo} {and} \bibinfo{person}{Scott
  Sanner}.} \bibinfo{year}{2010}\natexlab{}.
\newblock \showarticletitle{Probabilistic latent maximal marginal relevance}.
  In \bibinfo{booktitle}{{\em Proceedings of the 33rd international ACM SIGIR
  conference on Research and development in information retrieval}}. ACM,
  \bibinfo{pages}{833--834}.
\newblock


\bibitem[\protect\citeauthoryear{Haveliwala}{Haveliwala}{2002}]%
        {haveliwala2002topic}
\bibfield{author}{\bibinfo{person}{Taher~H Haveliwala}.}
  \bibinfo{year}{2002}\natexlab{}.
\newblock \showarticletitle{Topic-sensitive pagerank}. In
  \bibinfo{booktitle}{{\em Proceedings of the 11th international conference on
  World Wide Web}}. ACM, \bibinfo{pages}{517--526}.
\newblock


\bibitem[\protect\citeauthoryear{Hermann, Kocisky, Grefenstette, Espeholt, Kay,
  Suleyman, and Blunsom}{Hermann et~al\mbox{.}}{2015}]%
        {hermann2015teaching}
\bibfield{author}{\bibinfo{person}{Karl~Moritz Hermann}, \bibinfo{person}{Tomas
  Kocisky}, \bibinfo{person}{Edward Grefenstette}, \bibinfo{person}{Lasse
  Espeholt}, \bibinfo{person}{Will Kay}, \bibinfo{person}{Mustafa Suleyman},
  {and} \bibinfo{person}{Phil Blunsom}.} \bibinfo{year}{2015}\natexlab{}.
\newblock \showarticletitle{Teaching machines to read and comprehend}. In
  \bibinfo{booktitle}{{\em Advances in Neural Information Processing Systems}}.
  \bibinfo{pages}{1693--1701}.
\newblock


\bibitem[\protect\citeauthoryear{Hochreiter, Bengio, Frasconi, Schmidhuber,
  et~al\mbox{.}}{Hochreiter et~al\mbox{.}}{2001}]%
        {hochreiter2001gradient}
\bibfield{author}{\bibinfo{person}{Sepp Hochreiter}, \bibinfo{person}{Yoshua
  Bengio}, \bibinfo{person}{Paolo Frasconi}, \bibinfo{person}{J{\"u}rgen
  Schmidhuber}, {et~al\mbox{.}}} \bibinfo{year}{2001}\natexlab{}.
\newblock \bibinfo{title}{Gradient flow in recurrent nets: the difficulty of
  learning long-term dependencies}.
\newblock   (\bibinfo{year}{2001}).
\newblock


\bibitem[\protect\citeauthoryear{Hochreiter and Schmidhuber}{Hochreiter and
  Schmidhuber}{1997}]%
        {hochreiter1997long}
\bibfield{author}{\bibinfo{person}{Sepp Hochreiter} {and}
  \bibinfo{person}{J{\"u}rgen Schmidhuber}.} \bibinfo{year}{1997}\natexlab{}.
\newblock \showarticletitle{Long short-term memory}.
\newblock \bibinfo{journal}{{\em Neural computation\/}} \bibinfo{volume}{9},
  \bibinfo{number}{8} (\bibinfo{year}{1997}), \bibinfo{pages}{1735--1780}.
\newblock


\bibitem[\protect\citeauthoryear{Holtzman, Buys, Du, Forbes, and Choi}{Holtzman
  et~al\mbox{.}}{2019}]%
        {holtzman2019curious}
\bibfield{author}{\bibinfo{person}{Ari Holtzman}, \bibinfo{person}{Jan Buys},
  \bibinfo{person}{Li Du}, \bibinfo{person}{Maxwell Forbes}, {and}
  \bibinfo{person}{Yejin Choi}.} \bibinfo{year}{2019}\natexlab{}.
\newblock \showarticletitle{The curious case of neural text degeneration}.
\newblock \bibinfo{journal}{{\em arXiv preprint arXiv:1904.09751\/}}
  (\bibinfo{year}{2019}).
\newblock


\bibitem[\protect\citeauthoryear{Hsu, Lin, Lee, Min, Tang, and Sun}{Hsu
  et~al\mbox{.}}{2018}]%
        {hsu2018unified}
\bibfield{author}{\bibinfo{person}{Wan-Ting Hsu}, \bibinfo{person}{Chieh-Kai
  Lin}, \bibinfo{person}{Ming-Ying Lee}, \bibinfo{person}{Kerui Min},
  \bibinfo{person}{Jing Tang}, {and} \bibinfo{person}{Min Sun}.}
  \bibinfo{year}{2018}\natexlab{}.
\newblock \showarticletitle{A Unified Model for Extractive and Abstractive
  Summarization using Inconsistency Loss}.
\newblock \bibinfo{journal}{{\em arXiv preprint arXiv:1805.06266\/}}
  (\bibinfo{year}{2018}).
\newblock


\bibitem[\protect\citeauthoryear{Hu}{Hu}{2018}]%
        {hu2018introductory}
\bibfield{author}{\bibinfo{person}{Dichao Hu}.}
  \bibinfo{year}{2018}\natexlab{}.
\newblock \showarticletitle{An Introductory Survey on Attention Mechanisms in
  NLP Problems}.
\newblock \bibinfo{journal}{{\em arXiv preprint arXiv:1811.05544\/}}
  (\bibinfo{year}{2018}).
\newblock


\bibitem[\protect\citeauthoryear{Huang, Wu, and Wang}{Huang
  et~al\mbox{.}}{2020}]%
        {huang2020knowledge}
\bibfield{author}{\bibinfo{person}{Luyang Huang}, \bibinfo{person}{Lingfei Wu},
  {and} \bibinfo{person}{Lu Wang}.} \bibinfo{year}{2020}\natexlab{}.
\newblock \showarticletitle{Knowledge Graph-Augmented Abstractive Summarization
  with Semantic-Driven Cloze Reward}.
\newblock \bibinfo{journal}{{\em arXiv preprint arXiv:2005.01159\/}}
  (\bibinfo{year}{2020}).
\newblock


\bibitem[\protect\citeauthoryear{Inan, Khosravi, and Socher}{Inan
  et~al\mbox{.}}{2016}]%
        {inan2016tying}
\bibfield{author}{\bibinfo{person}{Hakan Inan}, \bibinfo{person}{Khashayar
  Khosravi}, {and} \bibinfo{person}{Richard Socher}.}
  \bibinfo{year}{2016}\natexlab{}.
\newblock \showarticletitle{Tying word vectors and word classifiers: A loss
  framework for language modeling}.
\newblock \bibinfo{journal}{{\em arXiv preprint arXiv:1611.01462\/}}
  (\bibinfo{year}{2016}).
\newblock


\bibitem[\protect\citeauthoryear{Ippolito, Kriz, Sedoc, Kustikova, and
  Callison-Burch}{Ippolito et~al\mbox{.}}{2019}]%
        {ippolito2019comparison}
\bibfield{author}{\bibinfo{person}{Daphne Ippolito}, \bibinfo{person}{Reno
  Kriz}, \bibinfo{person}{Joao Sedoc}, \bibinfo{person}{Maria Kustikova}, {and}
  \bibinfo{person}{Chris Callison-Burch}.} \bibinfo{year}{2019}\natexlab{}.
\newblock \showarticletitle{Comparison of Diverse Decoding Methods from
  Conditional Language Models}. In \bibinfo{booktitle}{{\em Proceedings of the
  57th Annual Meeting of the Association for Computational Linguistics}}.
  \bibinfo{pages}{3752--3762}.
\newblock


\bibitem[\protect\citeauthoryear{Jiang and Bansal}{Jiang and Bansal}{2018}]%
        {jiang2018closed}
\bibfield{author}{\bibinfo{person}{Yichen Jiang} {and} \bibinfo{person}{Mohit
  Bansal}.} \bibinfo{year}{2018}\natexlab{}.
\newblock \showarticletitle{Closed-Book Training to Improve Summarization
  Encoder Memory}. In \bibinfo{booktitle}{{\em Proceedings of the 2018
  Conference on Empirical Methods in Natural Language Processing}}.
  \bibinfo{pages}{4067--4077}.
\newblock


\bibitem[\protect\citeauthoryear{Kalchbrenner, Espeholt, Simonyan, Oord,
  Graves, and Kavukcuoglu}{Kalchbrenner et~al\mbox{.}}{2016}]%
        {kalchbrenner2016neural}
\bibfield{author}{\bibinfo{person}{Nal Kalchbrenner}, \bibinfo{person}{Lasse
  Espeholt}, \bibinfo{person}{Karen Simonyan}, \bibinfo{person}{Aaron van~den
  Oord}, \bibinfo{person}{Alex Graves}, {and} \bibinfo{person}{Koray
  Kavukcuoglu}.} \bibinfo{year}{2016}\natexlab{}.
\newblock \showarticletitle{Neural machine translation in linear time}.
\newblock \bibinfo{journal}{{\em arXiv preprint arXiv:1610.10099\/}}
  (\bibinfo{year}{2016}).
\newblock


\bibitem[\protect\citeauthoryear{Keneshloo, Ramakrishnan, and Reddy}{Keneshloo
  et~al\mbox{.}}{2019}]%
        {keneshloo2019deep}
\bibfield{author}{\bibinfo{person}{Yaser Keneshloo}, \bibinfo{person}{Naren
  Ramakrishnan}, {and} \bibinfo{person}{Chandan~K Reddy}.}
  \bibinfo{year}{2019}\natexlab{}.
\newblock \showarticletitle{Deep transfer reinforcement learning for text
  summarization}. In \bibinfo{booktitle}{{\em Proceedings of the 2019 SIAM
  International Conference on Data Mining}}. SIAM, \bibinfo{pages}{675--683}.
\newblock


\bibitem[\protect\citeauthoryear{Keneshloo, Shi, Ramakrishnan, and
  Reddy}{Keneshloo et~al\mbox{.}}{2018}]%
        {keneshloo2018deep}
\bibfield{author}{\bibinfo{person}{Yaser Keneshloo}, \bibinfo{person}{Tian
  Shi}, \bibinfo{person}{Naren Ramakrishnan}, {and} \bibinfo{person}{Chandan~K.
  Reddy}.} \bibinfo{year}{2018}\natexlab{}.
\newblock \showarticletitle{Deep Reinforcement Learning on Sequence to Sequence
  Models}.
\newblock \bibinfo{journal}{{\em arXiv preprint arXiv:\/}}
  (\bibinfo{year}{2018}).
\newblock


\bibitem[\protect\citeauthoryear{Kim, Kim, and Kim}{Kim et~al\mbox{.}}{2019}]%
        {kim2019abstractive}
\bibfield{author}{\bibinfo{person}{Byeongchang Kim}, \bibinfo{person}{Hyunwoo
  Kim}, {and} \bibinfo{person}{Gunhee Kim}.} \bibinfo{year}{2019}\natexlab{}.
\newblock \showarticletitle{Abstractive Summarization of Reddit Posts with
  Multi-level Memory Networks}. In \bibinfo{booktitle}{{\em Proceedings of the
  2019 Conference of the North American Chapter of the Association for
  Computational Linguistics: Human Language Technologies, Volume 1 (Long and
  Short Papers)}}. \bibinfo{pages}{2519--2531}.
\newblock


\bibitem[\protect\citeauthoryear{Kingma and Ba}{Kingma and Ba}{2014}]%
        {kingma2014adam}
\bibfield{author}{\bibinfo{person}{Diederik~P Kingma} {and}
  \bibinfo{person}{Jimmy Ba}.} \bibinfo{year}{2014}\natexlab{}.
\newblock \showarticletitle{Adam: A method for stochastic optimization}.
\newblock \bibinfo{journal}{{\em arXiv preprint arXiv:1412.6980\/}}
  (\bibinfo{year}{2014}).
\newblock


\bibitem[\protect\citeauthoryear{Kingma and Welling}{Kingma and
  Welling}{2013}]%
        {kingma2013auto}
\bibfield{author}{\bibinfo{person}{Diederik~P Kingma} {and}
  \bibinfo{person}{Max Welling}.} \bibinfo{year}{2013}\natexlab{}.
\newblock \showarticletitle{Auto-encoding variational bayes}.
\newblock \bibinfo{journal}{{\em arXiv preprint arXiv:1312.6114\/}}
  (\bibinfo{year}{2013}).
\newblock


\bibitem[\protect\citeauthoryear{Klein, Kim, Deng, Senellart, and Rush}{Klein
  et~al\mbox{.}}{2017}]%
        {klein2017opennmt}
\bibfield{author}{\bibinfo{person}{Guillaume Klein}, \bibinfo{person}{Yoon
  Kim}, \bibinfo{person}{Yuntian Deng}, \bibinfo{person}{Jean Senellart}, {and}
  \bibinfo{person}{Alexander Rush}.} \bibinfo{year}{2017}\natexlab{}.
\newblock \showarticletitle{{OpenNMT}: Open-Source Toolkit for Neural Machine
  Translation}.
\newblock \bibinfo{journal}{{\em Proceedings of ACL 2017, System
  Demonstrations\/}} (\bibinfo{year}{2017}), \bibinfo{pages}{67--72}.
\newblock


\bibitem[\protect\citeauthoryear{Koupaee and Wang}{Koupaee and Wang}{2018}]%
        {koupaee2018wikihow}
\bibfield{author}{\bibinfo{person}{Mahnaz Koupaee} {and}
  \bibinfo{person}{William~Yang Wang}.} \bibinfo{year}{2018}\natexlab{}.
\newblock \showarticletitle{Wikihow: A large scale text summarization dataset}.
\newblock \bibinfo{journal}{{\em arXiv preprint arXiv:1810.09305\/}}
  (\bibinfo{year}{2018}).
\newblock


\bibitem[\protect\citeauthoryear{Krause, Johnson, Krishna, and Fei-Fei}{Krause
  et~al\mbox{.}}{2017}]%
        {krause2017hierarchical}
\bibfield{author}{\bibinfo{person}{Jonathan Krause}, \bibinfo{person}{Justin
  Johnson}, \bibinfo{person}{Ranjay Krishna}, {and} \bibinfo{person}{Li
  Fei-Fei}.} \bibinfo{year}{2017}\natexlab{}.
\newblock \showarticletitle{A hierarchical approach for generating descriptive
  image paragraphs}. In \bibinfo{booktitle}{{\em Proceedings of the IEEE
  conference on computer vision and pattern recognition}}.
  \bibinfo{pages}{317--325}.
\newblock


\bibitem[\protect\citeauthoryear{Krizhevsky, Sutskever, and Hinton}{Krizhevsky
  et~al\mbox{.}}{2012}]%
        {krizhevsky2012imagenet}
\bibfield{author}{\bibinfo{person}{Alex Krizhevsky}, \bibinfo{person}{Ilya
  Sutskever}, {and} \bibinfo{person}{Geoffrey~E Hinton}.}
  \bibinfo{year}{2012}\natexlab{}.
\newblock \showarticletitle{Imagenet classification with deep convolutional
  neural networks}. In \bibinfo{booktitle}{{\em Advances in neural information
  processing systems}}. \bibinfo{pages}{1097--1105}.
\newblock


\bibitem[\protect\citeauthoryear{Kryscinski, Keskar, McCann, Xiong, and
  Socher}{Kryscinski et~al\mbox{.}}{2019}]%
        {kryscinski2019neural}
\bibfield{author}{\bibinfo{person}{Wojciech Kryscinski},
  \bibinfo{person}{Nitish~Shirish Keskar}, \bibinfo{person}{Bryan McCann},
  \bibinfo{person}{Caiming Xiong}, {and} \bibinfo{person}{Richard Socher}.}
  \bibinfo{year}{2019}\natexlab{}.
\newblock \showarticletitle{Neural text summarization: A critical evaluation}.
  In \bibinfo{booktitle}{{\em Proceedings of the 2019 Conference on Empirical
  Methods in Natural Language Processing and the 9th International Joint
  Conference on Natural Language Processing (EMNLP-IJCNLP)}}.
  \bibinfo{pages}{540--551}.
\newblock


\bibitem[\protect\citeauthoryear{Kry{\'s}ci{\'n}ski, Paulus, Xiong, and
  Socher}{Kry{\'s}ci{\'n}ski et~al\mbox{.}}{2018}]%
        {kryscinski2018improving}
\bibfield{author}{\bibinfo{person}{Wojciech Kry{\'s}ci{\'n}ski},
  \bibinfo{person}{Romain Paulus}, \bibinfo{person}{Caiming Xiong}, {and}
  \bibinfo{person}{Richard Socher}.} \bibinfo{year}{2018}\natexlab{}.
\newblock \showarticletitle{Improving Abstraction in Text Summarization}. In
  \bibinfo{booktitle}{{\em Proceedings of the 2018 Conference on Empirical
  Methods in Natural Language Processing}}. \bibinfo{pages}{1808--1817}.
\newblock


\bibitem[\protect\citeauthoryear{Lewis, Liu, Goyal, Ghazvininejad, Mohamed,
  Levy, Stoyanov, and Zettlemoyer}{Lewis et~al\mbox{.}}{2019}]%
        {lewis2019bart}
\bibfield{author}{\bibinfo{person}{Mike Lewis}, \bibinfo{person}{Yinhan Liu},
  \bibinfo{person}{Naman Goyal}, \bibinfo{person}{Marjan Ghazvininejad},
  \bibinfo{person}{Abdelrahman Mohamed}, \bibinfo{person}{Omer Levy},
  \bibinfo{person}{Ves Stoyanov}, {and} \bibinfo{person}{Luke Zettlemoyer}.}
  \bibinfo{year}{2019}\natexlab{}.
\newblock \showarticletitle{Bart: Denoising sequence-to-sequence pre-training
  for natural language generation, translation, and comprehension}.
\newblock \bibinfo{journal}{{\em arXiv preprint arXiv:1910.13461\/}}
  (\bibinfo{year}{2019}).
\newblock


\bibitem[\protect\citeauthoryear{Li, Xu, Li, and Gao}{Li
  et~al\mbox{.}}{2018b}]%
        {li2018guiding}
\bibfield{author}{\bibinfo{person}{Chenliang Li}, \bibinfo{person}{Weiran Xu},
  \bibinfo{person}{Si Li}, {and} \bibinfo{person}{Sheng Gao}.}
  \bibinfo{year}{2018}\natexlab{b}.
\newblock \showarticletitle{Guiding Generation for Abstractive Text
  Summarization Based on Key Information Guide Network}. In
  \bibinfo{booktitle}{{\em Proceedings of the 2018 Conference of the North
  American Chapter of the Association for Computational Linguistics: Human
  Language Technologies, Volume 2 (Short Papers)}}, Vol.~\bibinfo{volume}{2}.
  \bibinfo{pages}{55--60}.
\newblock


\bibitem[\protect\citeauthoryear{Li, Galley, Brockett, Gao, and Dolan}{Li
  et~al\mbox{.}}{2016a}]%
        {li2016diversity}
\bibfield{author}{\bibinfo{person}{Jiwei Li}, \bibinfo{person}{Michel Galley},
  \bibinfo{person}{Chris Brockett}, \bibinfo{person}{Jianfeng Gao}, {and}
  \bibinfo{person}{Bill Dolan}.} \bibinfo{year}{2016}\natexlab{a}.
\newblock \showarticletitle{A Diversity-Promoting Objective Function for Neural
  Conversation Models}. In \bibinfo{booktitle}{{\em Proceedings of the 2016
  Conference of the North American Chapter of the Association for Computational
  Linguistics: Human Language Technologies}}. \bibinfo{pages}{110--119}.
\newblock


\bibitem[\protect\citeauthoryear{Li and Jurafsky}{Li and Jurafsky}{2016}]%
        {li2016mutual}
\bibfield{author}{\bibinfo{person}{Jiwei Li} {and} \bibinfo{person}{Dan
  Jurafsky}.} \bibinfo{year}{2016}\natexlab{}.
\newblock \showarticletitle{Mutual information and diverse decoding improve
  neural machine translation}.
\newblock \bibinfo{journal}{{\em arXiv preprint arXiv:1601.00372\/}}
  (\bibinfo{year}{2016}).
\newblock


\bibitem[\protect\citeauthoryear{Li, Monroe, and Jurafsky}{Li
  et~al\mbox{.}}{2016b}]%
        {li2016simple}
\bibfield{author}{\bibinfo{person}{Jiwei Li}, \bibinfo{person}{Will Monroe},
  {and} \bibinfo{person}{Dan Jurafsky}.} \bibinfo{year}{2016}\natexlab{b}.
\newblock \showarticletitle{A simple, fast diverse decoding algorithm for
  neural generation}.
\newblock \bibinfo{journal}{{\em arXiv preprint arXiv:1611.08562\/}}
  (\bibinfo{year}{2016}).
\newblock


\bibitem[\protect\citeauthoryear{Li, Monroe, Ritter, Jurafsky, Galley, and
  Gao}{Li et~al\mbox{.}}{2016c}]%
        {li2016deep}
\bibfield{author}{\bibinfo{person}{Jiwei Li}, \bibinfo{person}{Will Monroe},
  \bibinfo{person}{Alan Ritter}, \bibinfo{person}{Dan Jurafsky},
  \bibinfo{person}{Michel Galley}, {and} \bibinfo{person}{Jianfeng Gao}.}
  \bibinfo{year}{2016}\natexlab{c}.
\newblock \showarticletitle{Deep Reinforcement Learning for Dialogue
  Generation}. In \bibinfo{booktitle}{{\em Proceedings of the 2016 Conference
  on Empirical Methods in Natural Language Processing}}.
  \bibinfo{pages}{1192--1202}.
\newblock


\bibitem[\protect\citeauthoryear{Li, Bing, and Lam}{Li et~al\mbox{.}}{2018a}]%
        {li2018actor}
\bibfield{author}{\bibinfo{person}{Piji Li}, \bibinfo{person}{Lidong Bing},
  {and} \bibinfo{person}{Wai Lam}.} \bibinfo{year}{2018}\natexlab{a}.
\newblock \showarticletitle{Actor-Critic based Training Framework for
  Abstractive Summarization}.
\newblock \bibinfo{journal}{{\em arXiv preprint arXiv:1803.11070\/}}
  (\bibinfo{year}{2018}).
\newblock


\bibitem[\protect\citeauthoryear{Li, Lam, Bing, and Wang}{Li
  et~al\mbox{.}}{2017}]%
        {li2017deep}
\bibfield{author}{\bibinfo{person}{Piji Li}, \bibinfo{person}{Wai Lam},
  \bibinfo{person}{Lidong Bing}, {and} \bibinfo{person}{Zihao Wang}.}
  \bibinfo{year}{2017}\natexlab{}.
\newblock \showarticletitle{Deep Recurrent Generative Decoder for Abstractive
  Text Summarization}. In \bibinfo{booktitle}{{\em Proceedings of the 2017
  Conference on Empirical Methods in Natural Language Processing}}.
  \bibinfo{pages}{2091--2100}.
\newblock


\bibitem[\protect\citeauthoryear{Lin}{Lin}{2004}]%
        {lin2004rouge}
\bibfield{author}{\bibinfo{person}{Chin-Yew Lin}.}
  \bibinfo{year}{2004}\natexlab{}.
\newblock \showarticletitle{{ROUGE}: A package for automatic evaluation of
  summaries}.
\newblock \bibinfo{journal}{{\em Text Summarization Branches Out\/}}
  (\bibinfo{year}{2004}).
\newblock


\bibitem[\protect\citeauthoryear{Lin, SUN, Ma, and Su}{Lin
  et~al\mbox{.}}{2018}]%
        {lin2018global}
\bibfield{author}{\bibinfo{person}{Junyang Lin}, \bibinfo{person}{Xu SUN},
  \bibinfo{person}{Shuming Ma}, {and} \bibinfo{person}{Qi Su}.}
  \bibinfo{year}{2018}\natexlab{}.
\newblock \showarticletitle{Global Encoding for Abstractive Summarization}. In
  \bibinfo{booktitle}{{\em Proceedings of the 56th Annual Meeting of the
  Association for Computational Linguistics (Volume 2: Short Papers)}}.
  \bibinfo{publisher}{Association for Computational Linguistics},
  \bibinfo{pages}{163--169}.
\newblock


\bibitem[\protect\citeauthoryear{Ling and Rush}{Ling and Rush}{2017}]%
        {ling2017coarse}
\bibfield{author}{\bibinfo{person}{Jeffrey Ling} {and}
  \bibinfo{person}{Alexander Rush}.} \bibinfo{year}{2017}\natexlab{}.
\newblock \showarticletitle{Coarse-to-Fine Attention Models for Document
  Summarization}. In \bibinfo{booktitle}{{\em Proceedings of the Workshop on
  New Frontiers in Summarization}}. \bibinfo{pages}{33--42}.
\newblock


\bibitem[\protect\citeauthoryear{Liu, Lu, Yang, Qu, Zhu, and Li}{Liu
  et~al\mbox{.}}{2017}]%
        {liu2017generative}
\bibfield{author}{\bibinfo{person}{Linqing Liu}, \bibinfo{person}{Yao Lu},
  \bibinfo{person}{Min Yang}, \bibinfo{person}{Qiang Qu}, \bibinfo{person}{Jia
  Zhu}, {and} \bibinfo{person}{Hongyan Li}.} \bibinfo{year}{2017}\natexlab{}.
\newblock \showarticletitle{Generative Adversarial Network for Abstractive Text
  Summarization}.
\newblock \bibinfo{journal}{{\em arXiv preprint arXiv:1711.09357\/}}
  (\bibinfo{year}{2017}).
\newblock


\bibitem[\protect\citeauthoryear{Liu, He, Chen, and Gao}{Liu
  et~al\mbox{.}}{2019a}]%
        {liu2019multi}
\bibfield{author}{\bibinfo{person}{Xiaodong Liu}, \bibinfo{person}{Pengcheng
  He}, \bibinfo{person}{Weizhu Chen}, {and} \bibinfo{person}{Jianfeng Gao}.}
  \bibinfo{year}{2019}\natexlab{a}.
\newblock \showarticletitle{Multi-Task Deep Neural Networks for Natural
  Language Understanding}. In \bibinfo{booktitle}{{\em Proceedings of the 57th
  Annual Meeting of the Association for Computational Linguistics}}.
  \bibinfo{pages}{4487--4496}.
\newblock


\bibitem[\protect\citeauthoryear{Liu and Lapata}{Liu and Lapata}{2019}]%
        {liu2019text}
\bibfield{author}{\bibinfo{person}{Yang Liu} {and} \bibinfo{person}{Mirella
  Lapata}.} \bibinfo{year}{2019}\natexlab{}.
\newblock \showarticletitle{Text Summarization with Pretrained Encoders}. In
  \bibinfo{booktitle}{{\em Proceedings of the 2019 Conference on Empirical
  Methods in Natural Language Processing and the 9th International Joint
  Conference on Natural Language Processing (EMNLP-IJCNLP)}}.
  \bibinfo{pages}{3721--3731}.
\newblock


\bibitem[\protect\citeauthoryear{Liu, Ott, Goyal, Du, Joshi, Chen, Levy, Lewis,
  Zettlemoyer, and Stoyanov}{Liu et~al\mbox{.}}{2019b}]%
        {liu2019roberta}
\bibfield{author}{\bibinfo{person}{Yinhan Liu}, \bibinfo{person}{Myle Ott},
  \bibinfo{person}{Naman Goyal}, \bibinfo{person}{Jingfei Du},
  \bibinfo{person}{Mandar Joshi}, \bibinfo{person}{Danqi Chen},
  \bibinfo{person}{Omer Levy}, \bibinfo{person}{Mike Lewis},
  \bibinfo{person}{Luke Zettlemoyer}, {and} \bibinfo{person}{Veselin
  Stoyanov}.} \bibinfo{year}{2019}\natexlab{b}.
\newblock \showarticletitle{Roberta: A robustly optimized bert pretraining
  approach}.
\newblock \bibinfo{journal}{{\em arXiv preprint arXiv:1907.11692\/}}
  (\bibinfo{year}{2019}).
\newblock


\bibitem[\protect\citeauthoryear{Lloret and Palomar}{Lloret and
  Palomar}{2012}]%
        {lloret2012text}
\bibfield{author}{\bibinfo{person}{Elena Lloret} {and} \bibinfo{person}{Manuel
  Palomar}.} \bibinfo{year}{2012}\natexlab{}.
\newblock \showarticletitle{Text summarisation in progress: a literature
  review}.
\newblock \bibinfo{journal}{{\em Artificial Intelligence Review\/}}
  \bibinfo{volume}{37}, \bibinfo{number}{1} (\bibinfo{year}{2012}),
  \bibinfo{pages}{1--41}.
\newblock


\bibitem[\protect\citeauthoryear{Lopyrev}{Lopyrev}{2015}]%
        {lopyrev2015generating}
\bibfield{author}{\bibinfo{person}{Konstantin Lopyrev}.}
  \bibinfo{year}{2015}\natexlab{}.
\newblock \showarticletitle{Generating news headlines with recurrent neural
  networks}.
\newblock \bibinfo{journal}{{\em arXiv preprint arXiv:1512.01712\/}}
  (\bibinfo{year}{2015}).
\newblock


\bibitem[\protect\citeauthoryear{Luong, Pham, and Manning}{Luong
  et~al\mbox{.}}{2015}]%
        {luong2015effective}
\bibfield{author}{\bibinfo{person}{Thang Luong}, \bibinfo{person}{Hieu Pham},
  {and} \bibinfo{person}{Christopher~D Manning}.}
  \bibinfo{year}{2015}\natexlab{}.
\newblock \showarticletitle{Effective Approaches to Attention-based Neural
  Machine Translation}. In \bibinfo{booktitle}{{\em Proceedings of the 2015
  Conference on Empirical Methods in Natural Language Processing}}.
  \bibinfo{pages}{1412--1421}.
\newblock


\bibitem[\protect\citeauthoryear{Mani and Maybury}{Mani and Maybury}{1999}]%
        {mani1999advances}
\bibfield{author}{\bibinfo{person}{Inderjeet Mani} {and}
  \bibinfo{person}{Mark~T Maybury}.} \bibinfo{year}{1999}\natexlab{}.
\newblock \bibinfo{booktitle}{{\em Advances in automatic text summarization}}.
\newblock \bibinfo{publisher}{MIT press}.
\newblock


\bibitem[\protect\citeauthoryear{Maynez, Narayan, Bohnet, and McDonald}{Maynez
  et~al\mbox{.}}{2020}]%
        {maynez2020faithfulness}
\bibfield{author}{\bibinfo{person}{Joshua Maynez}, \bibinfo{person}{Shashi
  Narayan}, \bibinfo{person}{Bernd Bohnet}, {and} \bibinfo{person}{Ryan
  McDonald}.} \bibinfo{year}{2020}\natexlab{}.
\newblock \showarticletitle{On Faithfulness and Factuality in Abstractive
  Summarization}.
\newblock \bibinfo{journal}{{\em arXiv preprint arXiv:2005.00661\/}}
  (\bibinfo{year}{2020}).
\newblock


\bibitem[\protect\citeauthoryear{McCann, Keskar, Xiong, and Socher}{McCann
  et~al\mbox{.}}{2018}]%
        {mccann2018natural}
\bibfield{author}{\bibinfo{person}{Bryan McCann},
  \bibinfo{person}{Nitish~Shirish Keskar}, \bibinfo{person}{Caiming Xiong},
  {and} \bibinfo{person}{Richard Socher}.} \bibinfo{year}{2018}\natexlab{}.
\newblock \showarticletitle{The Natural Language {Decathlon}: Multitask
  Learning as Question Answering}.
\newblock \bibinfo{journal}{{\em arXiv preprint arXiv:1806.08730\/}}
  (\bibinfo{year}{2018}).
\newblock


\bibitem[\protect\citeauthoryear{Miao and Blunsom}{Miao and Blunsom}{2016}]%
        {miao2016language}
\bibfield{author}{\bibinfo{person}{Yishu Miao} {and} \bibinfo{person}{Phil
  Blunsom}.} \bibinfo{year}{2016}\natexlab{}.
\newblock \showarticletitle{Language as a Latent Variable: Discrete Generative
  Models for Sentence Compression}. In \bibinfo{booktitle}{{\em Proceedings of
  the 2016 Conference on Empirical Methods in Natural Language Processing}}.
  \bibinfo{pages}{319--328}.
\newblock


\bibitem[\protect\citeauthoryear{Miao, Gowayyed, and Metze}{Miao
  et~al\mbox{.}}{2015}]%
        {miao2015eesen}
\bibfield{author}{\bibinfo{person}{Yajie Miao}, \bibinfo{person}{Mohammad
  Gowayyed}, {and} \bibinfo{person}{Florian Metze}.}
  \bibinfo{year}{2015}\natexlab{}.
\newblock \showarticletitle{{EESEN}: End-to-end speech recognition using deep
  RNN models and WFST-based decoding}. In \bibinfo{booktitle}{{\em Automatic
  Speech Recognition and Understanding (ASRU), 2015 IEEE Workshop on}}. IEEE,
  \bibinfo{pages}{167--174}.
\newblock


\bibitem[\protect\citeauthoryear{Mihalcea and Tarau}{Mihalcea and
  Tarau}{2004}]%
        {mihalcea2004textrank}
\bibfield{author}{\bibinfo{person}{Rada Mihalcea} {and} \bibinfo{person}{Paul
  Tarau}.} \bibinfo{year}{2004}\natexlab{}.
\newblock \showarticletitle{Textrank: Bringing order into text}. In
  \bibinfo{booktitle}{{\em Proceedings of the 2004 conference on empirical
  methods in natural language processing}}.
\newblock


\bibitem[\protect\citeauthoryear{Moratanch and Chitrakala}{Moratanch and
  Chitrakala}{2016}]%
        {moratanch2016survey}
\bibfield{author}{\bibinfo{person}{N Moratanch} {and} \bibinfo{person}{S
  Chitrakala}.} \bibinfo{year}{2016}\natexlab{}.
\newblock \showarticletitle{A survey on abstractive text summarization}. In
  \bibinfo{booktitle}{{\em Circuit, Power and Computing Technologies (ICCPCT),
  2016 International Conference on}}. IEEE, \bibinfo{pages}{1--7}.
\newblock


\bibitem[\protect\citeauthoryear{Nallapati, Zhai, and Zhou}{Nallapati
  et~al\mbox{.}}{2017}]%
        {nallapati2017summarunner}
\bibfield{author}{\bibinfo{person}{Ramesh Nallapati}, \bibinfo{person}{Feifei
  Zhai}, {and} \bibinfo{person}{Bowen Zhou}.} \bibinfo{year}{2017}\natexlab{}.
\newblock \showarticletitle{SummaRuNNer: A Recurrent Neural Network Based
  Sequence Model for Extractive Summarization of Documents.}
\newblock


\bibitem[\protect\citeauthoryear{Nallapati, Zhou, dos Santos, glar
  Gul{\c{c}}ehre, and Xiang}{Nallapati et~al\mbox{.}}{2016}]%
        {nallapati2016abstractive}
\bibfield{author}{\bibinfo{person}{Ramesh Nallapati}, \bibinfo{person}{Bowen
  Zhou}, \bibinfo{person}{Cicero dos Santos}, \bibinfo{person}{{\c{C}}a glar
  Gul{\c{c}}ehre}, {and} \bibinfo{person}{Bing Xiang}.}
  \bibinfo{year}{2016}\natexlab{}.
\newblock \showarticletitle{Abstractive Text Summarization using
  Sequence-to-sequence {RNNs} and Beyond}.
\newblock \bibinfo{journal}{{\em CoNLL 2016\/}} (\bibinfo{year}{2016}),
  \bibinfo{pages}{280}.
\newblock


\bibitem[\protect\citeauthoryear{Napoles, Gormley, and Van~Durme}{Napoles
  et~al\mbox{.}}{2012}]%
        {napoles2012annotated}
\bibfield{author}{\bibinfo{person}{Courtney Napoles}, \bibinfo{person}{Matthew
  Gormley}, {and} \bibinfo{person}{Benjamin Van~Durme}.}
  \bibinfo{year}{2012}\natexlab{}.
\newblock \showarticletitle{Annotated gigaword}. In \bibinfo{booktitle}{{\em
  Proceedings of the Joint Workshop on Automatic Knowledge Base Construction
  and Web-scale Knowledge Extraction}}. Association for Computational
  Linguistics, \bibinfo{pages}{95--100}.
\newblock


\bibitem[\protect\citeauthoryear{Narayan, Cohen, and Lapata}{Narayan
  et~al\mbox{.}}{2018}]%
        {narayan2018don}
\bibfield{author}{\bibinfo{person}{Shashi Narayan}, \bibinfo{person}{Shay~B
  Cohen}, {and} \bibinfo{person}{Mirella Lapata}.}
  \bibinfo{year}{2018}\natexlab{}.
\newblock \showarticletitle{Don’t Give Me the Details, Just the Summary!
  Topic-Aware Convolutional Neural Networks for Extreme Summarization}. In
  \bibinfo{booktitle}{{\em Proceedings of the 2018 Conference on Empirical
  Methods in Natural Language Processing}}. \bibinfo{pages}{1797--1807}.
\newblock


\bibitem[\protect\citeauthoryear{Nema, Khapra, Laha, and Ravindran}{Nema
  et~al\mbox{.}}{2017}]%
        {nema2017diversity}
\bibfield{author}{\bibinfo{person}{Preksha Nema}, \bibinfo{person}{Mitesh~M
  Khapra}, \bibinfo{person}{Anirban Laha}, {and} \bibinfo{person}{Balaraman
  Ravindran}.} \bibinfo{year}{2017}\natexlab{}.
\newblock \showarticletitle{Diversity driven attention model for query-based
  abstractive summarization}. In \bibinfo{booktitle}{{\em Proceedings of the
  55th Annual Meeting of the Association for Computational Linguistics (Volume
  1: Long Papers)}}, Vol.~\bibinfo{volume}{1}. \bibinfo{pages}{1063--1072}.
\newblock


\bibitem[\protect\citeauthoryear{Nenkova, McKeown, et~al\mbox{.}}{Nenkova
  et~al\mbox{.}}{2011}]%
        {nenkova2011automatic}
\bibfield{author}{\bibinfo{person}{Ani Nenkova}, \bibinfo{person}{Kathleen
  McKeown}, {et~al\mbox{.}}} \bibinfo{year}{2011}\natexlab{}.
\newblock \showarticletitle{Automatic summarization}.
\newblock \bibinfo{journal}{{\em Foundations and Trends{\textregistered} in
  Information Retrieval\/}} \bibinfo{volume}{5}, \bibinfo{number}{2--3}
  (\bibinfo{year}{2011}), \bibinfo{pages}{103--233}.
\newblock


\bibitem[\protect\citeauthoryear{Och}{Och}{2003}]%
        {och2003minimum}
\bibfield{author}{\bibinfo{person}{Franz~Josef Och}.}
  \bibinfo{year}{2003}\natexlab{}.
\newblock \showarticletitle{Minimum error rate training in statistical machine
  translation}. In \bibinfo{booktitle}{{\em Proceedings of the 41st Annual
  Meeting on Association for Computational Linguistics-Volume 1}}. Association
  for Computational Linguistics, \bibinfo{pages}{160--167}.
\newblock


\bibitem[\protect\citeauthoryear{Page, Brin, Motwani, and Winograd}{Page
  et~al\mbox{.}}{1999}]%
        {page1999pagerank}
\bibfield{author}{\bibinfo{person}{Lawrence Page}, \bibinfo{person}{Sergey
  Brin}, \bibinfo{person}{Rajeev Motwani}, {and} \bibinfo{person}{Terry
  Winograd}.} \bibinfo{year}{1999}\natexlab{}.
\newblock \bibinfo{booktitle}{{\em The PageRank citation ranking: Bringing
  order to the web.}}
\newblock \bibinfo{type}{{T}echnical {R}eport}. \bibinfo{institution}{Stanford
  InfoLab}.
\newblock


\bibitem[\protect\citeauthoryear{Papineni, Roukos, Ward, and Zhu}{Papineni
  et~al\mbox{.}}{2002}]%
        {papineni2002bleu}
\bibfield{author}{\bibinfo{person}{Kishore Papineni}, \bibinfo{person}{Salim
  Roukos}, \bibinfo{person}{Todd Ward}, {and} \bibinfo{person}{Wei-Jing Zhu}.}
  \bibinfo{year}{2002}\natexlab{}.
\newblock \showarticletitle{BLEU: a method for automatic evaluation of machine
  translation}. In \bibinfo{booktitle}{{\em Proceedings of the 40th annual
  meeting on association for computational linguistics}}. Association for
  Computational Linguistics, \bibinfo{pages}{311--318}.
\newblock


\bibitem[\protect\citeauthoryear{Pascanu, Mikolov, and Bengio}{Pascanu
  et~al\mbox{.}}{2013}]%
        {pascanu2013difficulty}
\bibfield{author}{\bibinfo{person}{Razvan Pascanu}, \bibinfo{person}{Tomas
  Mikolov}, {and} \bibinfo{person}{Yoshua Bengio}.}
  \bibinfo{year}{2013}\natexlab{}.
\newblock \showarticletitle{On the difficulty of training recurrent neural
  networks}. In \bibinfo{booktitle}{{\em International Conference on Machine
  Learning}}. \bibinfo{pages}{1310--1318}.
\newblock


\bibitem[\protect\citeauthoryear{Pasunuru and Bansal}{Pasunuru and
  Bansal}{2018}]%
        {pasunuru2018multi}
\bibfield{author}{\bibinfo{person}{Ramakanth Pasunuru} {and}
  \bibinfo{person}{Mohit Bansal}.} \bibinfo{year}{2018}\natexlab{}.
\newblock \showarticletitle{Multi-Reward Reinforced Summarization with Saliency
  and Entailment}. In \bibinfo{booktitle}{{\em Proceedings of the 2018
  Conference of the North American Chapter of the Association for Computational
  Linguistics: Human Language Technologies, Volume 2 (Short Papers)}},
  Vol.~\bibinfo{volume}{2}. \bibinfo{pages}{646--653}.
\newblock


\bibitem[\protect\citeauthoryear{Pasunuru, Guo, and Bansal}{Pasunuru
  et~al\mbox{.}}{2017}]%
        {pasunuru2017towards}
\bibfield{author}{\bibinfo{person}{Ramakanth Pasunuru}, \bibinfo{person}{Han
  Guo}, {and} \bibinfo{person}{Mohit Bansal}.} \bibinfo{year}{2017}\natexlab{}.
\newblock \showarticletitle{Towards Improving Abstractive Summarization via
  Entailment Generation}. In \bibinfo{booktitle}{{\em Proceedings of the
  Workshop on New Frontiers in Summarization}}. \bibinfo{pages}{27--32}.
\newblock


\bibitem[\protect\citeauthoryear{Paulus, Xiong, and Socher}{Paulus
  et~al\mbox{.}}{2017}]%
        {paulus2017deep}
\bibfield{author}{\bibinfo{person}{Romain Paulus}, \bibinfo{person}{Caiming
  Xiong}, {and} \bibinfo{person}{Richard Socher}.}
  \bibinfo{year}{2017}\natexlab{}.
\newblock \showarticletitle{A deep reinforced model for abstractive
  summarization}.
\newblock \bibinfo{journal}{{\em arXiv preprint arXiv:1705.04304\/}}
  (\bibinfo{year}{2017}).
\newblock


\bibitem[\protect\citeauthoryear{Rachabathuni}{Rachabathuni}{2017}]%
        {rachabathuni2017survey}
\bibfield{author}{\bibinfo{person}{Pavan~Kartheek Rachabathuni}.}
  \bibinfo{year}{2017}\natexlab{}.
\newblock \showarticletitle{A survey on abstractive summarization techniques}.
  In \bibinfo{booktitle}{{\em Inventive Computing and Informatics (ICICI),
  International Conference on}}. IEEE, \bibinfo{pages}{762--765}.
\newblock


\bibitem[\protect\citeauthoryear{Radev, Hovy, and McKeown}{Radev
  et~al\mbox{.}}{2002}]%
        {radev2002introduction}
\bibfield{author}{\bibinfo{person}{Dragomir~R Radev}, \bibinfo{person}{Eduard
  Hovy}, {and} \bibinfo{person}{Kathleen McKeown}.}
  \bibinfo{year}{2002}\natexlab{}.
\newblock \showarticletitle{Introduction to the special issue on
  summarization}.
\newblock \bibinfo{journal}{{\em Computational linguistics\/}}
  \bibinfo{volume}{28}, \bibinfo{number}{4} (\bibinfo{year}{2002}),
  \bibinfo{pages}{399--408}.
\newblock


\bibitem[\protect\citeauthoryear{Raffel, Shazeer, Roberts, Lee, Narang, Matena,
  Zhou, Li, and Liu}{Raffel et~al\mbox{.}}{2019}]%
        {raffel2019exploring}
\bibfield{author}{\bibinfo{person}{Colin Raffel}, \bibinfo{person}{Noam
  Shazeer}, \bibinfo{person}{Adam Roberts}, \bibinfo{person}{Katherine Lee},
  \bibinfo{person}{Sharan Narang}, \bibinfo{person}{Michael Matena},
  \bibinfo{person}{Yanqi Zhou}, \bibinfo{person}{Wei Li}, {and}
  \bibinfo{person}{Peter~J Liu}.} \bibinfo{year}{2019}\natexlab{}.
\newblock \showarticletitle{Exploring the limits of transfer learning with a
  unified text-to-text transformer}.
\newblock \bibinfo{journal}{{\em arXiv preprint arXiv:1910.10683\/}}
  (\bibinfo{year}{2019}).
\newblock


\bibitem[\protect\citeauthoryear{Ranzato, Chopra, Auli, and Zaremba}{Ranzato
  et~al\mbox{.}}{2015}]%
        {ranzato2015sequence}
\bibfield{author}{\bibinfo{person}{Marc'Aurelio Ranzato},
  \bibinfo{person}{Sumit Chopra}, \bibinfo{person}{Michael Auli}, {and}
  \bibinfo{person}{Wojciech Zaremba}.} \bibinfo{year}{2015}\natexlab{}.
\newblock \showarticletitle{Sequence level training with recurrent neural
  networks}.
\newblock \bibinfo{journal}{{\em arXiv preprint arXiv:1511.06732\/}}
  (\bibinfo{year}{2015}).
\newblock


\bibitem[\protect\citeauthoryear{Rennie, Marcheret, Mroueh, Ross, and
  Goel}{Rennie et~al\mbox{.}}{2017}]%
        {rennie2017self}
\bibfield{author}{\bibinfo{person}{Steven~J Rennie}, \bibinfo{person}{Etienne
  Marcheret}, \bibinfo{person}{Youssef Mroueh}, \bibinfo{person}{Jerret Ross},
  {and} \bibinfo{person}{Vaibhava Goel}.} \bibinfo{year}{2017}\natexlab{}.
\newblock \showarticletitle{Self-Critical Sequence Training for Image
  Captioning}. In \bibinfo{booktitle}{{\em Computer Vision and Pattern
  Recognition (CVPR), 2017 IEEE Conference on}}. IEEE,
  \bibinfo{pages}{1179--1195}.
\newblock


\bibitem[\protect\citeauthoryear{Rezende, Mohamed, and Wierstra}{Rezende
  et~al\mbox{.}}{2014}]%
        {rezende2014stochastic}
\bibfield{author}{\bibinfo{person}{Danilo~Jimenez Rezende},
  \bibinfo{person}{Shakir Mohamed}, {and} \bibinfo{person}{Daan Wierstra}.}
  \bibinfo{year}{2014}\natexlab{}.
\newblock \showarticletitle{Stochastic Backpropagation and Approximate
  Inference in Deep Generative Models}. In \bibinfo{booktitle}{{\em
  International Conference on Machine Learning}}. \bibinfo{pages}{1278--1286}.
\newblock


\bibitem[\protect\citeauthoryear{Rush, Chopra, and Weston}{Rush
  et~al\mbox{.}}{2015}]%
        {rush2015neural}
\bibfield{author}{\bibinfo{person}{Alexander~M Rush}, \bibinfo{person}{Sumit
  Chopra}, {and} \bibinfo{person}{Jason Weston}.}
  \bibinfo{year}{2015}\natexlab{}.
\newblock \showarticletitle{A Neural Attention Model for Abstractive Sentence
  Summarization}. In \bibinfo{booktitle}{{\em Proceedings of the 2015
  Conference on Empirical Methods in Natural Language Processing}}.
  \bibinfo{pages}{379--389}.
\newblock


\bibitem[\protect\citeauthoryear{Saggion and Poibeau}{Saggion and
  Poibeau}{2013}]%
        {saggion2013automatic}
\bibfield{author}{\bibinfo{person}{Horacio Saggion} {and}
  \bibinfo{person}{Thierry Poibeau}.} \bibinfo{year}{2013}\natexlab{}.
\newblock \showarticletitle{Automatic text summarization: Past, present and
  future}.
\newblock In \bibinfo{booktitle}{{\em Multi-source, multilingual information
  extraction and summarization}}. \bibinfo{publisher}{Springer},
  \bibinfo{pages}{3--21}.
\newblock


\bibitem[\protect\citeauthoryear{Sankaran, Mi, Al-Onaizan, and
  Ittycheriah}{Sankaran et~al\mbox{.}}{2016}]%
        {sankaran2016temporal}
\bibfield{author}{\bibinfo{person}{Baskaran Sankaran}, \bibinfo{person}{Haitao
  Mi}, \bibinfo{person}{Yaser Al-Onaizan}, {and} \bibinfo{person}{Abe
  Ittycheriah}.} \bibinfo{year}{2016}\natexlab{}.
\newblock \showarticletitle{Temporal Attention Model for Neural Machine
  Translation}.
\newblock \bibinfo{journal}{{\em CoRR\/}}  \bibinfo{volume}{abs/1608.02927}
  (\bibinfo{year}{2016}).
\newblock


\bibitem[\protect\citeauthoryear{See, Liu, and Manning}{See
  et~al\mbox{.}}{2017}]%
        {see2017get}
\bibfield{author}{\bibinfo{person}{Abigail See}, \bibinfo{person}{Peter~J.
  Liu}, {and} \bibinfo{person}{Christopher~D. Manning}.}
  \bibinfo{year}{2017}\natexlab{}.
\newblock \showarticletitle{Get To The Point: Summarization with
  Pointer-Generator Networks}. In \bibinfo{booktitle}{{\em Proceedings of the
  55th Annual Meeting of the Association for Computational Linguistics (Volume
  1: Long Papers)}}. \bibinfo{publisher}{Association for Computational
  Linguistics}, \bibinfo{pages}{1073--1083}.
\newblock


\bibitem[\protect\citeauthoryear{Sellam, Das, and Parikh}{Sellam
  et~al\mbox{.}}{2020}]%
        {sellam2020bleurt}
\bibfield{author}{\bibinfo{person}{Thibault Sellam}, \bibinfo{person}{Dipanjan
  Das}, {and} \bibinfo{person}{Ankur~P Parikh}.}
  \bibinfo{year}{2020}\natexlab{}.
\newblock \showarticletitle{BLEURT: Learning Robust Metrics for Text
  Generation}.
\newblock \bibinfo{journal}{{\em arXiv preprint arXiv:2004.04696\/}}
  (\bibinfo{year}{2020}).
\newblock


\bibitem[\protect\citeauthoryear{Sharma, Li, and Wang}{Sharma
  et~al\mbox{.}}{2019}]%
        {sharma2019bigpatent}
\bibfield{author}{\bibinfo{person}{Eva Sharma}, \bibinfo{person}{Chen Li},
  {and} \bibinfo{person}{Lu Wang}.} \bibinfo{year}{2019}\natexlab{}.
\newblock \showarticletitle{BIGPATENT: A Large-Scale Dataset for Abstractive
  and Coherent Summarization}. In \bibinfo{booktitle}{{\em Proceedings of the
  57th Annual Meeting of the Association for Computational Linguistics}}.
  \bibinfo{pages}{2204--2213}.
\newblock


\bibitem[\protect\citeauthoryear{Shen, Cheng, He, He, Wu, Sun, and Liu}{Shen
  et~al\mbox{.}}{2016}]%
        {shen2015minimum}
\bibfield{author}{\bibinfo{person}{Shiqi Shen}, \bibinfo{person}{Yong Cheng},
  \bibinfo{person}{Zhongjun He}, \bibinfo{person}{Wei He}, \bibinfo{person}{Hua
  Wu}, \bibinfo{person}{Maosong Sun}, {and} \bibinfo{person}{Yang Liu}.}
  \bibinfo{year}{2016}\natexlab{}.
\newblock \showarticletitle{Minimum Risk Training for Neural Machine
  Translation}. In \bibinfo{booktitle}{{\em Proceedings of the 54th Annual
  Meeting of the Association for Computational Linguistics (Volume 1: Long
  Papers)}}, Vol.~\bibinfo{volume}{1}. \bibinfo{pages}{1683--1692}.
\newblock


\bibitem[\protect\citeauthoryear{Shen, Lin, Tu, Zhao, Liu, Sun,
  et~al\mbox{.}}{Shen et~al\mbox{.}}{2017}]%
        {shen2017recent}
\bibfield{author}{\bibinfo{person}{Shi-Qi Shen}, \bibinfo{person}{Yan-Kai Lin},
  \bibinfo{person}{Cun-Chao Tu}, \bibinfo{person}{Yu Zhao},
  \bibinfo{person}{Zhi-Yuan Liu}, \bibinfo{person}{Mao-Song Sun},
  {et~al\mbox{.}}} \bibinfo{year}{2017}\natexlab{}.
\newblock \showarticletitle{Recent advances on neural headline generation}.
\newblock \bibinfo{journal}{{\em Journal of Computer Science and Technology\/}}
  \bibinfo{volume}{32}, \bibinfo{number}{4} (\bibinfo{year}{2017}),
  \bibinfo{pages}{768--784}.
\newblock


\bibitem[\protect\citeauthoryear{Shi, Wang, and Reddy}{Shi
  et~al\mbox{.}}{2019}]%
        {shi2019leafnats}
\bibfield{author}{\bibinfo{person}{Tian Shi}, \bibinfo{person}{Ping Wang},
  {and} \bibinfo{person}{Chandan~K Reddy}.} \bibinfo{year}{2019}\natexlab{}.
\newblock \showarticletitle{LeafNATS: An Open-Source Toolkit and Live Demo
  System for Neural Abstractive Text Summarization}. In
  \bibinfo{booktitle}{{\em Proceedings of the 2019 Conference of the North
  American Chapter of the Association for Computational Linguistics
  (Demonstrations)}}. \bibinfo{pages}{66--71}.
\newblock


\bibitem[\protect\citeauthoryear{Song, Zhao, and Liu}{Song
  et~al\mbox{.}}{2018}]%
        {song2018structure}
\bibfield{author}{\bibinfo{person}{Kaiqiang Song}, \bibinfo{person}{Lin Zhao},
  {and} \bibinfo{person}{Fei Liu}.} \bibinfo{year}{2018}\natexlab{}.
\newblock \showarticletitle{Structure-Infused Copy Mechanisms for Abstractive
  Summarization}. In \bibinfo{booktitle}{{\em Proceedings of the 27th
  International Conference on Computational Linguistics}}.
  \bibinfo{publisher}{Association for Computational Linguistics},
  \bibinfo{pages}{1717--1729}.
\newblock


\bibitem[\protect\citeauthoryear{Sutskever}{Sutskever}{2013}]%
        {sutskever2013training}
\bibfield{author}{\bibinfo{person}{Ilya Sutskever}.}
  \bibinfo{year}{2013}\natexlab{}.
\newblock \bibinfo{booktitle}{{\em Training recurrent neural networks}}.
\newblock \bibinfo{publisher}{University of Toronto Toronto, Ontario, Canada}.
\newblock


\bibitem[\protect\citeauthoryear{Sutskever, Vinyals, and Le}{Sutskever
  et~al\mbox{.}}{2014}]%
        {sutskever2014sequence}
\bibfield{author}{\bibinfo{person}{Ilya Sutskever}, \bibinfo{person}{Oriol
  Vinyals}, {and} \bibinfo{person}{Quoc~V Le}.}
  \bibinfo{year}{2014}\natexlab{}.
\newblock \showarticletitle{Sequence to sequence learning with neural
  networks}. In \bibinfo{booktitle}{{\em Advances in neural information
  processing systems}}. \bibinfo{pages}{3104--3112}.
\newblock


\bibitem[\protect\citeauthoryear{Sutton and Barto}{Sutton and Barto}{1998}]%
        {sutton1998reinforcement}
\bibfield{author}{\bibinfo{person}{Richard~S Sutton} {and}
  \bibinfo{person}{Andrew~G Barto}.} \bibinfo{year}{1998}\natexlab{}.
\newblock \bibinfo{booktitle}{{\em Reinforcement learning: An introduction}}.
\newblock \bibinfo{publisher}{MIT press}.
\newblock


\bibitem[\protect\citeauthoryear{Takase, Suzuki, Okazaki, Hirao, and
  Nagata}{Takase et~al\mbox{.}}{2016}]%
        {takase2016neural}
\bibfield{author}{\bibinfo{person}{Sho Takase}, \bibinfo{person}{Jun Suzuki},
  \bibinfo{person}{Naoaki Okazaki}, \bibinfo{person}{Tsutomu Hirao}, {and}
  \bibinfo{person}{Masaaki Nagata}.} \bibinfo{year}{2016}\natexlab{}.
\newblock \showarticletitle{Neural headline generation on abstract meaning
  representation}. In \bibinfo{booktitle}{{\em Proceedings of the 2016
  Conference on Empirical Methods in Natural Language Processing}}.
  \bibinfo{pages}{1054--1059}.
\newblock


\bibitem[\protect\citeauthoryear{Tan, Wan, and Xiao}{Tan et~al\mbox{.}}{2017}]%
        {tan2017abstractive}
\bibfield{author}{\bibinfo{person}{Jiwei Tan}, \bibinfo{person}{Xiaojun Wan},
  {and} \bibinfo{person}{Jianguo Xiao}.} \bibinfo{year}{2017}\natexlab{}.
\newblock \showarticletitle{Abstractive document summarization with a
  graph-based attentional neural model}. In \bibinfo{booktitle}{{\em
  Proceedings of the 55th Annual Meeting of the Association for Computational
  Linguistics (Volume 1: Long Papers)}}, Vol.~\bibinfo{volume}{1}.
  \bibinfo{pages}{1171--1181}.
\newblock


\bibitem[\protect\citeauthoryear{Tu, Lu, Liu, Liu, and Li}{Tu
  et~al\mbox{.}}{2016}]%
        {tu2016modeling}
\bibfield{author}{\bibinfo{person}{Zhaopeng Tu}, \bibinfo{person}{Zhengdong
  Lu}, \bibinfo{person}{Yang Liu}, \bibinfo{person}{Xiaohua Liu}, {and}
  \bibinfo{person}{Hang Li}.} \bibinfo{year}{2016}\natexlab{}.
\newblock \showarticletitle{Modeling Coverage for Neural Machine Translation}.
  In \bibinfo{booktitle}{{\em Proceedings of the 54th Annual Meeting of the
  Association for Computational Linguistics (Volume 1: Long Papers)}},
  Vol.~\bibinfo{volume}{1}. \bibinfo{pages}{76--85}.
\newblock


\bibitem[\protect\citeauthoryear{van~den Oord, Dieleman, Zen, Simonyan,
  Vinyals, Graves, Kalchbrenner, Senior, and Kavukcuoglu}{van~den Oord
  et~al\mbox{.}}{[n. d.]}]%
        {oord2016wavenet}
\bibfield{author}{\bibinfo{person}{A{\"a}ron van~den Oord},
  \bibinfo{person}{Sander Dieleman}, \bibinfo{person}{Heiga Zen},
  \bibinfo{person}{Karen Simonyan}, \bibinfo{person}{Oriol Vinyals},
  \bibinfo{person}{Alex Graves}, \bibinfo{person}{Nal Kalchbrenner},
  \bibinfo{person}{Andrew Senior}, {and} \bibinfo{person}{Koray Kavukcuoglu}.}
  \bibinfo{year}{[n. d.]}\natexlab{}.
\newblock \showarticletitle{WaveNet: A Generative Model for Raw Audio}. In
  \bibinfo{booktitle}{{\em 9th ISCA Speech Synthesis Workshop}}.
  \bibinfo{pages}{125--125}.
\newblock


\bibitem[\protect\citeauthoryear{Vaswani, Shazeer, Parmar, Uszkoreit, Jones,
  Gomez, Kaiser, and Polosukhin}{Vaswani et~al\mbox{.}}{2017}]%
        {vaswani2017attention}
\bibfield{author}{\bibinfo{person}{Ashish Vaswani}, \bibinfo{person}{Noam
  Shazeer}, \bibinfo{person}{Niki Parmar}, \bibinfo{person}{Jakob Uszkoreit},
  \bibinfo{person}{Llion Jones}, \bibinfo{person}{Aidan~N Gomez},
  \bibinfo{person}{{\L}ukasz Kaiser}, {and} \bibinfo{person}{Illia
  Polosukhin}.} \bibinfo{year}{2017}\natexlab{}.
\newblock \showarticletitle{Attention is all you need}. In
  \bibinfo{booktitle}{{\em Advances in Neural Information Processing Systems}}.
  \bibinfo{pages}{6000--6010}.
\newblock


\bibitem[\protect\citeauthoryear{Venkatraman, Hebert, and Bagnell}{Venkatraman
  et~al\mbox{.}}{2015}]%
        {venkatraman2015improving}
\bibfield{author}{\bibinfo{person}{Arun Venkatraman}, \bibinfo{person}{Martial
  Hebert}, {and} \bibinfo{person}{J~Andrew Bagnell}.}
  \bibinfo{year}{2015}\natexlab{}.
\newblock \showarticletitle{Improving Multi-Step Prediction of Learned Time
  Series Models}. In \bibinfo{booktitle}{{\em Twenty-Ninth AAAI Conference on
  Artificial Intelligence}}.
\newblock


\bibitem[\protect\citeauthoryear{Verma and Lee}{Verma and Lee}{2017}]%
        {verma2017extractive}
\bibfield{author}{\bibinfo{person}{Rakesh~M. Verma} {and}
  \bibinfo{person}{Daniel Lee}.} \bibinfo{year}{2017}\natexlab{}.
\newblock \showarticletitle{Extractive Summarization: Limits, Compression,
  Generalized Model and Heuristics}.
\newblock \bibinfo{journal}{{\em Computaci{\'o}n y Sistemas\/}}
  \bibinfo{volume}{21} (\bibinfo{year}{2017}).
\newblock


\bibitem[\protect\citeauthoryear{Vijayakumar, Cogswell, Selvaraju, Sun, Lee,
  Crandall, and Batra}{Vijayakumar et~al\mbox{.}}{2016}]%
        {vijayakumar2016diverse}
\bibfield{author}{\bibinfo{person}{Ashwin~K Vijayakumar},
  \bibinfo{person}{Michael Cogswell}, \bibinfo{person}{Ramprasath~R Selvaraju},
  \bibinfo{person}{Qing Sun}, \bibinfo{person}{Stefan Lee},
  \bibinfo{person}{David Crandall}, {and} \bibinfo{person}{Dhruv Batra}.}
  \bibinfo{year}{2016}\natexlab{}.
\newblock \showarticletitle{Diverse beam search: Decoding diverse solutions
  from neural sequence models}.
\newblock \bibinfo{journal}{{\em arXiv preprint arXiv:1610.02424\/}}
  (\bibinfo{year}{2016}).
\newblock


\bibitem[\protect\citeauthoryear{Vinyals, Fortunato, and Jaitly}{Vinyals
  et~al\mbox{.}}{2015}]%
        {vinyals2015pointer}
\bibfield{author}{\bibinfo{person}{Oriol Vinyals}, \bibinfo{person}{Meire
  Fortunato}, {and} \bibinfo{person}{Navdeep Jaitly}.}
  \bibinfo{year}{2015}\natexlab{}.
\newblock \showarticletitle{Pointer networks}. In \bibinfo{booktitle}{{\em
  Advances in Neural Information Processing Systems}}.
  \bibinfo{pages}{2692--2700}.
\newblock


\bibitem[\protect\citeauthoryear{Wang, Yao, Tao, Zhong, Liu, and Du}{Wang
  et~al\mbox{.}}{2018}]%
        {wang2018reinforced}
\bibfield{author}{\bibinfo{person}{Li Wang}, \bibinfo{person}{Junlin Yao},
  \bibinfo{person}{Yunzhe Tao}, \bibinfo{person}{Li Zhong},
  \bibinfo{person}{Wei Liu}, {and} \bibinfo{person}{Qiang Du}.}
  \bibinfo{year}{2018}\natexlab{}.
\newblock \showarticletitle{A reinforced topic-aware convolutional
  sequence-to-sequence model for abstractive text summarization}. In
  \bibinfo{booktitle}{{\em Proceedings of the 27th International Joint
  Conference on Artificial Intelligence}}. AAAI Press,
  \bibinfo{pages}{4453--4460}.
\newblock


\bibitem[\protect\citeauthoryear{Weaver and Tao}{Weaver and Tao}{2001}]%
        {weaver2001optimal}
\bibfield{author}{\bibinfo{person}{Lex Weaver} {and} \bibinfo{person}{Nigel
  Tao}.} \bibinfo{year}{2001}\natexlab{}.
\newblock \showarticletitle{The optimal reward baseline for gradient-based
  reinforcement learning}. In \bibinfo{booktitle}{{\em Proceedings of the
  Seventeenth conference on Uncertainty in artificial intelligence}}. Morgan
  Kaufmann Publishers Inc., \bibinfo{pages}{538--545}.
\newblock


\bibitem[\protect\citeauthoryear{Wei~Zhao}{Wei~Zhao}{2019}]%
        {zhao2019moverscore}
\bibfield{author}{\bibinfo{person}{Fei Liu Yang Gao Christian M. Meyer
  Steffen~Eger Wei~Zhao, Maxime~Peyrard}.} \bibinfo{year}{2019}\natexlab{}.
\newblock \showarticletitle{MoverScore: Text Generation Evaluating with
  Contextualized Embeddings and Earth Mover Distance}. In
  \bibinfo{booktitle}{{\em Proceedings of the 2019 Conference on Empirical
  Methods in Natural Language Processing}}. \bibinfo{publisher}{Association for
  Computational Linguistics}, \bibinfo{address}{Hong Kong, China}.
\newblock


\bibitem[\protect\citeauthoryear{Werbos}{Werbos}{1990}]%
        {werbos1990backpropagation}
\bibfield{author}{\bibinfo{person}{Paul~J Werbos}.}
  \bibinfo{year}{1990}\natexlab{}.
\newblock \showarticletitle{Backpropagation through time: what it does and how
  to do it}.
\newblock \bibinfo{journal}{{\it Proc. IEEE}} \bibinfo{volume}{78},
  \bibinfo{number}{10} (\bibinfo{year}{1990}), \bibinfo{pages}{1550--1560}.
\newblock


\bibitem[\protect\citeauthoryear{Williams}{Williams}{1992}]%
        {williams1992simple}
\bibfield{author}{\bibinfo{person}{Ronald~J Williams}.}
  \bibinfo{year}{1992}\natexlab{}.
\newblock \showarticletitle{Simple statistical gradient-following algorithms
  for connectionist reinforcement learning}.
\newblock In \bibinfo{booktitle}{{\em Reinforcement Learning}}.
  \bibinfo{publisher}{Springer}, \bibinfo{pages}{5--32}.
\newblock


\bibitem[\protect\citeauthoryear{Williams and Zipser}{Williams and
  Zipser}{1989}]%
        {williams1989learning}
\bibfield{author}{\bibinfo{person}{Ronald~J Williams} {and}
  \bibinfo{person}{David Zipser}.} \bibinfo{year}{1989}\natexlab{}.
\newblock \showarticletitle{A learning algorithm for continually running fully
  recurrent neural networks}.
\newblock \bibinfo{journal}{{\em Neural computation\/}} \bibinfo{volume}{1},
  \bibinfo{number}{2} (\bibinfo{year}{1989}), \bibinfo{pages}{270--280}.
\newblock


\bibitem[\protect\citeauthoryear{Wu and Hu}{Wu and Hu}{2018}]%
        {wu2018learning}
\bibfield{author}{\bibinfo{person}{Yuxiang Wu} {and} \bibinfo{person}{Baotian
  Hu}.} \bibinfo{year}{2018}\natexlab{}.
\newblock \bibinfo{title}{Learning to Extract Coherent Summary via Deep
  Reinforcement Learning}.
\newblock   (\bibinfo{year}{2018}).
\newblock


\bibitem[\protect\citeauthoryear{Wu, Schuster, Chen, Le, Norouzi, Macherey,
  Krikun, Cao, Gao, Macherey, et~al\mbox{.}}{Wu et~al\mbox{.}}{2016}]%
        {wu2016google}
\bibfield{author}{\bibinfo{person}{Yonghui Wu}, \bibinfo{person}{Mike
  Schuster}, \bibinfo{person}{Zhifeng Chen}, \bibinfo{person}{Quoc~V Le},
  \bibinfo{person}{Mohammad Norouzi}, \bibinfo{person}{Wolfgang Macherey},
  \bibinfo{person}{Maxim Krikun}, \bibinfo{person}{Yuan Cao},
  \bibinfo{person}{Qin Gao}, \bibinfo{person}{Klaus Macherey}, {et~al\mbox{.}}}
  \bibinfo{year}{2016}\natexlab{}.
\newblock \showarticletitle{Google's neural machine translation system:
  Bridging the gap between human and machine translation}.
\newblock \bibinfo{journal}{{\em arXiv preprint arXiv:1609.08144\/}}
  (\bibinfo{year}{2016}).
\newblock


\bibitem[\protect\citeauthoryear{Xia, Tian, Wu, Lin, Qin, Yu, and Liu}{Xia
  et~al\mbox{.}}{2017}]%
        {xia2017deliberation}
\bibfield{author}{\bibinfo{person}{Yingce Xia}, \bibinfo{person}{Fei Tian},
  \bibinfo{person}{Lijun Wu}, \bibinfo{person}{Jianxin Lin},
  \bibinfo{person}{Tao Qin}, \bibinfo{person}{Nenghai Yu}, {and}
  \bibinfo{person}{Tie-Yan Liu}.} \bibinfo{year}{2017}\natexlab{}.
\newblock \showarticletitle{Deliberation networks: Sequence generation beyond
  one-pass decoding}. In \bibinfo{booktitle}{{\em Advances in Neural
  Information Processing Systems}}. \bibinfo{pages}{1782--1792}.
\newblock


\bibitem[\protect\citeauthoryear{Xu, Ba, Kiros, Cho, Courville, Salakhudinov,
  Zemel, and Bengio}{Xu et~al\mbox{.}}{2015}]%
        {xu2015show}
\bibfield{author}{\bibinfo{person}{Kelvin Xu}, \bibinfo{person}{Jimmy Ba},
  \bibinfo{person}{Ryan Kiros}, \bibinfo{person}{Kyunghyun Cho},
  \bibinfo{person}{Aaron Courville}, \bibinfo{person}{Ruslan Salakhudinov},
  \bibinfo{person}{Rich Zemel}, {and} \bibinfo{person}{Yoshua Bengio}.}
  \bibinfo{year}{2015}\natexlab{}.
\newblock \showarticletitle{Show, attend and tell: Neural image caption
  generation with visual attention}. In \bibinfo{booktitle}{{\em International
  conference on machine learning}}. \bibinfo{pages}{2048--2057}.
\newblock


\bibitem[\protect\citeauthoryear{Yan, Qi, Gong, Liu, Duan, Chen, Zhang, and
  Zhou}{Yan et~al\mbox{.}}{2020}]%
        {yan2020prophetnet}
\bibfield{author}{\bibinfo{person}{Yu Yan}, \bibinfo{person}{Weizhen Qi},
  \bibinfo{person}{Yeyun Gong}, \bibinfo{person}{Dayiheng Liu},
  \bibinfo{person}{Nan Duan}, \bibinfo{person}{Jiusheng Chen},
  \bibinfo{person}{Ruofei Zhang}, {and} \bibinfo{person}{Ming Zhou}.}
  \bibinfo{year}{2020}\natexlab{}.
\newblock \showarticletitle{ProphetNet: Predicting Future N-gram for
  Sequence-to-Sequence Pre-training}.
\newblock \bibinfo{journal}{{\em arXiv preprint arXiv:2001.04063\/}}
  (\bibinfo{year}{2020}).
\newblock


\bibitem[\protect\citeauthoryear{Yang, Dai, Yang, Carbonell, Salakhutdinov, and
  Le}{Yang et~al\mbox{.}}{2019}]%
        {yang2019xlnet}
\bibfield{author}{\bibinfo{person}{Zhilin Yang}, \bibinfo{person}{Zihang Dai},
  \bibinfo{person}{Yiming Yang}, \bibinfo{person}{Jaime Carbonell},
  \bibinfo{person}{Russ~R Salakhutdinov}, {and} \bibinfo{person}{Quoc~V Le}.}
  \bibinfo{year}{2019}\natexlab{}.
\newblock \showarticletitle{Xlnet: Generalized autoregressive pretraining for
  language understanding}. In \bibinfo{booktitle}{{\em Advances in neural
  information processing systems}}. \bibinfo{pages}{5754--5764}.
\newblock


\bibitem[\protect\citeauthoryear{Yang, Yang, Dyer, He, Smola, and Hovy}{Yang
  et~al\mbox{.}}{2016}]%
        {yang2016hierarchical}
\bibfield{author}{\bibinfo{person}{Zichao Yang}, \bibinfo{person}{Diyi Yang},
  \bibinfo{person}{Chris Dyer}, \bibinfo{person}{Xiaodong He},
  \bibinfo{person}{Alex Smola}, {and} \bibinfo{person}{Eduard Hovy}.}
  \bibinfo{year}{2016}\natexlab{}.
\newblock \showarticletitle{Hierarchical attention networks for document
  classification}. In \bibinfo{booktitle}{{\em Proceedings of the 2016
  Conference of the North American Chapter of the Association for Computational
  Linguistics: Human Language Technologies}}. \bibinfo{pages}{1480--1489}.
\newblock


\bibitem[\protect\citeauthoryear{Zaremba and Sutskever}{Zaremba and
  Sutskever}{2015}]%
        {zaremba2015reinforcement}
\bibfield{author}{\bibinfo{person}{Wojciech Zaremba} {and}
  \bibinfo{person}{Ilya Sutskever}.} \bibinfo{year}{2015}\natexlab{}.
\newblock \showarticletitle{Reinforcement learning neural turing
  machines-revised}.
\newblock \bibinfo{journal}{{\em arXiv preprint arXiv:1505.00521\/}}
  (\bibinfo{year}{2015}).
\newblock


\bibitem[\protect\citeauthoryear{Zeng, Luo, Fidler, and Urtasun}{Zeng
  et~al\mbox{.}}{2016}]%
        {zeng2016efficient}
\bibfield{author}{\bibinfo{person}{Wenyuan Zeng}, \bibinfo{person}{Wenjie Luo},
  \bibinfo{person}{Sanja Fidler}, {and} \bibinfo{person}{Raquel Urtasun}.}
  \bibinfo{year}{2016}\natexlab{}.
\newblock \showarticletitle{Efficient summarization with read-again and copy
  mechanism}.
\newblock \bibinfo{journal}{{\em arXiv preprint arXiv:1611.03382\/}}
  (\bibinfo{year}{2016}).
\newblock


\bibitem[\protect\citeauthoryear{Zhang, Zhao, Saleh, and Liu}{Zhang
  et~al\mbox{.}}{2019b}]%
        {zhang2019pegasus}
\bibfield{author}{\bibinfo{person}{Jingqing Zhang}, \bibinfo{person}{Yao Zhao},
  \bibinfo{person}{Mohammad Saleh}, {and} \bibinfo{person}{Peter~J Liu}.}
  \bibinfo{year}{2019}\natexlab{b}.
\newblock \showarticletitle{PEGASUS: Pre-training with Extracted Gap-sentences
  for Abstractive Summarization}.
\newblock \bibinfo{journal}{{\em arXiv preprint arXiv:1912.08777\/}}
  (\bibinfo{year}{2019}).
\newblock


\bibitem[\protect\citeauthoryear{Zhang, Kishore, Wu, Weinberger, and
  Artzi}{Zhang et~al\mbox{.}}{2019a}]%
        {zhang2019bertscore}
\bibfield{author}{\bibinfo{person}{Tianyi Zhang}, \bibinfo{person}{Varsha
  Kishore}, \bibinfo{person}{Felix Wu}, \bibinfo{person}{Kilian~Q Weinberger},
  {and} \bibinfo{person}{Yoav Artzi}.} \bibinfo{year}{2019}\natexlab{a}.
\newblock \showarticletitle{Bertscore: Evaluating text generation with bert}.
\newblock \bibinfo{journal}{{\em arXiv preprint arXiv:1904.09675\/}}
  (\bibinfo{year}{2019}).
\newblock


\bibitem[\protect\citeauthoryear{Zhang and Lapata}{Zhang and Lapata}{2017}]%
        {zhang2017sentence}
\bibfield{author}{\bibinfo{person}{Xingxing Zhang} {and}
  \bibinfo{person}{Mirella Lapata}.} \bibinfo{year}{2017}\natexlab{}.
\newblock \showarticletitle{Sentence Simplification with Deep Reinforcement
  Learning}. In \bibinfo{booktitle}{{\em Proceedings of the 2017 Conference on
  Empirical Methods in Natural Language Processing}}.
  \bibinfo{pages}{584--594}.
\newblock


\bibitem[\protect\citeauthoryear{Zhang, Lapata, Wei, and Zhou}{Zhang
  et~al\mbox{.}}{2018b}]%
        {zhang2018neural}
\bibfield{author}{\bibinfo{person}{Xingxing Zhang}, \bibinfo{person}{Mirella
  Lapata}, \bibinfo{person}{Furu Wei}, {and} \bibinfo{person}{Ming Zhou}.}
  \bibinfo{year}{2018}\natexlab{b}.
\newblock \showarticletitle{Neural Latent Extractive Document Summarization}.
  In \bibinfo{booktitle}{{\em Proceedings of the 2018 Conference on Empirical
  Methods in Natural Language Processing}}. \bibinfo{pages}{779--784}.
\newblock


\bibitem[\protect\citeauthoryear{Zhang, Ding, Qian, Manning, and
  Langlotz}{Zhang et~al\mbox{.}}{2018a}]%
        {zhang2018learning}
\bibfield{author}{\bibinfo{person}{Yuhao Zhang}, \bibinfo{person}{Daisy~Yi
  Ding}, \bibinfo{person}{Tianpei Qian}, \bibinfo{person}{Christopher~D
  Manning}, {and} \bibinfo{person}{Curtis~P Langlotz}.}
  \bibinfo{year}{2018}\natexlab{a}.
\newblock \showarticletitle{Learning to Summarize Radiology Findings}.
\newblock \bibinfo{journal}{{\em EMNLP 2018\/}} (\bibinfo{year}{2018}),
  \bibinfo{pages}{204}.
\newblock


\bibitem[\protect\citeauthoryear{Zhou, Yang, Wei, Huang, Zhou, and Zhao}{Zhou
  et~al\mbox{.}}{2018}]%
        {zhou2018neural}
\bibfield{author}{\bibinfo{person}{Qingyu Zhou}, \bibinfo{person}{Nan Yang},
  \bibinfo{person}{Furu Wei}, \bibinfo{person}{Shaohan Huang},
  \bibinfo{person}{Ming Zhou}, {and} \bibinfo{person}{Tiejun Zhao}.}
  \bibinfo{year}{2018}\natexlab{}.
\newblock \showarticletitle{Neural Document Summarization by Jointly Learning
  to Score and Select Sentences}. In \bibinfo{booktitle}{{\em Proceedings of
  the 56th Annual Meeting of the Association for Computational Linguistics
  (Volume 1: Long Papers)}}, Vol.~\bibinfo{volume}{1}.
  \bibinfo{pages}{654--663}.
\newblock


\bibitem[\protect\citeauthoryear{Zhou, Yang, Wei, and Zhou}{Zhou
  et~al\mbox{.}}{2017}]%
        {zhou2017selective}
\bibfield{author}{\bibinfo{person}{Qingyu Zhou}, \bibinfo{person}{Nan Yang},
  \bibinfo{person}{Furu Wei}, {and} \bibinfo{person}{Ming Zhou}.}
  \bibinfo{year}{2017}\natexlab{}.
\newblock \showarticletitle{Selective Encoding for Abstractive Sentence
  Summarization}. In \bibinfo{booktitle}{{\em Proceedings of the 55th Annual
  Meeting of the Association for Computational Linguistics (Volume 1: Long
  Papers)}}, Vol.~\bibinfo{volume}{1}. \bibinfo{pages}{1095--1104}.
\newblock


\end{thebibliography}

\end{document}